\documentclass[acmsmall,screen]{acmart}

\AtBeginDocument{%
  }

\setcopyright{acmcopyright}
\copyrightyear{2024}
\acmYear{2024}

\makeatletter
\@printpermissionfalse
\@printcopyrightfalse
\@acmownedfalse
\@ACM@nonacmtrue
\makeatother

\usepackage{graphics}
\usepackage{multirow, makecell}
\usepackage{algorithmic}
\usepackage{algorithm}

\usepackage{xcolor}
\usepackage{tikz}
\usepackage[edges]{forest}
\definecolor{hidden-draw}{RGB}{20,68,106}
\definecolor{hidden-pink}{RGB}{255,245,247}
\definecolor{deepgreen}{rgb}{0.0, 0.5, 0.0} 
\definecolor{deepblue}{rgb}{0.0, 0.0, 0.5} 
\usepackage{amssymb}
\usepackage[utf8]{inputenc}
\usepackage[T1]{fontenc}
\usepackage{placeins}
\usepackage{array}
\usepackage{graphicx}
\usepackage{colortbl}
\usepackage{caption}
\usepackage{subcaption}
\usepackage{pbox}
\usepackage{pifont}
\begin{document}

\title{Next Token Prediction Towards Multimodal Intelligence: A Comprehensive Survey}

\author{Liang Chen$^{1\dagger}$ Zekun Wang$^{2*}$ Shuhuai Ren$^{1*}$ Lei Li$^{3*}$ Haozhe Zhao$^{1*}$ Yunshui Li$^{4*}$ Zefan Cai$^1$ Hongcheng Guo$^2$  Lei Zhang$^4$  Yizhe Xiong$^5$ Yichi Zhang$^1$ Ruoyu Wu$^1$ Qingxiu Dong$^1$ Ge Zhang$^6$ Jian Yang$^8$ Lingwei Meng$^7$ Shujie Hu$^7$ Yulong Chen$^9$ Junyang Lin$^8$ Shuai Bai$^8$  Andreas Vlachos$^9$ Xu Tan$^{10}$ Minjia Zhang$^{11}$ Wen Xiao$^{10}$ Aaron Yee$^{12,13}$ Tianyu Liu$^8$ Baobao Chang$^1$}


\renewcommand{\shortauthors}{Chen et al.}

\renewcommand{\thefootnote}{}

\begin{abstract} 
\textbf{Abstract:} Building on the foundations of language modeling in natural language processing, Next Token Prediction (NTP) has evolved into a versatile training objective for machine learning tasks across various modalities, achieving considerable success. As Large Language Models (LLMs) have advanced to unify understanding and generation tasks within the textual modality, recent research has shown that tasks from different modalities can also be effectively encapsulated within the NTP framework, transforming the multimodal information into tokens and predict the next one given the context.  This survey introduces a comprehensive taxonomy that unifies both understanding and generation within multimodal learning through the lens of NTP. The proposed taxonomy covers five key aspects: Multimodal tokenization, MMNTP model architectures, unified task representation, datasets \& evaluation, and open challenges. 
This new taxonomy aims to aid researchers in their exploration of multimodal intelligence. An associated GitHub repository collecting the latest papers and repos is available at \href{https://github.com/LMM101/Awesome-Multimodal-Next-Token-Prediction}{https://github.com/LMM101/Awesome-Multimodal-Next-Token-Prediction}. \footnotetext{\textbf{Authors' affiliations:} $^1$Peking University $^2$Beihang Univerisy $^3$University of Hongkong $^4$Shenzhen Institute of Advanced  Technology, China Academy of Sciences $^5$Tsinghua University $^6$M-A-P $^{7}$The Chinese University of Hong Kong $^8$Alibaba Group $^9$University of Cambridge $^{10}$Microsoft Research $^{11}$UIUC $^{12}$Humanify Inc. $^{13}$Zhejiang University}   
\footnotetext{\textbf{Authors' contributions:} $^\dagger$ project lead. $^*$ core contributors. Full contributions are in Section~\ref{sec: ack}. }
\footnotetext{\textbf{Corresponding to:} Liang Chen <\href{mailto:leo.liang.chen@outlook.com}{leo.liang.chen@outlook.com}>, Baobao Chang <\href{mailto:chbb@pku.edu.cn}{chbb@pku.edu.cn}>}
\end{abstract}




\maketitle 
\renewcommand{\thefootnote}{\arabic{footnote}}

\begin{figure}[h]
\centering
\includegraphics[width=\textwidth]{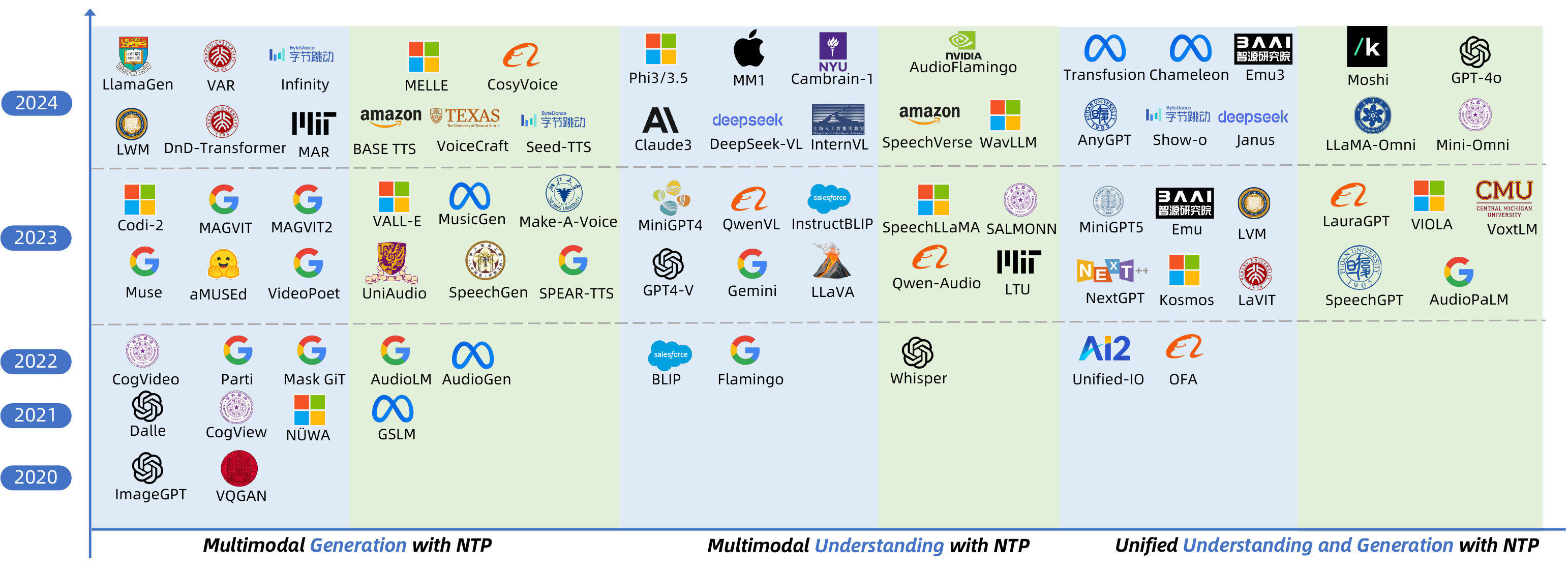}  
\caption{Historical development of LMMs utilizing Next-Token Prediction. Models featuring vision and more modalities are set in \textcolor{deepblue}{blue background} while models that support audio modality are set in \textcolor{deepgreen}{green backgrounds.}}
\label{fig:history}
\end{figure}

\section{Introduction}
\label{sec:intro}
Humans' engagement with the universe is a tapestry, interwoven with the threads of various modalities. Humans can see and sketch paintings, read and write epics, listen and compose music, touch and sculpture heroes, ponder and make movements. 
These modalities -- specific information types such as vision, sound, and language -- are the channels through which humans interpret and respond to the world. This multifaceted interaction highlights the intertwined nature of perception and response in human experience. As a specialized area within Artificial Intelligence (AI) research, multimodal Learning focuses on creating systems capable of understanding and generating various multimodal information~\citep{10.1109/TPAMI.2018.2798607}. 

\newcommand{\tbd}[1]{\textcolor{blue}{(#1 --TBD)}}



\begin{figure}[t]
\centering
\includegraphics[width=1\textwidth]{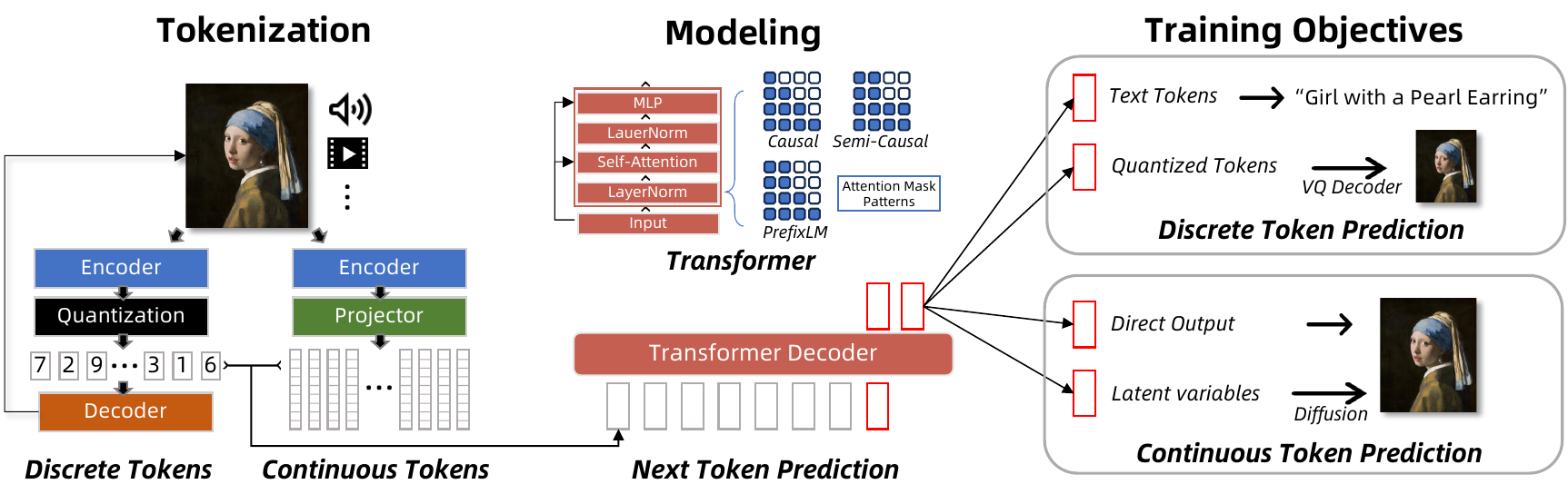}
\caption{General pipeline of Multimodal Learning with Next Token Prediction (MMNTP).}
\label{fig:methodology}
\end{figure}

A paradigm shift has emerged in the field of AI across multiple modalities, transitioning from specialized unimodal models trained for a single task to versatile multimodal ones dealing with a diverse array of tasks~\citep{hao2022languagemodelsgeneralpurposeinterfaces}. This shift is largely attributed to the advancement of Large Language Models (LLMs) in the Natural Language Processing (NLP) field such as GPT-3~\citep{gpt3}, ChatGPT~\citep{chatgpt} and LLaMA~\citep{touvron2023llama}, which unify multiple natural language understanding and generation tasks with a single Next Token Prediction (NTP) objective. The original task of NTP is to predict the next token (which can be a word, subword, or character) in a given sequence of text based on the context provided by preceding tokens. The NTP paradigm has been proven to be scalable given abundant data and computational resources in the lens of scaling law research~\cite{kaplan2020scalinglaw,zhai2022scaling}.

Simultaneously, researchers have explored the incorporation of non-textual input and output modalities into large language models, sparking interest within the community to develop powerful Large Multimodal Models (LMMs) featuring capabilities to conduct tasks across different modalities~\citep{cui2023surveymultimodallargelanguage,yin2024surveymultimodallargelanguage}. For a better understanding of the historical development of LMMs based on NTP, we demonstrate a timeline in Figure~\ref{fig:history}, categorized by models' understanding or generation ability and different modalities. 

In \autoref{fig:methodology}, we use the image modality as an example to illustrate the workflow of \textbf{Multimodal Learning with NTP (MMNTP)}. The process can be divided into three key components: Tokenization, Modeling, and Training Objectives, which will be explained and discussed in details in the rest of the survey. For vision modality, image and video understanding capabilities have been demonstrated in large vision-language models such as GPT4-V~\citep{gpt4v}, QwenVL~\citep{QwenVL}, LLaVA~\citep{liu2023llava}, Kosmos~\citep{peng2023kosmos2groundingmultimodallarge,lv2024kosmos25multimodalliteratemodel,huang2023languageneedaligningperception}, Phi-Vision~\citep{phi3_5} and Gemini~\citep{gemini1}, while Emu~\citep{sun2024emu} and Chameleon~\citep{chameleonteam2024chameleon} show visual generation could be achieved in NTP manner. Similarly, end-to-end audio understanding and generation have been achieved in NTP-based models such as GPT4-o and Moshi \citep{gpt4o,défossez2024moshispeechtextfoundationmodel}.


To equip LLMs with visual understanding capabilities, pioneering research such as Flamingo~\cite{alayrac2022flamingo}, BLIP2~\cite{li2023blip2}, GPT4V~\cite{gpt4v}, MiniGPT4~\cite{zhu2023minigpt4} and LLaVA~\cite{liu2023llava} has demonstrated that LLMs can be easily adapted to process multimodal inputs such as images and videos, by converting multimodal information into tokens with a straightforward tokenization module, such as a visual encoder like CLIP~\cite{radford2021clip} or a simple linear projection~\cite{fuyu}. Subsequently, these models perform multimodal instruction tuning based on image-query-answer triples using the same NTP objective. 

As Large Language Models bridge understanding and generation tasks in natural language processing, there is considerable interest in extending their capabilities to generate multimodal outputs. Recent advances in this direction include GPT-4o~\cite{gpt4o}, which can understand and generate text, audio, and images using a unified multimodal LLM. We have also witnessed tremendous improvements from the open source community. For visual modality, Chameleon~\cite{chameleonteam2024chameleon} and Emu3~\citep{wang2024emu3nexttokenpredictionneed} are two distinctive multimodal  that unify understanding and generation in both language and image modalities. For audio, Moshi~\citep{défossez2024moshispeechtextfoundationmodel} can conduct tasks such as automatic speech recognition (ASR) and speech generation in an NTP manner based a pretrained LLM. As a general and fundamental approach, NTP also has promising implications for diverse fields like AI-for-Science such as designing proteins in biology~\citep{benegas2024genomiclanguagemodelsopportunities} and composing molecule structure in chemistry~\citep{flam2022language}.


To generate multimodal content using the NTP approach, it is crucial to recognize that unlike language, which is structured from discrete symbols, multimodal data like images and sounds inherently exist in a continuous space. A common technique to address this challenge is quantization. Vector Quantization (VQ) is a classical method that allows for the modeling of probability density functions for continuous multimodal data through discrete vector distributions~\cite{Pags2015IntroductionTV,vq}. This technique aligns well with NTP modeling. With the rise of deep learning, neural VQ methods such as VQVAE~\cite{Oord2017NeuralDR} and VQGAN~\cite{Esser2020TamingTF} have been developed, establishing a foundation for linking visual and audio generation with NTP. Significant work has emerged leveraging these VQ methodologies and the language modeling task. Examples include innovative systems such as DALL-E~\cite{ramesh2021zeroshot}, CogView~\cite{CogView}, CM3Leon~\cite{cm3leon}, Parti~\cite{text2image2}, Muse~\cite{chang2023muse}, VideoPoet~\cite{kondratyuk2023videopoet}, LVM~\cite{bai2023sequential}, Chameleon~\cite{chameleonteam2024chameleon} and Infinity~\citep{han2024infinityscalingbitwiseautoregressive}. These methods often rely on external models, like VQGAN decoders, for image generation, making their approach a form of indirect multimodal generation. Parallel explorations have been conducted utilizing the NTP objective to directly generate images in continuous spaces, such as VAE's latent space~\cite{tschannen2024givt}, or by simulating a diffusion process~\cite{li2024autoregressive-without,Transfusion}. Unlike the indirect methods, only a few initiatives like ImageGPT~\citep{imagegpt} perform direct multimodal generation by predicting pixels from scratch. Additionally, NTP models can be augmented with various external models to facilitate multimodal generation. Notable examples include Emu~\cite{sun2023emu1}, MiniGPT5~\cite{zheng2023minigpt5}, and CoDi2~\cite{tang2023codi2}. These approaches utilize the NTP framework to incorporate external diffusion models for image generation, showcasing another form of indirect multimodal generation.



We have covered powerful models that can understand or generate information across different modalities within the NTP paradigm. However, developing a single model that can both comprehend and produce information across multiple modalities, similar to human abilities, remains an intriguing goal in the pursuit of Artificial General Intelligence (AGI). Recently, a new research trend has emerged, focusing on the development of LMMs that unifies multimodal understanding and generation in the NTP paradigm. Notable examples include Unified-IO~\citep{lu2023unifiedio,lu2023unifiedio2}, Chameleon~\citep{chameleonteam2024chameleon}, Transfusion~\citep{Transfusion}, Show-o~\citep{Show-o}, Moshi~\citep{Moshi}, and Emu3~\citep{Emu3}. Unifying understanding and generation presents unique challenges, including the diversity of modalities and resolving conflicts between them. We will discuss these issues further in Section~\ref{sec:challenges}.

\begin{table}[t]
\centering
\caption{Key Tables and Figures of the Survey.}
\label{table:key_tables}
\resizebox{1\textwidth}{!}{
\begin{tabular}{llcc}
\toprule
 \textbf{Content} & \textbf{Section}  & \textbf{Reference} & \textbf{Examples/Key-Words}  \\\midrule
\multicolumn{4}{c}{\textbf{\textit{Tables}}}\\\midrule

Multimodal Tokenizers & \S \ref{sec:tokenization} Multimodal Tokenization & Table~\ref{table:multimodal_tokenization} & VQVAE~\citep{rolfe2017discrete}, CLIP~\citep{radford2021clip}, HuBERT~\citep{hsu2021hubert}  \\
MMNTP Models & \S \ref{sec:model} Backbone Model for Multimodal Next Token Prediction & Table~\ref{table:mmntp_structure_summary} & Flamingo~\citep{alayrac2022flamingo},DALLE~\citep{DALLE}, Unified-IO~\citep{lu2022unifiedio} \\
Multimodal Prompt Engineering & \S \ref{sec:training} Training with Unified Multimodal Task Representation & Table~\ref{table:multimodal_ICL_summary} & Multimodal ICL, Multimodal CoT\\
Training Dataset & \S \ref{sec:datasets} Datasets and Evaluation & Table~\ref{tab:data_pt}, ~\ref{tab:data_ins} & mC4~\cite{xue2021mt5}, Laion-5B~\cite{laion5b},LLaVA~\citep{liu2023llava} \\
Evaluation Dataset & \S \ref{sec:datasets} Datasets and Evaluation & Table~\ref{tab:benchmarks_tab} & MME \citep{fu2023mme},MMBench \citep{liu2023mmbench},MMMU \citep{yue2023mmmu}  \\ \midrule
\multicolumn{4}{c}{\textbf{\textit{Figures}}} \\ \midrule

Historical Development of MMNTP Models  & \S \ref{sec:intro} Introduction  & Fig.~\ref{fig:history} & Multimodal Generation and Understanding \\
Pipeline of MMNTP & \S \ref{sec:intro} Introduction& Fig. ~\ref{fig:methodology} & Tokenization, Modeling, Training Objectives \\
Illustration of Multimodal tokenizations & \S \ref{sec:tokenization} Multimodal Tokenization & Fig. ~\ref{fig:tokenizations} & Discrete Tokens, Continous Tokens\\
Tokenizer Training Methods & \S \ref{sec:tokenization} Multimodal Tokenization & Fig. ~\ref{fig:keyfeatures} & Auto-Encoding, Contrastive Learning, ...\\
Illustration of Discrete Tokenization (VQ) & \S \ref{sec:tokenization} Multimodal Tokenization & Fig. ~\ref{fig:vqvae} & VQVAE, Quantization, Codebook \\
Illustration of Continuous Tokenization & \S \ref{sec:tokenization} Multimodal Tokenization & Fig. ~\ref{fig:soft-tokens} & Encoder, Decoder, Aligner \\
Categorization of MMNTP Models & \S \ref{sec:model} Backbone Model for Multimodal Next Token Prediction & Fig. ~\ref{fig:two type of MMNTP models} & Compositional Model, Unified Model \\
Backbone of MMNTP Models & \S \ref{sec:model} Backbone Model for Multimodal Next Token Prediction & Fig. ~\ref{fig:arch} &  Multimodal Transformer \\
Attention Patterns of MMNTP Models & \S \ref{sec:model} Backbone Model for Multimodal Next Token Prediction & Fig. ~\ref{fig:attn-mask} & Causal, Semi-Causal, Prefix ... \\
Unified Structures for Vision Tasks & \S \ref{sec:model} Backbone Model for Multimodal Next Token Prediction & Fig. ~\ref{fig:image_ntp} & VQA, Text-to-Image, Image-to-Image, ... \\
Unified Structures for Audio Tasks & \S \ref{sec:model} Backbone Model for Multimodal Next Token Prediction & Fig. ~\ref{fig:audio_ntp} & Audio Understanding, Audio Generation, ... \\
Training Objectives Explain & \S \ref{sec:training} Training with Unified Multimodal Task Representation & Fig. ~\ref{fig:training_obj} & Discrete/Continuous Token Prediction \\
Training Stage Overview & \S \ref{sec:training} Training with Unified Multimodal Task Representation & Fig. ~\ref{fig:training_stage} & Alignment, Instruction, Preference \\
Examples of Multimodal Prompt Engineering & \S \ref{sec:training} Training with Unified Multimodal Task Representation & Fig. ~\ref{fig:inference_examples} & Flamingo~\citep{flam2022language}, PCA-Bench~\citep{chen2024pcabench} \\
Explanation of Multimodal Prompt Engineering & \S \ref{sec:training} Training with Unified Multimodal Task Representation & Fig. ~\ref{fig:inference_methods} & ICL, CoT \\
Performance Comparisons on Understanding Tasks& \S \ref{sec:datasets} Datasets and Evaluation & Fig. ~\ref{fig:comp_on_understanding} & VQAv2~\citep{balanced_vqa_v2}, MMMU~\citep{yue2023mmmu} \\
Performance Comparisons on Generation Tasks & \S \ref{sec:datasets} Datasets and Evaluation & Fig. ~\ref{fig:fig:comp_on_generation} & Imagenet~\citep{russakovsky2015imagenet}, GenEval~\citep{ghosh2023genevalobjectfocusedframeworkevaluating} \\

\bottomrule
\end{tabular}}
\end{table}


\tikzstyle{my-box}=[
    rectangle,
    draw=hidden-draw,
    rounded corners,
    text opacity=1,
    minimum height=1.5em,
    minimum width=5em,
    inner sep=2pt,
    align=center,
    fill opacity=.5,
    line width=0.8pt,
]
\tikzstyle{leaf}=[my-box, minimum height=1.5em,
    fill=hidden-pink!80, text=black, align=left,font=\tiny,
    inner xsep=2pt,
    inner ysep=4pt,
    line width=0.8pt,
]

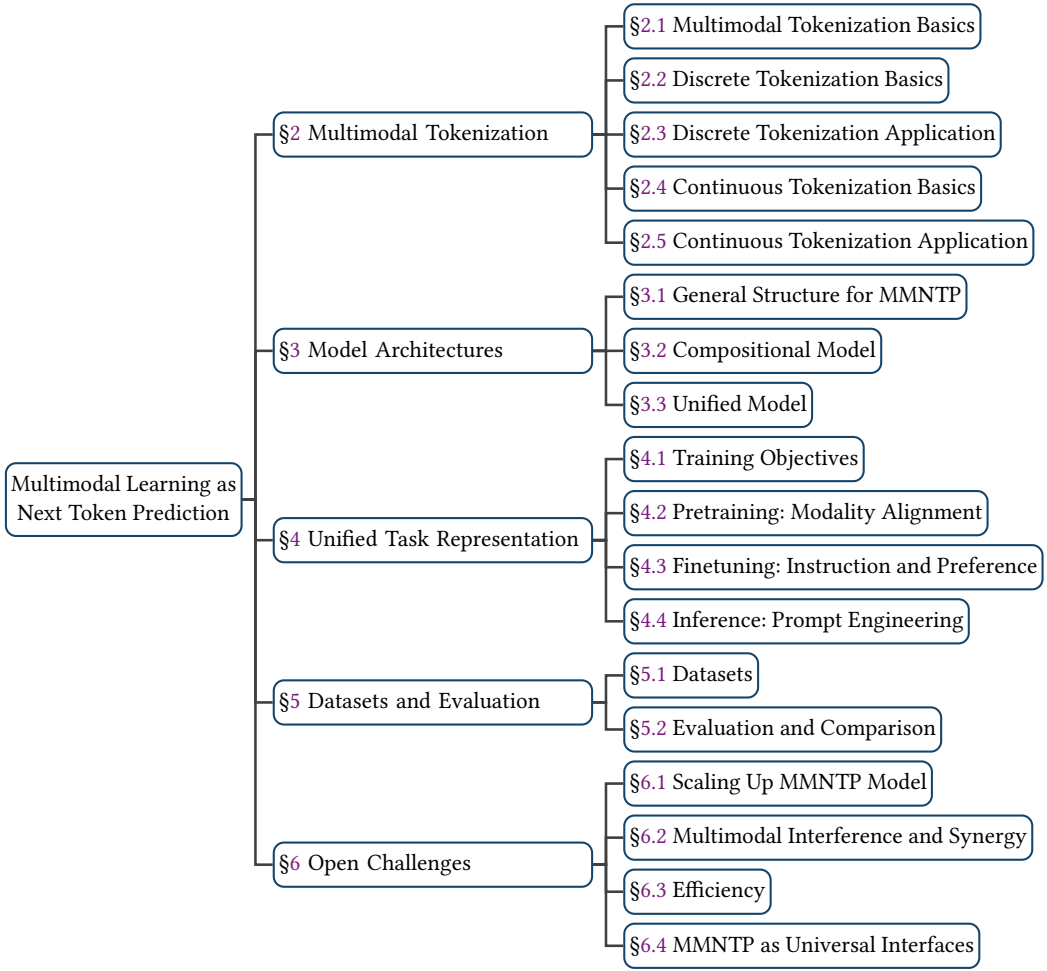
\begin{figure}[t!]
    \centering
    \resizebox{1\textwidth}{!}{
        \begin{forest}
            forked edges,
            for tree={
                grow=east,
                reversed=true,
                anchor=base west,
                parent anchor=east,
                child anchor=west,
                base=left,
                font=\small,
                rectangle,
                draw=hidden-draw,
                rounded corners,
                align=center,
                minimum width=2em,
                edge+={darkgray, line width=1pt},
                s sep=3pt,
                inner xsep=2pt,
                inner ysep=3pt,
                line width=0.8pt,
                ver/.style={rotate=90, child anchor=north, parent anchor=south, anchor=center},
            },
            where level=1{text width=12em,font=\small}{}, 
            [
                Multimodal Learning as \\Next Token Prediction
                [
                    \S \ref{sec:tokenization} Multimodal Tokenization 
                    [
                        \S \ref{sec: good tokenizer} Multimodal Tokenization Basics 
                    ]
                    [
                        \S \ref{sec:discrete tokens} Discrete Tokenization Basics 
                    ]
                    [
                        \S \ref{sec:vq app} Discrete Tokenization Application 
                    ]
                    [
                        \S \ref{sec:continuous tokens} Continuous Tokenization Basics 
                    ]
                    [
                        \S \ref{sec:continous app} Continuous Tokenization Application 
                    ]
                ]
                [
                    \S \ref{sec:model} Model Architectures 
                    [
                       \S \ref{sec: general structure} General Structure for MMNTP 
                    ]
                    [
                        \S \ref{sec:comp model} Compositional Model 
                    ]
                    [
                        \S \ref{sec:unified model} Unified Model 
                    ]
                ]
                [
                    \S \ref{sec:training} Unified Task Representation
                    [
                        \S \ref{sec:training_objectives} Training Objectives 
                    ]
                    [
                        \S \ref{subsub-ssl} Pretraining: Modality Alignment
                    ]
                    [
                        \S \ref{subsub-sl} Finetuning: Instruction and Preference 
                    ]
                    [
                        \S \ref{sub:inference} Inference: Prompt Engineering 
                    ]
                ]
                [
                     \S \ref{sec:datasets} Datasets and Evaluation
                     [
                        \S \ref{sec: training_dataset} Datasets 
                    ]
                    [
                        \S \ref{sec: evaluation_dataset} Evaluation and Comparison
                    ]
                ]
                [
                    \S \ref{sec:challenges} Open Challenges
                    [
                        \S \ref{sec:challange_scaling_law} Scaling Up MMNTP Model 
                    ]
                    [
                         \S \ref{sec:challange_opti} Multimodal Interference and Synergy
                    ]
                    [
                        \S \ref{sec:challange_eff} Efficiency 
                    ]
                    [
                        \S \ref{sec:challange_exte} MMNTP as Universal Interfaces
                    ]
                ]
            ]
        \end{forest}}
    \caption{Structure of the survey for Multimodal Learning with Next Token Prediction (MMNTP). }
    \label{taxonomy}
\end{figure}

\subsection{Overall Structure of the Survey}

The structure of the survey is shown in Figure~\ref{taxonomy}. Section~\ref{sec:tokenization} focuses on Multimodal Tokenization, and highlights the importance of tokenization as the bridge between raw multimodal data and their representations, distinguishing between discrete tokens that use Vector Quantization and continuous tokens. Section~\ref{sec:model} delves into the Multimodal Backbone Model for NTP, indicating that an auto-regressive model, often resembling a large language model, is employed to capture multimodal tokens, utilizing distinct attention masks for different modalities to account for their specific features. Section~\ref{sec:training} covers Training with Unified Multimodal Task Representation, explaining the training objectives varying from discrete to continuous token prediction, enabling multimodal output through VQ decoders or directly generating conditions for models like diffusion or VAE. The section also covers prompt engineering techniques such as In-Context Learning and Chain-of-Thought reasoning of MMNTP models adopted from  LLM research. Section~\ref{sec:datasets} introduces datasets and evaluation metrics, noting the superior performance of NTP models over non-NTP models in both understanding and generation tasks. Lastly, Section~\ref{sec:challenges} outlines unsolved challenges in MMNTP research, such as scaling up MMNTP, emergent abilities, modality-specific biases, modalities interference, and MMNTP as universal interfaces, and discusses approaches to mitigate these challenges.
Table~\ref{table:key_tables} outlines key tables and figures in our survey. 

\subsection{Related Work}

Several recent works have reviewed Large Multimodal Models (LMMs) in multimodal learning. For instance, ~\citet{yin2024surveymultimodallargelanguage} delve into the understanding capabilities of early vision-language models. Similarly, \citet{awais2023foundationalmodelsdefiningnew}, ~\citet{bordes2024introductionvisionlanguagemodeling}, ~\citet{ghosh2024exploringfrontiervisionlanguagemodels}, ~\citet{caffagni2024revolutionmultimodallargelanguage}, and ~\citet{zhang2024mmllmsrecentadvancesmultimodal} take a step forward and explore recent progress in multimodal learning with a focus on model architecture, training strategies, datasets, evaluation metrics and more. 
In addition, several surveys have reviewed multimodal learning in vision-language tasks, including pre-training~\citep{caffagni2024revolutionmultimodallargelanguage}, transfer learning~\citep{zhang2024visionlanguagemodelsvisiontasks}, reasoning~\citep{wang2024exploringreasoningabilitiesmultimodal}, and reinforcement learning from human feedback (RLHF)~\citep{zhang2024visionlanguagemodelsvisiontasks}.
Beyond the discussions on the general revolution of the LMMs, specialized surveys have investigated the application of LMMs in domains such as multimodal agents~\citep{xie2024largemultimodalagentssurvey,li2023multimodalfoundationmodelsspecialists} and autonomous driving~\citep{cui2023surveymultimodallargelanguage}. 
Recent surveys have also tackled key issues in multimodal learning, such as hallucinations in LMMs~\citep{liu2024surveyhallucinationlargevisionlanguage,rohleder2024variationalapproachhotspots} and efficiency of LMMs~\citep{jin2024efficientmultimodallargelanguage,xu2024surveyresourceefficientllmmultimodal}.

Diverging from prior work that primarily focused on the understanding abilities of multimodal LLMs,  our survey adopts a systematic perspective by integrating both understanding and generation in multimodal learning through the paradigm of next-token prediction. To the best of our knowledge, this is the first survey that reviews LMMs from the perspective of next token prediction, aiming to aid researchers in their exploration of multimodal intelligence.

In summary, in this survey, we aim to provide a holistic review on current multimodal models that rely on next token prediction.  An associated GitHub link collecting the latest papers is at \href{https://github.com/LMM101/Awesome-Multimodal-Next-Token-Prediction}{https://github.com/LMM101/Awesome-Multimodal-Next-Token-Prediction}.

\section{Multimodal Tokenization}
\label{sec:tokenization}
Tokenization is the first and a fundamental step for multimodal sequential modeling under the next token prediction framework. It decomposes information from various sources, such as images, videos, and audio clips, into a sequence of minimal, manageable units known as tokens for the NTP model to learn. Table~\ref{table:multimodal_tokenization} provides an overview of the tokenizers used across various modalities in recent research.

\begin{figure}[h]
    \centering
    \includegraphics[width=\linewidth]{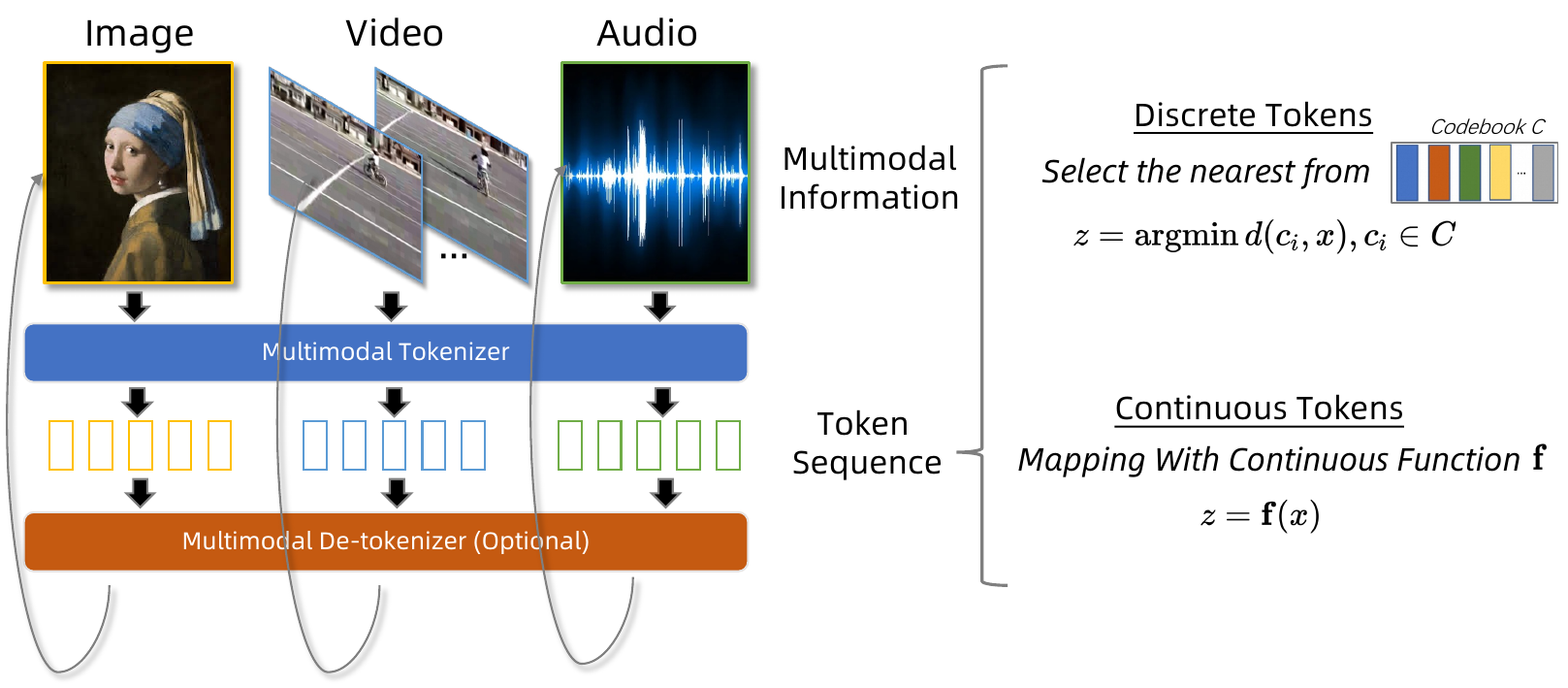}
    \caption{Illustrations of multimodal tokenization.}
    \label{fig:tokenizations}
\end{figure}

\begin{table*}[t!]
\centering

\caption{Summary of tokenizers for different modalities. Training method includes AE (Auto-Encoding), DAE (Denoising Auto-Encoding), SP (Supervised Pretraining) and CL (Contrastive Learning), $\dagger$ means training with auxiliary losses.}
\label{table:multimodal_tokenization}

\resizebox{1\textwidth}{!}{
\begin{tabular}{cc|cccccc}
\toprule
   Tokenizer & Year & Modality & Tokenizer Type & Backbone Structure & Training Method & Quantization Method & Used for Generation \\ \midrule
VQVAE~\citep{rolfe2017discrete} & 2017 & Image, Video, Audio & Discrete & CNN & AE & Vanilla VQ & \ding{51} \\
3D ConvNets~\citep{diba2017temporal} & 2017 & Video & Continuous & 3D-CNN & SP & \ding{55} & \ding{55} \\
VQVAE-2~\citep{Razavi2019GeneratingDH} & 2019 & Image & Discrete & CNN & AE & Multi-scale VQ & \ding{51} \\
vq-wav2vec~\citep{baevski2019vq} & 2019 & Audio & Discrete & Transformer & CL & \ding{55} & \ding{55} \\
wav2vec 2.0~\citep{baevski2020wav2vec} & 2020 & Audio & Continuous & Transformer & CL & \ding{55} & \ding{55} \\
VQGAN~\citep{Esser2020TamingTF} & 2020 & Image & Discrete & CNN & AE & Vanilla VQ $^\dagger$ & \ding{51} \\
SoundStream~\citep{zeghidour2021soundstream} & 2021 & Audio & Discrete & CNN & AE & Vanilla VQ & \ding{51} \\
WavLM~\citep{chen2022wavlm} & 2021 & Audio & Continuous & Transformer & DAE & \ding{55} & \ding{55} \\
HuBERT~\citep{hsu2021hubert} & 2021 & Audio & Continuous & Transformer & DAE & \ding{55} & \ding{55} \\
NFNet~\citep{nfnet} & 2021 & Image & Continuous & CNN & SP & \ding{55} & \ding{55} \\
BEiT~\citep{beit} & 2021 & Image & Continuous & Transformer & DAE & Vanilla VQ & \ding{55} \\
BEiTv2~\citep{peng2022beitv2maskedimage} & 2022 & Image & Continuous & Transformer & DAE & Vanilla VQ & \ding{55} \\
MaskDistill~\citep{peng2023a} & 2023 & Image & Continuous & Transformer & DAE & \ding{55} & \ding{55} \\
CLIP~\citep{radford2021clip} & 2021 & Image & Continuous & ViT & CL & \ding{55} & \ding{55} \\
ViT-VQGAN~\citep{yu2022vectorquantized} & 2021 & Image & Discrete & Transformer & AE & Vanilla VQ$^\dagger$ & \ding{51} \\
ViViT~\citep{arnab2021vivit} & 2021 & Video & Continuous & ViT & SP & \ding{55} & \ding{55} \\
data2vec~\citep{baevski2022data2vec} & 2022 & Audio, Image & Continuous & Transformer & DAE & \ding{55} & \ding{55} \\
Whisper~\citep{radford2023robust} & 2022 & Audio & Continuous & Transformer & SP & \ding{55} & \ding{55} \\
CLAP~\citep{clap} & 2022 & Audio & Continuous & Transformer & CL & \ding{55} & \ding{55} \\
Encodec~\citep{encodec} & 2022 & Audio & Discrete & CNN, LSTM & AE & Residual VQ & \ding{51} \\
FlexiViT~\citep{flexivit} & 2022 & Image & Continuous & Transformer & SP & \ding{55} & \ding{55} \\
Pix2Struct~\citep{lee2022pix2struct} & 2022 & Image & Continuous & Transformer & SP & \ding{55} & \ding{55} \\
RQVAE~\citep{lee2022RQVAE} & 2022 & Image & Discrete & CNN & AE & Residual VQ$^\dagger$ & \ding{51} \\
MAGViT~\citep{magvit} & 2022 & Video & Discrete & 3D-CNN & AE & Vanilla VQ$^\dagger$ & \ding{51} \\
C-ViViT~\citep{villegas2022phenaki} & 2022 & Video & Discrete & Transformer & AE & Vanilla VQ & \ding{51} \\
CoCa~\citep{coca} & 2022 & Image & Continuous & Transformer & CL+SP & \ding{55} & \ding{55} \\
EVA-CLIP~\citep{eva-clip} & 2023 & Image & Continuous & Transformer & CL & \ding{55} & \ding{55} \\
SAM-CLIP~\citep{samclip} & 2023 & Image & Continuous & Transformer & CL+SP & \ding{55} & \ding{55} \\
NaViT~\citep{navit} & 2023 & Image & Continuous & Transformer & CL+SP & \ding{55} & \ding{55} \\
InternVIT~\citep{InternVL} & 2023 & Image & Continuous & Transformer & CL+SP & \ding{55} & \ding{55} \\
SEED-Tokenizer~\citep{seed-tokenizer} & 2023 & Image & Discrete & ViT & AE+CL & Vanilla VQ & \ding{51} \\
USM~\citep{usm} & 2023 & Audio & Continuous & Conformer & DAE & \ding{55} & \ding{55} \\
DAC~\citep{kumar2024high} & 2023 & Audio & Discrete & CNN & AE & Improved Residual VQ & \ding{51} \\
LMCodec~\citep{jenrungrot2023lmcodec} & 2023 & Audio & Discrete & CNN & AE & Residual VQ$^\dagger$ & \ding{51} \\
HiFiCodec~\citep{yang2023hifi} & 2023 & Audio & Discrete & CNN, LSTM & AE & Group VQ & \ding{51} \\
SpeechTokenizer~\citep{zhang2023speechtokenizer} & 2023 & Audio & Discrete & CNN, LSTM & AE & Residual VQ$^\dagger$ & \ding{51} \\
MAGViT-v2~\citep{magvit2} & 2024 & Image, Video & Discrete & 3D-CNN & AE & LFQ$^\dagger$ & \ding{51} \\
LaViT~\citep{lavit} & 2024 & Image & Discrete & ViT+U-Net & AE & Vanilla VQ$^\dagger$ & \ding{51} \\
Video-LaVit~\citep{video-lavit} & 2024 & Video & Discrete & ViT+U-Net & AE & Vanilla VQ, MPEG-4 & \ding{51} \\
SPAE~\citep{spae} & 2024 & Image & Discrete & CNN & AE & Vanilla VQ & \ding{51} \\
FACodec~\citep{ju2024naturalspeech} & 2024 & Audio & Discrete & Transformer & AE & Group VQ & \ding{51} \\
SemantiCodec~\citep{liu2024semanticodec} & 2024 & Audio & Discrete & AudioMAE (ViT) & AE & Vanilla VQ & \ding{51} \\
WavTokenizer~\citep{ji2024wavtokenizer} & 2024 & Audio & Discrete & CNN, LSTM & AE & Vanilla VQ & \ding{51} \\
Mimi~\citep{defossez2024moshi} & 2024 & Audio & Discrete & Transformer & AE & Residual VQ$^\dagger$ & \ding{51} \\
VAR~\citep{VAR} & 2024 & Image & Discrete & CNN & AE & Multi-scale VQ$^\dagger$ & \ding{51} \\
QwenVL2-ViT~\citep{Qwen2vl} & 2024 & Image, Video & Continuous & Transformer & CL & \ding{55} & \ding{55} 
\\ \bottomrule
\end{tabular}}
\end{table*}

Despite being derived from various modalities, these tokenization methods can all be categorized into two prototypes: \textbf{discrete tokenization}  and \textbf{continuous tokenization}. In this section, we will initially introduce the general definition and basics techniques of training multimodal tokenizers (\S~\ref{sec: good tokenizer}), then the fundamentals and applications of discrete tokens (\S~\ref{sec:discrete tokens},~\ref{sec:vq app}) and continuous tokens (\S~\ref{sec:continuous tokens},~\ref{sec:continous app}) in NTP framework. 


\subsection{Tokenization of Different Modalities}
\label{sec: good tokenizer}


We first define the tokenization process as a function $f$ that maps a sample $x$ from the raw multimodal space $X$ to a representation $z$ in the tokenizer's output representation space $Z_f$.
\begin{equation}
    f(x) = z,
    \label{eq:tokenization}
\end{equation}
where $x\in X$ and $z\in Z_f$.

\subsubsection{Tokenizer Type}

As illustrated in Fig.~\ref{fig:tokenizations}, tokenizers for multimodal information can be categorized into two types: discrete and continuous. This classification is based on how tokens are derived from the original data. Both tokenization methods encode the original information into a latent representation space, but they differ in their approach. 

Discrete tokenization performs quantization on the latent space, utilizing a fixed-size, discrete space similar to the vocabulary of language models. In contrast, continuous tokenization does not involve quantization, resulting in a much larger representation space. 

\paragraph{Discrete}

In Equation~\ref{eq:tokenization}, a discrete token implies that the representation space $Z_f$ comprises a finite number of discrete symbols. The output space is called the codebook $C = \{c_1, c_2, ..., c_N\}$, where $c_i \in \mathbb{R}^0$, and each representation $z$ is composed of codes from this codebook, i.e., $z = \{z_1, z_2, ..., z_n\}$ with $z_i \in C$. Language tokens are inherently discrete because they originate from a finite vocabulary. Each word or subword unit is mapped to a unique token from this predefined set. In contrast, modalities such as audio and images exist in continuous, high-dimensional spaces. To process these modalities within the same framework (i.e., NTP) as for discrete language tokens, they need to be transformed into a discrete representation.


Quantization is a process that maps values from a continuous space to a discrete space, typically resulting in a much smaller representation space. It is a default operation when a discrete representation is desired for tokenizing multimodal information. Quantization is often combined with auto-encoder techniques to reduce the size of the latent space. Typical examples include VQ-series tokenizers such as VQVAE~\cite{vq} and VQGAN~\cite{Esser2020TamingTF}, which inherently feature discrete representations. Details of the quantization process are introduced in Section~\ref{sec:discrete tokens}.

\paragraph{Continuous}

In contrast to discrete tokenization, continuous tokenization represents data using a continuous space where tokens are derived directly from the data's inherent properties without enforcing quantization into a predefined codebook. In this approach, the representation space \( Z_f \) is not limited to a finite set of predetermined codes; rather, it preserves the continuous nature of the data. Each token \( z \) is sampled from a continuous distribution, allowing for a more nuanced and flexible representation that can capture the subtleties of the input data. Continuous tokenization is particularly advantageous for modalities that naturally exist in a continuous form and require a rich representational capacity to capture their complex patterns. For instance, in audio and visual data, continuous representations can effectively retain fine-grained temporal and spatial information that might be lost during discrete tokenization.

\begin{figure}[h]
    \centering    
    \includegraphics[width=1\linewidth]{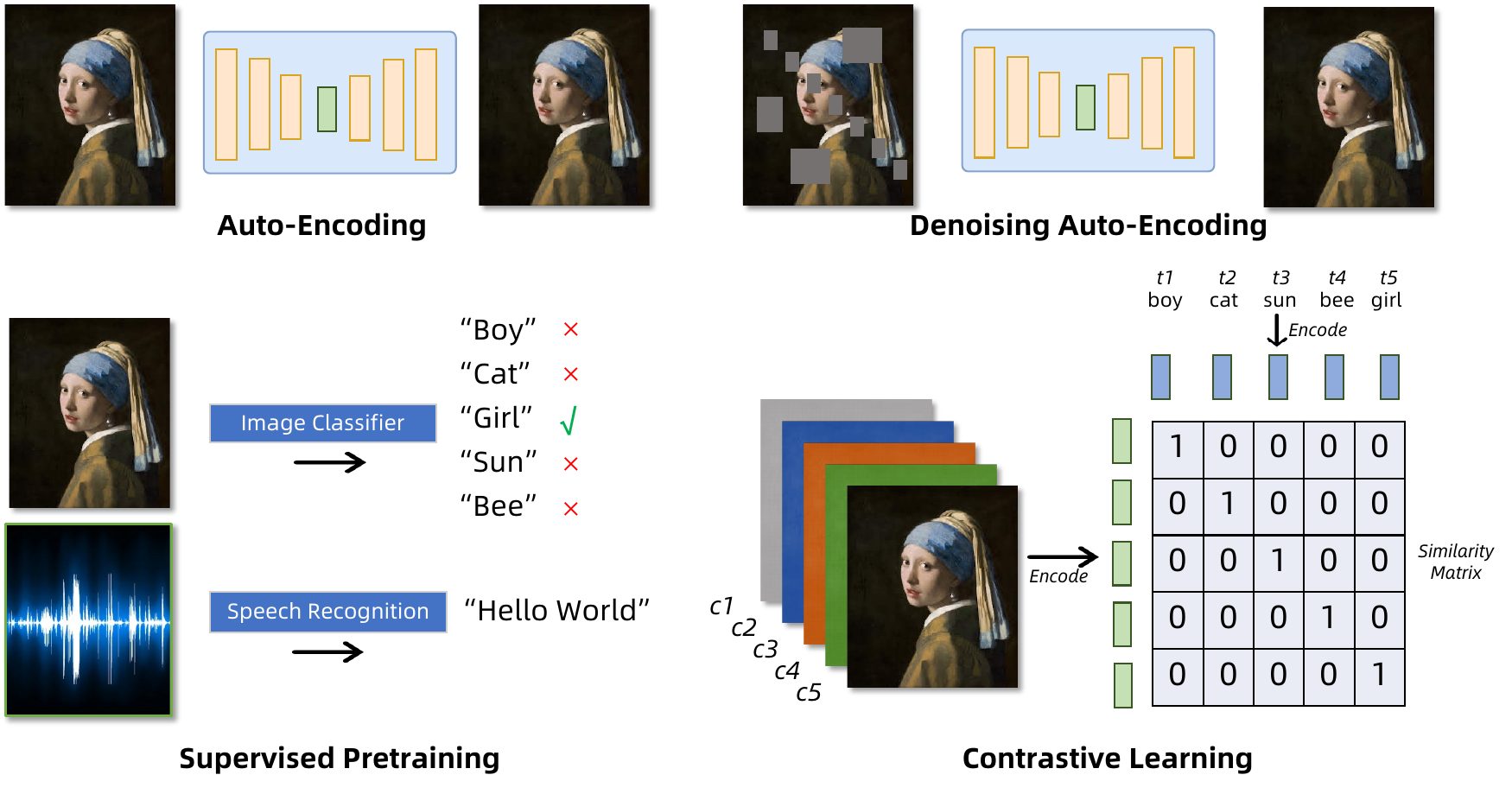}
    \caption{Illustrations of the training method of tokenizers. }
    \label{fig:keyfeatures}
\end{figure}

\subsubsection{Features of Tokenizers}

Before diving into different tokenization techniques, we summarize the basic two features (Representation and Reconstruction) that an ideal multimodal tokenizer should possess to achieve better understanding and generation capabilities in the NTP framework.

\paragraph{Representation Ability:} Effective representation encodes semantically relevant information into the latent space $Z$ while removing redundant information. This is crucial for various downstream tasks that learn a conditional probability $P(Y|X)$ over the label space $Y$, conditioned on the multimodal input space $X$, by replacing it with $P(Y|Z)$. Prominent tokenizers known for better representation include language-guided contrastive learning methods such as CLIP~\cite{radford2021clip} and fully self-supervised methods like DINO~\cite{caron2021emerging}.

\paragraph{Reconstruction Ability:} For generating multimodal information, it is expected that the tokenization function $f$ is invertible or nearly invertible, meaning there is a detokenization function $g$ that can recover the original input from the representation space, satisfying $g(f(x)) = x$ or $g(f(x)) \approx x$. Notable works that excel in reconstruction include Auto-Encoder (AE) series models such as Variational Auto-Encoder~\cite{kingma2022autoencoding} (VAE) and VQVAE~\cite{vq}.

It is important to note that these abilities are not mutually exclusive; their balance depends on the training techniques used.

\subsubsection{Training Methods for Tokenizers}

The training methodologies for tokenizers can be categorized into four groups, based on their respective training objectives: Auto-Encoding, Denoising Auto-Encoding, Supervised Training, and Contrastive Learning, as depicted in Figure~\ref{fig:keyfeatures}. Herein, we provide a summary of the core concepts associated with various tokenizers.


\paragraph{Auto-Encoding}

Auto-Encoder (AE) is a type of artificial neural network designed to learn efficient data representations. It consists of two main components: an encoder, which maps input data to a latent space with reduced dimensions, and a decoder, which reconstructs the input data from this latent representation. The training goal for an Auto-Encoder is to minimize the reconstruction error, ensuring the decoded output closely resembles the original input. Variants like Variational Auto-Encoders~\cite{kingma2022autoencoding} (VAEs) use probabilistic approaches to generate more robust and informative embeddings. In multimodal generation models, tokenizers trained with auto-encoder methodologies are used to restore the multimodal input from the latent representation. A special case is diffusion models~\cite{dieleman2022diffusion}, which can also be viewed as an Auto-Encoder, enabling generation in a non-autoregressive manner~\cite{li2024autoregressive-without}. Discrete tokens are typically generated by quantizing~\citep{rolfe2017discrete} the continuous data representation within the latent space of auto-encoders.

\paragraph{Denoising Auto-Encoding}

A Denoising Auto-Encoder (DAE) builds on the basic auto-encoder concept by introducing noise into the input data and training the model to reconstruct the original, noise-free version. This approach encourages the model to learn robust features capable of handling data corruption, thereby improving its generalization capabilities. In transformer-based models, a common technique known as Masked Language Modeling~\citep{devlin2019bert} involves masking parts of the input tokens and training the model to predict them, which can be viewed as a special type of denoising auto-encoder. This method has become mainstream across various modalities, popularized in language by BERT~\citep{devlin2019bert}, in vision by BEiT~\citep{beit} and MAE~\citep{he2021maskedautoencodersscalablevision}, and in audio by HuBERT~\citep{hsu2021hubert}.

\paragraph{Supervised Pretraining}

Some tokenizers are pretrained on specific tasks using supervised learning, aiming to acquire task-specific representations through labeled datasets. These models are initially trained on large-scale datasets to capture specific features of the input data. In the vision modality, supervised tasks include semantic segmentation, object detection, and depth estimation. Models trained for these tasks, such as SAM~\citep{sam,samclip}, ViTDet~\citep{vitdet}, and MiDaS~\citep{midas}, are later used in LMMs as tokenizers, like in DeepSeek-VL~\citep{DeepSeek-VL} and Cambrain-1~\citep{Cambrian-1}, to extract diverse visual features from input data. In the audio modality, Whisper~\citep{radford2023robust} is trained with 680,000 hours of labeled audio data in a weakly supervised manner. Thanks to its robust and powerful speech feature extraction capabilities, Whisper is widely used in Speech LLMs~\cite{tang2023salmonn, chu2023qwen, hu2024wavllm} for extracting speech embeddings.

\paragraph{Contrastive Learning}

Contrastive Learning is a self-supervised learning method that focuses on learning representations by distinguishing between positive and negative pairs. The core idea is to bring similar (positive) examples closer together in the representation space while pushing dissimilar (negative) examples further apart. The items in each pair can belong to the same or different modalities. For example, DINO~\citep{caron2021emerging} uses image-image pairs to enhance vision representation, while CLIP~\citep{radford2021clip} employs text-image pairs to improve language alignment within vision representation.

Currently, LMMs that only feature multimodal understanding capabilities, such as InstructBLIP~\cite{dai2023instructblip} and LLaVA~\cite{liu2023llava}, opt for tokenizers with superior representation abilities like CLIP~\citep{radford2021clip}, as they do not require reconstruction of the multimodal information. Conversely, LMMs supporting multimodal generation capabilities tend to choose VQVAE as the tokenizer, exemplified by models like Unified-IO~\cite{lu2022unifiedio}, Chameleon~\cite{chameleonteam2024chameleon}, Emu3~\citep{Emu3}, among others~\citep{wang2024mio, seedllama, wang2022ofa}.

\subsection{Discrete Tokenization Basics}
\label{sec:discrete tokens}
Unlike the language modality, which inherently comprises discrete symbols (e.g., tokens or words), most other modalities naturally exist in a continuous space. To bridge the gap, the core technique is \textbf{Vector Quantization (VQ)}, which aims to map the original continuous information into a compressed, finite representation space, i.e. discrete tokens. The discrete tokens can have 2-dimensional or 3-dimensional structures for images and videos. These tokens are initially linearized based on a specific order, such as left to right and top to bottom, transforming them into a 1-dimensional sequence. This linearization allows for effective modeling usingthe next token prediction objective.

In this section, we will first elaborate on modern vector quantization techniques widely used as multimodal tokenizers, such as VQVAE (\S~\ref{sec:vq}) and its variants. Following that, we will introduce the specific optimizations of discrete tokenization in different modalities (\S~\ref{sec:vq app}).

\subsubsection{Vector Quantization Methods} 
\label{sec:vq}

\begin{figure}[t]
    \centering
    \includegraphics[width=\linewidth]{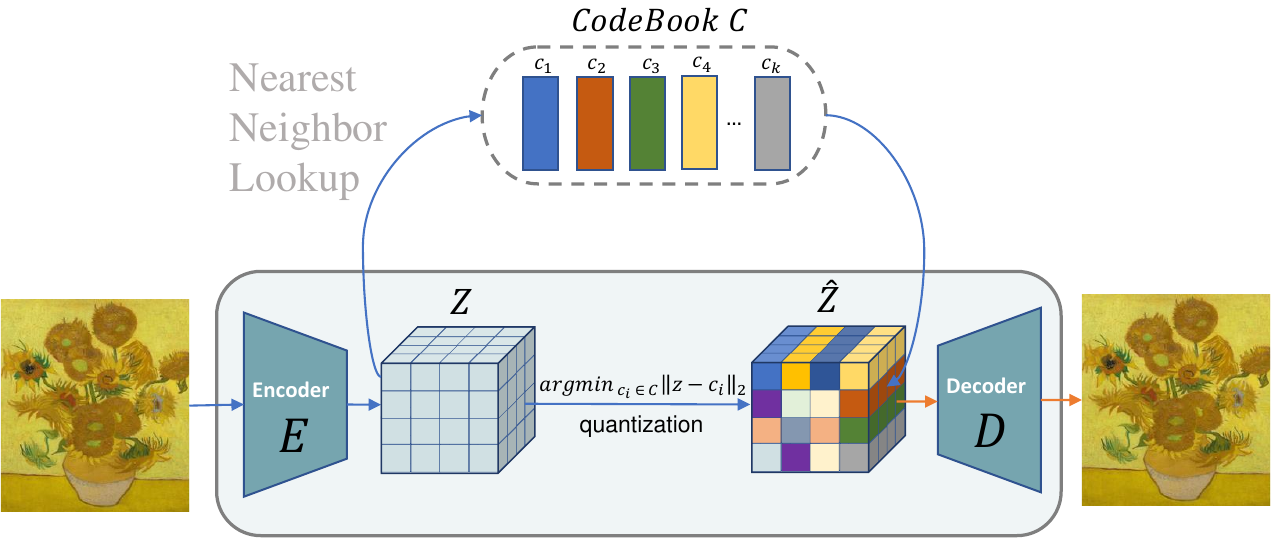}
    \caption{Illustration of Vector Quantization. Blue lines denote the encoding and quantization process, while the orange lines denote the reconstruction process. The encoder transforms the input image into a latent representation, which is quantized by mapping each vector in \( Z \) to its nearest codebook entry in \( C \). The quantized representation \( \hat{Z} \) is then passed through the decoder to reconstruct the image.}
    \label{fig:vqvae}
\end{figure}

The origins of VQ method trace back to the 1950s at Bell Laboratories, where researchers endeavored to optimize signal transmission through the development of suitable discretization procedures~\cite{Pags2015IntroductionTV}. In essence, quantization is the process of mapping an infinite set of continuous values to a smaller, discrete set of finite values. The primary objective of vector quantization is to reconstruct all the information in the original data as accurately as possible with a finite set of vectors, which is also called the \emph{codebook}. 

\paragraph{Vanilla VQ}

The original VQVAE proposed by \citet{Oord2017NeuralDR} is a milestone of many successive vector quantization methods. As shown in Figure~\ref{fig:vqvae}, a VQVAE consists of three main components: the encoder, the quantizer, and the decoder. The encoder comprises the input data to a compact latent space, the quantizer select the nearest code vectors from the finite codebook to approximate the continuous latents, the decoder reconstruct the input data using the discrete codes. When training the VQVAE, three main loss components are crucial: reconstruction loss, codebook loss, and commitment loss~\citep{Oord2017NeuralDR}. The reconstruction loss, often implemented as mean squared error or binary cross-entropy, ensures accurate data reconstruction by minimizing differences between input and output. Codebook loss, or vector quantization loss, enables effective encoding by aligning encoder outputs with nearest codebook entries, ensuring discrete latent variables. Meanwhile, commitment loss acts as a regularizer, encouraging encoder outputs to stay close to codebook entries to maintain stable learning, preventing erratic mapping. As gradient can not pass the quantization operator (finding the nearest code), the straight-through estimator~\cite{bengio2013estimatingpropagatinggradientsstochastic} is adopted to let the gradient flow normally.




Recent advancements in vector quantization methods have focused on achieving better image reconstruction and enhancing generative capabilities. To improve reconstruction quality, both architectural innovations and codebook designs have been proposed. Transformer-based frameworks, such as ViT-VQGAN~\citep{yu2022vectorquantized}, Swin-MAE~\citep{xu2023swin}, Swin-Unet~\citep{cao2021swinunet}, and Efficient-VQGAN~\citep{cao2023efficientvqgan}, replace traditional CNN encoders and decoders with more robust modules like ViT~\citep{vit} and Swin-Transformer~\citep{liu2021swinTransformer,liu2022swinV2}, leading to better feature representations and reconstruction fidelity. Additionally, several methods such as LFQ~\citep{magvit2} and FSQ~\citep{FSQ} are proposed to address the significant challenge of codebook collapse during \textbf{codebook learning}, where a large portion of code embeddings are not used when enlarging the codebook size, causing a redundancy in the codebook and limiting the
expressive power of the generative model~\citep{baykal2024edvaemitigatingcodebookcollapse}. For improved generative performance and efficiency, several approaches have been introduced. \citet{tian2024VAR} propose Visual Autoregressive modeling, which facilitates image generation through "next-scale prediction", moving away from the traditional raster-scan "next-token prediction" used in standard VQVAE-based models. RQ-Transformer~\citep{lee2022RQVAE} employs residual quantization (RQ) to precisely approximate feature maps and reduce spatial resolution. RQ helps the RQ-Transformer to significantly reduce computational costs and effectively learn long-range interactions in inputs. RAR~\citep{RAR} introduces a randomness annealing strategy with a permuted objective, enhancing the model's ability to learn bidirectional contexts while retaining the autoregressive framework. TiTok~\citep{yu2024imageworth32tokens} tokenizes images into 1D latent sequences, providing a more compact latent representation that is substantially more efficient and effective than conventional techniques. It greatly reduces the number of tokens required to encode an image compared to previous methods~\citep{cao2023efficientvqgan,yu2022vectorquantized}.

\paragraph{VQ with Auxiliary Losses}
The primary goal of the vanilla VQVAE is to accurately reconstruct input data by minimizing the mean squared error loss. However, this auto-encoding objective doesn't always align with human perception of the quality of reconstructed data. For example, in the visual modality, the vanilla MSE loss often results in images with blurred details, particularly in human faces~\citep{larsen2016autoencodingpixelsusinglearned}. To address this issue, several approaches introduce higher-level training objectives aimed at improving the overall quality of the output data. In the realm of vision, perceptual loss~\citep{johnson2016perceptuallossesrealtimestyle} is widely used to enhance the quality of reconstructed images by leveraging a pre-trained CNN. VQGAN~\citep{cao2023efficientvqgan} incorporates a discriminator network to enhance image fidelity by adding an adversarial training objective. The role of the discriminator is to discern between the reconstructed and original images, while the VQ-VAE is optimized to deceive the discriminator, thereby improving the quality of the reconstructed images. In the audio modality, it is essential to decouple the audio into its acoustic and semantic components to achieve both powerful audio reconstruction quality and LLM modeling. SpeechTokenizer~\citep{zhang2023speechtokenizer} and Mimi~\citep{defossez2024moshi} introduce the loss of semantic distillation at the first layer of Residual VQ, using self-supervised models, such as HuBERT~\citep{hsu2021hubert} and WavLM~\citep{chen2022wavlm}.

\paragraph{Residual Vector Quantization} Residual vector quantization (RVQ) has been used for image~\citep{Lee_Kim_Kim_Cho_Han_2022} and audio~\citep{Zeghidour_Luebs_Omran_Skoglund_Tagliasacchi_2022} generation, where quantized codes are refined by storing additional quantized residuals.
\citet{lee2022RQVAE} propose the RQVAE that also introduces a residual quantization to recursively quantize the feature map in a coarse-to-fine manner, employing a fixed-size codebook to maintain both precision and code diversity.

\paragraph{Product Quantization}~\citet{PO-VAE} propose product quantization (PQ), to factor the codebook into a product of smaller codebooks, allowing for high-quality quantizers without the requirement of intractably large codebooks.

\paragraph{Multi-scale Quantization} ~\citet{tian2024VAR} introduce the Visual Autoregressive modeling (VAR), which develops a multi-scale quantization autoencoder that encodes images into $K$ multi-scale discrete token maps using a shared codebook. It aids the model in generating images through "next-scale prediction," instead of the raster-scan "next-token prediction" typically used in standard VQVAE-based models. The multi-scale quantization enables the model to learn visual distributions and demonstrates strong generalization capabilities.

\paragraph{Finite Scalar Quantization}
To generate concise and expressive tokens using a larger token vocabulary and avoid codebook collapse, \citet{FSQ} propose finite scalar quantization (FSQ). FSQ projects the VAE representation down to a few dimensions that can be quantized into fixed values, creating an implicit codebook.

\paragraph{Look-up Free Quantization}
LFQ~\citep{yu2023language} reduces the embedding dimension of the codebook to zero, effectively replacing the codebook with an integer set. It allows VQVAE to improve the quality of image reconstruction and generation by vastly increasing the vocabulary size by magnitudes. For example, the rFID on Imagenet decreases from 2.5 to 1.4 when the LFQ vocabulary size increases from $2^10$ to $2^16$ on ImageNet dataset.

\paragraph{Embedding-Free Quantization}
Maskbit~\cite{maskbit} explores an embedding-free tokenization approach that utilizes binary quantization. It projects latent embeddings into K dimensions and then quantizes them based on their sign values to produce bit token representations. The generated bit tokens exhibit highly structured semantic representations, which are crucial for generation tasks.

\paragraph{Group Vector Quantization} Unlike RVQ which models the information residually, Group Vector Quantization models the information across different dimensions. In the audio domain, HiFi-Codec~\citep{yang2023hifi} proposes a group-residual vector
quantization technique to reduce the number of codebooks, while FACodec~\cite{ju2024naturalspeech} disentangles speech into prosody information, content information, and acoustic details using three-factorized vector quantizers.

\subsubsection{Evaluation of VQ Tokenizers}

When evaluating VQVAEs, two critical metrics are commonly considered: \textbf{reconstruction ability} and \textbf{generation ability}.

Reconstruction ability refers to how well the VQVAE can reproduce the original input data after encoding and decoding. This metric evaluates the fidelity of the model in terms of how accurately it can reconstruct the input data from its latent representations. L2 distance, Peak Signal-Noise Ratio (PSNR), and reconstruction Fréchet Inception Distance (rFID) are often applied to assess the reconstruction ability.

Generation ability assesses the model’s capacity to generate new, plausible samples from the learned distribution in the codebook space. This metric evaluates the creativity and diversity of the VQVAE in producing new data that is consistent with the training data distribution. To quantitatively evaluate generation ability, metrics such as the Inception Score (IS) and generation Fréchet Inception Distance (gFID)~\citep{heusel2018ganstrainedtimescaleupdate} are often used.

rFIDs are often computed between ImageNet validation images and their reconstructed images. gFIDs are usually computed against the training set with ADM's evaluation suite~\cite{dhariwal2021diffusionmodelsbeatgans}.

\subsection{Discrete Tokenization for Different Modalities}
\label{sec:vq app}
Generic quantization methods provide basic ways to convert continuous data into discrete tokens. However, there isn't a single quantizer that works well for all modalities because each modality has unique characteristics. Therefore, it is important to create specific tokenizers for each modality. This section will explain the unique features of different modalities and showcase some examples of tokenizers for images, audio, and video, among others.

\subsubsection{Image}
Images can be tokenized into discrete symbols with the previously introduced VQVAE structure. Compared to text tokens, images diverge in three fundamental aspects that significantly impact how they should be tokenized:

\begin{itemize}
    \item[1.] Rich Information Granularity: Unlike text, which primarily encapsulates high-level semantic meaning, images are contain with a myriad of perceptual details. These encompass low-level visual elements such as colors, shapes, and textures, alongside more abstract concepts like objects and actions.
    \item[2. ] Dense Information: Images inhabit a densely packed representational realm, where each pixel, across multiple dimensions including height, width, and color channels (RGB being a common example), carries information. This stands in stark contrast to the discreteness of text in nature, characterized by sequentially arranged words.
    \item[3. ] Two-Dimensional Spatial Structure: Images are inherently structured in two dimensions, spread across a grid defined by height and width. This 2D layout differs fundamentally from the straightforward, one-dimensional sequence that characterizes textual data, introducing unique complexities in their processing and analysis.
\end{itemize}



Given these differences, bridging the gap between text and image modalities in the training of LLMs based on discrete image tokens requires a robust image tokenizer, which must balance the fusion of sufficient alignment with LLM's language ability (referred to as ``representation''), the retention of rich original image information (referred to as ``reconstruction''), and the efficient use of tokens given the growing inference cost of transformer decoder (referred to as ``token efficiency''). These factors possess a trade-off~\citep{seedllama,seed-tokenizer, magvit2, sun2023generative}, making it crucial for the construction of an image tokenizer to maintain equilibrium among these factors.

In terms of better representation, models like ViT~\citep{vit} are commonly employed, often aligned with a text encoder through contrastive loss~\citep{radford2021clip, peng2022beit}, or aligned with text modalities through generative loss~\citep{coca}. Additionally, modules like Q-Former~\citep{li2023blip2} can also be used for image feature transformation~\citep{li2023blip2, seedllama}. Consequently, the resultant image features integrate higher-level semantics and gradually compress high-dimensional images into lower-dimensional representations aligned with text. While the initial arrangement of image patches follows a raster order, preserving intrinsic sequential relationships, this configuration lacks causal semantics, posing challenges for language modeling.

Regarding reconstruction ability, an image decoder is often layered atop the image encoder to reconstruct the original image from its representation, incorporating reconstruction loss into the training process~\citep{amused, seedllama, lavit, Esser2020TamingTF}. Training labels typically use the original images, but with advancements in diffusion models, more research is incorporating latents for diffusion models as reconstruction labels~\citep{lavit, seedllama}.

For token efficiency, modules like selectors or mergers for image tokens are utilized to truncate their length (i.e., the number of tokens per image). For instance, SEED-LLaMA~\citep{seedllama} compresses longer image features encoded by ViT into 32 continuous tokens using a Causal Q-Former and then discretizes them through quantization. LaViT~\citep{lavit} first predicts whether each patch token should be selected using a shared MLP, and then compresses the image length by employing selected patches as queries and unselected patches as keys and values in cross-attention blocks~\citep{seedllama}.

Beyond these aspects, some studies also focus on the unique properties of specific image types or tasks. For example, VQ-IMG aims to enhance the modeling capabilities of image tokenizers for faces~\citep{make-a-scene}, while LVM integrates tasks like segmentation and object detection during the training of models based on VQGAN to enrich the representation of image tokens~\citep{bai2023sequential}. StrokeNVWA introduces a VQ-Stroke method to discretize vector graphic images into stroke tokens~\citep{strokenvwa}.

\subsubsection{Audio} 

Raw audios are typically stored as 16-bit integer values with a sampling rate that exceeds tens of thousands values per second, which leads to extremely long sequences and renders next token prediction training more difficult. Versatile quantization methodologies have been investigated for audio tokenization.
Initially aimed at audio compression, these methodologies have more recently been developed to create compact semantic and acoustic representations in the context of NTP language modeling. 

As a traditional companding algorithm, $\mu$-law/A-law algorithm is commonly employed in speech generative models such as WaveNet~\citep{van2016wavenet}. While this algorithm projects each audio frame to an 8-bit value, it does not reduce the sampling rate, thereby preserving overlong sequences. Self-supervised learned models have shown exceptional performance in various speech-related tasks, sparking interest in clustering their speech representations for speech quantization. The vq-wav2vec~\citep{baevski2019vq} uses either a Gumbel-Softmax or online k-means clustering to quantize the SSL-learned dense representation. HuBERT~\citep{hsu2021hubert} is trained with a masked prediction task, whose targets are obtained through k-means clustering of learned features from earlier iterations. Utilizing quantized tokens learned with Self-Supervised Learning (SSL), GSLM~\cite{lakhotia2021gslm} and VQTTS~\citep{du2022vqtts} demonstrate faster speed in speech generation tasks compared with WaveNet. Because SSL tokens are extracted with highly abstracted semantics while discarding low-level acoustic information, the reconstruction quality is relatively low, and speaker identity is lost~\citep{borsos2023audiolm}. Neural codec models typically apply a VQ-VAE on the raw audios with residual vector quantization, exemplified by SoundStream~\citep{zeghidour2021soundstream} and EnCodec~\citep{encodec}. They are originally designed for audio compression, have the capability to encode waveforms into discrete codes and faithfully reconstruct them back into high-quality waveforms. Recently, they are widely used in audio generation models such as AudioLM~\citep{borsos2023audiolm}, VALL-E~\citep{wang2023neural} and their variants~\citep{han2024valler,song2024ellav, wang2023viola}, and reach new state-of-the-art performance on various tasks. Compared with traditional $\mu$-law/A-law algorithms, codec models can efficiently reduce the length of token sequences. It can also maintain multi-scale acoustic information indicating speaker identity compared with highly-abstracted SSL-learned discrete tokens such as HuBERT~\citep{hsu2021hubert} tokens. Additionally, the codec models are typically off-the-shelf and lightweight. 

Latest works have attempted to impose additional supervision on the discrete codes extracted by codec models. The objective is to enhance their ability to extract and encode higher-level semantic information, thereby improving language modeling. SpeechTokenizer~\citep{zhang2023speechtokenizer} is an RVQ-based codec model, where its first-layer codebook incorporates semantic information through the semantic distillation process, using HuBERT~\citep{hsu2021hubert} representations as the semantic teacher. Mimi, used by Moshi~\citep{défossez2024moshispeechtextfoundationmodel}, further improves upon this by replacing the semantic teacher from HuBERT with WavLM~\citep{chen2022wavlm}. Additionally, it isolates the first-layer codebook from the RVQ process to achieve better semantic and acoustic disentanglement. To enhance the compression rate, WavTokenizer~\citep{ji2024wavtokenizer} is capable of quantizing one-second audio into 75 or 40 tokens with a single quantizer.

\subsubsection{Video}

Compared to images, videos introduce an additional temporal dimension that must be considered during the tokenization process. A straightforward strategy is to utilize an image-based VQVAE model to tokenize the video frame-by-frame. This approach is employed by several multimodal foundation models, such as LVM~\cite{bai2023sequential}, LWM~\cite{liu2023world}, and Unified-IO series~\cite{lu2022unifiedio,lu2023unifiedio2}. However, a significant drawback of frame-by-frame tokenization is its inability to compress video data over time, resulting in a high degree of token redundancy across frames—particularly in long-form videos—thereby imposing substantial computational demands~\citep{song2024moviechatdensetokensparse}. Furthermore, using an image-based tokenizer fails to model temporal relationships between frames, leading to issues of temporal inconsistency.

To address token redundancy and enhance temporal modeling, several studies have proposed training a 3D tokenizer that compresses videos across spatial and temporal dimensions. For example, VideoGPT~\citep{yan2021videogpt} applies a 3D-CNN architecture in the encoder and decoder of the video tokenizer. C-ViViT~\cite{villegas2022phenaki} uses a transformer architecture to split videos into 3D cubes, which are then discretized into token IDs. 

There are two additional desirable features for a video tokenizer:
\textbf{(1) Joint Image-Video Tokenization}. The MAGVIT series~\cite{magvit2} enables tokenizing images and videos with a shared vocabulary. 
To achieve this, the number of frames in an input video, $T$, must satisfy $T=1+n \times F_T$, meaning the video comprises an initial frame followed by $n$ clips, each containing $F_T$ frames. 
When $n=0$, the video contains only the initial frame, thus simplifying the video to an image. Accordingly, both the initial frame and each subsequent clip are discretized into a $(1, H', W')$ token map, where $H'$ and $W'$ are the height and weight of the token map. 
\textbf{(2) Temporal Causality}. Compared to vanilla 3D architectures, using causal 3D architecture can ensure the tokenization and detokenization of each clip depend only on the preceding clips, facilitating autoregressive modeling along the temporal dimension. 

\subsubsection{More Modalities}

Modeling various information as discrete tokens has gone far beyond the traditional text, image, video and audio modalities. In the computer vision field, we can unify the output spaces of tasks like object detection, semantic segmentation, and depth mapping into images. These can then be tokenized into discrete image tokens, allowing us to train a single NTP model to handle all these tasks \cite{wang2022ofa,wang2023images,bai2023sequential}. In \textbf{robotics and embodied AI} domain, the robots actions in response to the environments can be coded into various discrete tokens and learn the policy in NTP manner as shown in recent studies such as VIMA~\cite{jiang2023vima}, RT2~\cite{brohan2023rt2} and Locomotion NTP~\cite{humanoid}.
In \textbf{AI4Science}, by factorizing various proteins into DNA token sequences, protein language models are capable of learning from a wide array of sequences that span the evolutionary tree. These models have demonstrated their efficacy as powerful tools for sequence design and protein engineering, as highlighted in studies ~\cite{madani2023large,ruffolo2024designing}.

\subsection{Continuous Tokenization Basics} 
\label{sec:continuous tokens}

\begin{figure}[h]
    \centering
    \includegraphics[width=\linewidth]{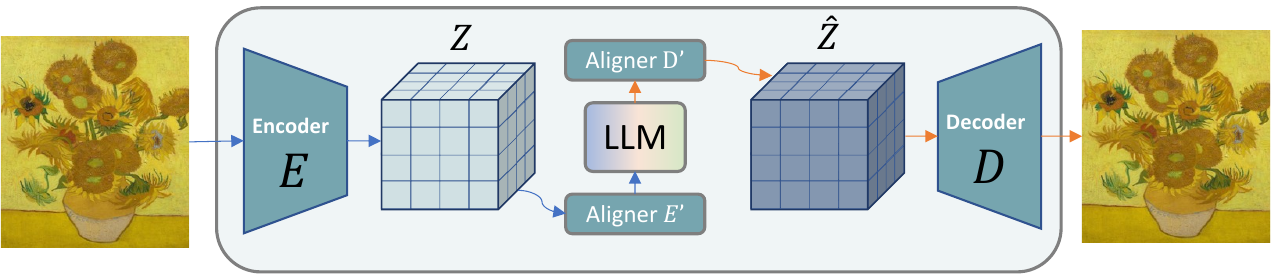}
    \caption{Illustration of Continuous Tokens. Blue lines denote the encoding process, where the encoder transforms the input image into a latent representation \( Z \), and aligner \( E' \) generates continuous tokens as input for the LLM to understand image content. Orange lines denote the generation process, where the LLM produces continuous tokens via aligner \( D' \), creating a latent representation \( \hat{Z} \) for the decoder to reconstruct or generate an image.}
    \label{fig:soft-tokens}
\end{figure}

Continuous tokens represent non-textual modalities in a continuous feature space, offering less information loss~\citep{dnd-transformer} and improved data representation compared to discrete tokens~\citep{xie2024showosingletransformerunify}. However, their dense feature encapsulation makes direct mapping to a fixed vocabulary challenging, unlike discrete tokens. It poses a challenge for LLMs aiming to comprehend and generate such information in a NTP manner. 

To handle continuous multimodal token inputs for LLM to understand, transformations or adapters are necessary to balance data representation and text alignment. For multimodal generation, modifying the output head to align with non-textual modality specific decoders' input feature space is also crucial. The following subsections introduce the basic designs and change for LLMs to accommodate continuous multimodal token from multimodal understanding (\S\ref{sec: Tokenize Continuous Input}) and generation (\S\ref{sec: De-tokenize Continuous Output}) perspectives.

\subsubsection{Tokenize Continuous Input for Understanding}
\label{sec: Tokenize Continuous Input}

To effectively integrate raw non-textual modality data into Large Language Models (LLMs), two key steps are typically undertaken: (1) encoding the data into a more suitable representation space, and (2) aligning it with the LLM’s feature space. 

\paragraph{Encoding} 
The encoding of non-textual modality data aims to capture meaningful features and important nuances that are essential for the understanding of the data. This can be achieved through different types of encoders such as Transformer-based encoders~\citep{li2023blip2, liu2023llava, liu2023llava15,zhu2023minigpt4, radford2021clip} or CNN-based encoders~\citep{davinci, zhang2023universal, jiang2023vima, alayrac2022flamingo}. There's also an option to go encoder-free~\citep{kim2021vilt, fuyu}, which allows for raw data to be fed directly into the model.

Transformer-based encoders are widely used for their robust representation capabilities and generalizability~\citep{vaswani2017attention, vit}. For a non-textual modality sample, the input is initially divided into patches and transformed into a 1D sequence, with each patch represented as a soft token. This sequence is then processed through the Transformer's encoder layers, employing self-attention mechanisms to capture relationships between patches. Consequently, the model produces a rich representation of the input. Typically, there are two types of encoders: (1) unimodal encoders, designed to process information from a single modality~\citep{vit, sam, arnab2021vivit, usm, mert, prismer, beit, liu2021swinTransformer}; and (2) multi-modal encoders, capable of integrating information from multiple modalities~\citep{radford2021clip, imagebind, eva-clip, clap, anymal, imu2clip, coca}. For instance, PaLM-E~\citep{driess2023palme}, Unified-IO-2~\citep{lu2023unifiedio}, and PaLI~\citep{pali} use ViT~\citep{vit} encoders trained solely on visual data. Conversely, LLaVA~\citep{liu2023llava}, Emu~\citep{sun2023emu1, sun2023generative}, and Qwen-VL~\citep{QwenVL} utilize CLIP~\citep{radford2021clip} or EVA-CLIP~\citep{eva-clip} encoders with contrastive loss to align textual and non-textual representations. NExT-GPT~\citep{nextgpt}, CoDi-2~\citep{tang2023codi2}, and BuboGPT~\citep{zhao2023bubogpt} employ ImageBind~\citep{imagebind} as their non-textual encoder, aligning various modalities like audio, text, and heat maps with image representations.

In comparison, CNN-based encoders are less frequently used but remain vital due to their flexibility in image resolution generalization~\citep{magvit, magvit2} and ability to capture local features~\citep{jiang2023vima}. For example, DaVinCi~\citep{davinci} uses ResNet~\citep{resnet} as the visual encoder. Flamingo~\citep{alayrac2022flamingo} utilizes NFNet~\citep{nfnet}, a normalizer-free ResNet, for image encoding.

Beyond encoders, Fuyu-8B~\citep{fuyu} directly processes raw image patches after a single linear projection to accommodate images of varying resolutions and aspect ratios, similar to ViLT~\citep{kim2021vilt}. However, Fuyu-8B adds the flexibility of an any-resolution setting using a decoder-only model, benefiting from architectural simplicity but showing reduced downstream performance compared to encoder-based models. Moreover, ImageGPT~\citep{imagegpt} trains a decoder-only generative model on raw image pixel sequences, which, despite its effectiveness in image generation and understanding, requires significant computational resources and is limited to low-resolution images.

\paragraph{Input Alignment.} 
After encoding non-textual modality data, we obtain a meaningful representation. However, this representation often lacks alignment with the textual embedding space of large language models, leading to a failure in properly understanding these inputs. Although multi-modal encoders like CLIP~\citep{radford2021clip} have made strides in narrowing the gap, they still encounter two significant challenges: (1) the presence of redundant continuous tokens~\citep{alayrac2022flamingo, perceiver, li2023blip2}; and (2) a lack of contextual semantics, such as causal semantics, because they are typically trained only with image-caption paired data rather than image-text interleaved data or image-prompt instructional data~\citep{seed-tokenizer, gemini1, laurencon2023obelics, zhu2023multimodal}. Therefore, it is crucial to establish a connection between the representation space of non-textual modality data and the LLM textual embedding space. There are typically two approaches to construct such a bridge: (1) Slot-based Resampler~\citep{alayrac2022flamingo, li2023blip2}; and (2) Projection~\citep{fuyu, liu2023llava, liu2023llava15,QwenVL}. 

The Slot-based Resampler compresses redundant non-textual modality tokens from the encoding stage into fewer learned query vectors, known as slots. This is typically accomplished using multiple Transformer blocks with a cross-attention mechanism. For instance, BLIP-2~\citep{li2023blip2} employs a Q-Former and linear projection to bridge the image encoder with the LLM backbone. The Q-Former blocks consist of a self-attention layer on the learned queries, a cross-attention layer between the encoded image representation and the learned queries, and a feed-forward layer. Initially, it is trained for image-text matching, image-text contrastive learning, and image-grounded text generation, followed by training for next token prediction with the frozen LLM backbone. Another model using this approach is Flamingo~\citep{alayrac2022flamingo}, which utilizes a Perceiver Resampler~\citep{perceiver} to compress byte arrays into latent vectors in a modality-agnostic manner. Specifically, Perceiver~\citep{perceiver} employs multiple cascaded attention mechanisms: the latents act as queries and initially cross-attend to keys and values calculated from the byte array (e.g., an image), followed by processing with a self-attention block, iterating several times. PerceiverIO~\citep{perceiverio} enhances this with an additional cross-attention block between an output query array and the slots (i.e., the latents). The Hierarchical Perceiver~\citep{hip} decomposes the input array into multiple groups, compresses each group, and merges the resulting latents to obtain the output array.

Compared to a slot-based resampler, projection is much simpler in architecture, involving only a single linear projection~\citep{fuyu, liu2023llava} or an Multi-layer Perceptron (MLP) ~\citep{liu2023llava15}. For instance, LLaVA~\citep{liu2023llava} employs a linear projection to convert encoded image representations into the language embedding space. Similarly, Fuyu-8B~\citep{fuyu} projects raw image patches onto the embedding space. LLaVA-1.5~\citep{liu2023llava15} enhances LLaVA by substituting the linear projection with an MLP.

There are also other approaches to connect the non-textual modality encoder with the LLM backbone. For example, Emu~\citep{sun2023emu1} leverages a Causal Transformer (i.e., C-Former) to convert the image tokens autoregressively; Emu2~\citep{sun2023generative} replaces the C-Former with mean pooling followed by a linear projection.

\subsubsection{De-tokenize Continuous Output for Generation}
\label{sec: De-tokenize Continuous Output}

The backbone of large language models is inherently designed for language generation. Typically, their output layers function as classification heads that predict distributions over a language vocabulary. For discrete non-textual modalities, the discrete token vocabularies can be integrated into the LLM’s original text vocabulary since token generation is still managed by the classification heads. However, this approach does not work for continuous non-textual modalities. To enable the generation of continuous token outputs from LLM backbones, it is essential to modify their output layers (i.e., language modeling heads) to produce representations suited for non-textual modality data. These representations are then transformed to align with the input features of specific non-textual modality data decoders, such as a diffusion model~\citep{Rombach_Blattmann_Lorenz_Esser_Ommer_2022}. Recent work includes MAR~\citep{MAR} and Transfusion~\citep{Transfusion}. We will further elaborate on the decoding of continuous output in ~\S\ref{par:soft-token-output-decoding} and the transformations to the output feature in ~\S\ref{par:soft-token-output-transformation}.

\paragraph{Decoding\label{par:soft-token-output-decoding}}

Unlike pure text generation, multimodal generation requires the model to decide when to switch modalities during decoding, due to their intrinsic differences. We refer to this objective as \textbf{positioning}. There are typically two methods to achieve this: (1) using placeholders~\citep{zheng2023minigpt5, nextgpt, koh2023GILL}; and (2) employing a non-textual modality begin-of-sentence (BOS) token~\citep{sun2023emu1, sun2023generative, dreamllm}.

Firstly, special tokens can be introduced as placeholders for non-textual modality data. For instance, Mini-GPT5~\citep{zheng2023minigpt5} and GILL~\citep{koh2023GILL} utilize a sequence of image placeholder tokens ranging from [IMG1] to [IMGr], which can be interleaved with textual tokens, and these tokens are added to the model's vocabulary. Likewise, NExT-GPT~\citep{nextgpt} uses 5 image placeholder tokens, along with 9 audio and 25 video placeholder tokens. Secondly, the use of a single BOS token (sometimes accompanied by an EOS token) can simplify the process by signaling the position of non-textual modality data. For example, DreamLLM~\citep{dreamllm} employs a special <dream> token to mark the start of modality switching, allowing a single model run to process a sequence of queries. Emu~\citep{sun2023emu1} and Emu2~\citep{sun2023generative} use both image BOS and EOS tokens to encase encoded image features.

In addition to focusing on positioning, models must also learn to generate accurate features for non-textual modalities. Typically, the output layers of large language models (LLMs) feature classification heads for discrete token decoding, an objective we refer to as \textbf{output representation}. To enable continuous token outputs, modifications to these output layers are required. Generally, there are three approaches: (1) adapting the original language modeling head to be regressive~\citep{sun2023emu1, sun2023generative}; (2) introducing a new head for dense outputs~\citep{dreamllm}; and (3) utilizing the final hidden states before the language model head~\citep{zheng2023minigpt5, koh2023GILL}.

\paragraph{Output Alignment\label{par:soft-token-output-transformation}}
Typically, generated continuous tokens cannot be directly used for multimodal generation because they don't align with the input features of multimodal decoders like LDM~\citep{Rombach_Blattmann_Lorenz_Esser_Ommer_2022} and AudioLDM~\citep{audioldm}. To address this, additional modules are introduced to convert these tokens into representations suitable for multimodal decoders, ultimately generating the final non-textual modality data. For instance, NExT-GPT~\citep{nextgpt} employs a Transformer-based output projection, while Mini-GPT5~\citep{zheng2023minigpt5} and GILL~\citep{koh2023GILL} utilize a Q-Former-like architecture~\citep{li2023blip2} consisting of a Transformer encoder and decoder to transform continuous tokens into conditional latent features for the Stable Diffusion Model. DreamLLM~\citep{dreamllm} uses a linear layer, whereas Emu~\citep{sun2023emu1} and Emu2~\citep{sun2023generative} directly utilize the generated continuous tokens as latents for multimodal decoders.

\subsection{Continuous Tokenization for Different Modalities}
\label{sec:continous app}

While the aforementioned workflow and categorization outline a general approach to continuous multimodal tokenization, research indicates that employing modality-specific encoders, tailored to each modality, can significantly enhance performance~\citep{navit, fixres, anymal}. Given the unique characteristics of different modalities, these approaches introduce specific inductive biases into the tokenization process.

\subsubsection{Images}

For images, specific research directions include but are not limited to: \textbf{image augmentation}, \textbf{resolution and aspect ratio} and \textbf{heterogeneous images}.

(1) Image Augmentation: This involves enhancing image representation using elements like depth, edge, and segmentation~\citep{prismer, sam, samclip}. Prismer~\citep{prismer}, for instance, introduces features beyond traditional RGB patches, such as depth and normal patchification. These features are compressed with a shared experts resampler before being integrated by a unified image encoder. SAM-CLIP~\citep{samclip} leverages SAM~\citep{sam} and the CLIP text encoder for distillation training, boosting the semantic and spatial comprehension of the image encoder.

(2) Resolution and Aspect Ratio: This strategy includes support for high-resolution images, multi-resolution capabilities, and arbitrary aspect ratios~\citep{msvit, navit, fuyu, llava-uhd, ureader}. For example, Fuyu~\citep{fuyu} uses raw pixels as image encoding inputs for the LLM backbone via linear projection, employing a special image newline token for delineating raster-ordered patches. This enables support for various resolutions and aspect ratios. MS-ViT~\citep{msvit} suggests varying patchification based on image region complexity, introducing a gate mechanism to mark tokens needing finer patchification, which then undergoes encoding after position encoding interpolation.

(3) Heterogeneous Images: This includes encoding methods for specific image types like vector images, diagrams, charts, and PDFs~\citep{layoutlm, textmonkey, ureader}. Document images, for example, require detailed observation, as seen in TextMonkey~\citep{textmonkey}, which splits large document images into smaller sub-images. Each sub-image is encoded individually, and trainable shifted attention layers are added post-frozen ViT layers for interactive representation across sub-images. These are then compressed and fed into the LLM backbone via an image and token resampler.

\subsubsection{Audio}


Recently, MELLE~\citep{meng2024autoregressive} indicates that predicting continuous tokens in an NTP manner can generate audio with high quality and naturalness comparable to ground truth.
Traditionally, audio frames are converted from the temporal domain to the frequency domain using the Short-Time Fourier Transform (STFT)~\cite{griffin1984signal} or the Fast Fourier Transform (FFT)~\cite{duhamel1990fast}.
The magnitude of the Fourier-transformed frames is modeled as spectrogram, which is a 2D image showing how the frequency content of the signal evolves over time. 
Spectrograms or other transformations of raw audio signals are additionally going through the feature selection pipeline before converting into discrete tokens.
Mel-Frequency Cepstral Coefficients (MFCCs)~\cite{furui1986speaker} extracts coefficients that represent the short-term power spectrum of sound and is one of the most common features used in speech recognition.
Mel-spectrogram~\cite{furui1986speaker} converts the spectrogram to the mel scale, which is more perceptually relevant to human hearing. These continuous features are commonly used in audio generation tasks.

Pre-trained foundation models, typically learned in a self-supervised manner on large-scale corpora, have emerged as powerful speech and audio representation extractors~\citep{latif2023sparks}. To obtain general speech features, wav2vec 2.0~\citep{baevski2020wav2vec} masks speech input in the latent space and addresses a contrastive task defined over quantized latent representations that are learned simultaneously. data2vec~\citep{baevski2022data2vec} biases the query-key attention scores with a penalty proportional to their distance. HuBERT~\cite{hsu2021hubert} employs an offline clustering step to provide aligned target labels for a BERT-like prediction loss, which is applied solely on the masked regions. WavLM~\cite{chen2022wavlm} introduces denoising in pretraining, jointly with regular masked speech prediction, as HuBERT. Whisper~\citep{radford2023robust}  is a speech recognition model characterized by an attention-based encoder-decoder architecture, trained on web-scale labeled speech data. It is increasingly being employed as a foundational speech model, extending its applications beyond speech recognition tasks~\citep{hu2024wavllm,tang2023salmonn,meng24c_interspeech,meng2024llm}.

For continuous tokenization of audio, AST~\citep{ast} uses a convolution-free pure-transformer architecture to extract features for audio classification, drawing insights from ViT~\citep{vit}. Inspired by CLIP~\citep{radford2021clip}, CLAP~\citep{clap} introduces a contrastive language-audio pre-training task to learn text-enhanced audio representations using supervised audio and text pairs. Fine-tuned based on a pre-trained CLIP model, Wav2CLIP~\citep{wu2022wav2clip} and AudioCLIP~\citep{guzhov2022audioclip} incorporate an additional audio encoder using supervised pairs of audio and class labels. Audio-MAE~\citep{huang2022masked} adopts a Transformer-based encoder-decoder framework to learn audio representations. Similar to MAE, it uses a reconstruction pre-training task where the decoder is tasked with reconstructing masked patches from the encoded information of the unmasked patches. BEATs~\citep{chen2022beats} introduces a self-distilled tokenizer that converts continuous audio signals into discrete labels, facilitating classic mask and discrete label prediction pre-training.


\subsubsection{Video}

Video can be viewed as a sequence of images (frames) over time, making the modeling of temporal relationships between these frames a central focus. There are two common approaches to this modeling: post-temporal fusion and full-temporal fusion. 

In the case of \textbf{post-temporal fusion}, models such as CLIP4Clip~\citep{luo2022clip4clip} and CLIPBERT~\citep{lei2021less} first independently encode each frame using an image encoder. They then employ lightweight pooling, convolution, and attention mechanisms to temporally fuse the features from all frames. The advantage of this approach lies in its ability to leverage pre-trained image encoders, thereby reducing the computational overhead associated with adapting to video data. However, a significant drawback is its limited capacity to adequately model features in the temporal dimension.

On the other hand, \textbf{full spatial-temporal fusion} models, like Temporal 3D ConvNets~\citep{diba2017temporal}, VideoMAE~\citep{tong2022videomae}, and ViViT~\citep{arnab2021vivit}, utilize 3D convolutions or 3D attention structures, allowing for comprehensive interaction among inputs in the spatio-temporal dimension. This enables better modeling of dynamic changes in temporal order, effectively capturing the motion of objects and backgrounds. However, this approach requires substantial 3D computation, prompting common strategies such as decoupling temporal and spatial self-attention~\citep{bertasius2021space, ren2023testa} and implementing sparse 3D attention~\citep{lin2022swinbert} to enhance computational efficiency.

Recent advancements, such as TimeChat~\citep{ren2024timechat} and NumPro~\citep{wu2024number}, have explored the integration of timestamp information into continuous video tokens, facilitating explicit time-vision associations for improved temporal grounding and reasoning.

\section{Backbone Model for Multimodal Next Token Prediction} 
\label{sec:model}
\begin{figure}[t]
\centering
\includegraphics[width=\textwidth]{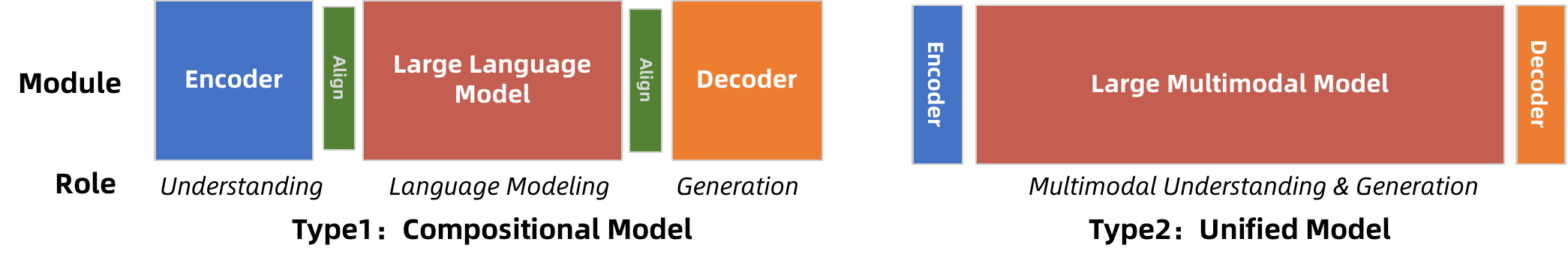}
\caption{Two types of Multimodal Next Token Prediction models. Compositional model utilizes powerful encoder and decoder for understanding and generation task, adding additional alignment layers. Unified model uses light-weighted encoder and decoder and leave most of the understanding and generation job to the backbone model.
}
\label{fig:two type of MMNTP models}
\end{figure}

\begin{table}[t!]
\centering

\caption{Summary of MMNTP model structures for different modalities.}
\label{table:mmntp_structure_summary}

\resizebox{1\textwidth}{!}{
\begin{tabular}{cc|cccccccc}
\toprule
   Model & Year & Modality & Und. & Gen. & Task & Backbone & Tokenization & Encoder & Decoder \\ \midrule
Flamingo~\citep{alayrac2022flamingo} & 2022 & Image & \ding{51} & \ding{55} & I2T & Compositional & Continuous & NFNet~\citep{brock2021highperformancelargescaleimagerecognition} & \ding{55} \\
DALLE~\citep{DALLE} & 2022 & Image & \ding{55} & \ding{51} & T2I & Unified & Discrete & dVAE~\citep{DALLE} & dVAE~\citep{DALLE} \\
Unified-IO~\citep{lu2022unifiedio} & 2022 & Image & \ding{51} & \ding{51} & I2T, T2I & Unified & Discrete & VQGAN~\citep{cao2023efficientvqgan} & VQGAN~\citep{cao2023efficientvqgan} \\
AudioLM~\citep{borsos2023audiolm} & 2022 & Audio & \ding{55} & \ding{51} & A2A & Compositional & Discrete & w2v-BERT~\citep{chung2021w2v}, SoundStream~\citep{zeghidour2021soundstream} & SoundStream~\citep{zeghidour2021soundstream} \\
AudioGen~\citep{kreuk2022audiogen} & 2022 & Audio & \ding{55} & \ding{51} & T2A & Unified & Discrete & Encodec~\citep{encodec} & Encodec~\citep{encodec} \\
MiniGPT4~\citep{zhu2023minigpt4} & 2023 & Image & \ding{51} & \ding{55} & I2T & Compositional & Continuous & EVA-CLIP~\citep{eva-clip} & \ding{55} \\
LLaVA~\citep{liu2023llava} & 2023 & Image & \ding{51} & \ding{55} & I2T & Compositional & Continuous & CLIP~\citep{radford2021clip} & \ding{55} \\
BLIP2~\citep{li2023blip2} & 2023 & Image & \ding{51} & \ding{55} & I2T & Compositional & Continuous & EVA-CLIP~\citep{eva-clip} & \ding{55} \\
Kosmos-1~\citep{peng2023kosmos} & 2023 & Image & \ding{51} & \ding{55} & I2T & Compositional & Continuous & CLIP~\citep{radford2021clip} & \ding{55} \\
InternVL~\citep{InternVL} & 2023 & Image & \ding{51} & \ding{51} & I2T & Compositional & Continuous & InternViT~\citep{InternVL} & \ding{55} \\
QwenVL~\citep{QwenVL} & 2023 & Image & \ding{51} & \ding{55} & I2T & Compositional & Continuous & OpenCLIP~\cite{OpenCLIP} & \ding{55} \\
Fuyu~\citep{fuyu} & 2023 & Image & \ding{51} & \ding{55} & I2T & Unified & Discrete & Image Patch~\citep{fuyu} & \ding{55} \\
Molom~\citep{Molmo} & 2023 & Image & \ding{51} & \ding{55} & I2T & Compositional & Continuous & CLIP~\citep{radford2021clip} & \ding{55} \\
Codi-2~\citep{tang2023codi2} & 2023 & Image, Video, Audio & \ding{51} & \ding{51} & I2T, T2I, A2T, T2A, T2V, V2T & Compositional & Continuous & ImageBind~\citep{imagebind} & Diffusion(SD\$\text{2.1}\$)~\cite{Rombach_Blattmann_Lorenz_Esser_Ommer_2022} $\text{+}$ AudioLDM2~\cite{AudioLDM2} $\text{+}$ zeroscope\footnote{\url{https://huggingface.co/cerspense/zeroscope_v2_576w}} \\
MiniGPT5~\citep{zheng2023minigpt5} & 2023 & Image & \ding{51} & \ding{51} & I2T, T2I & Compositional & Continuous & EVA-CLIP~\citep{eva-clip} & Diffusion(SD$\text{2.1}$)\cite{Rombach_Blattmann_Lorenz_Esser_Ommer_2022} \\
Blip-Diffusion~\citep{li2023blipdiffusionpretrainedsubjectrepresentation} & 2023 & Image & \ding{55} & \ding{51} & T2I & Compositional & Continuous & EVA-CLIP~\citep{eva-clip} & Diffusion(SD$\text{1.5}$)~\cite{sd} \\
Kosmos-G~\citep{Kosmos-G} & 2023 & Image & \ding{55} & \ding{51} & T2I & Compositional & Continuous & CLIP~\citep{radford2021clip} & Diffusion(SD$\text{1.5}$)~\cite{sd} \\
Unified-IO2~\citep{lu2023unifiedio2} & 2023 & Image, Video, Audio & \ding{51} & \ding{51} & I2T, T2I, A2T, T2A, V2T & Compositional & Continuous$\text{+}$Discrete & AST~\citep{ast} $\text{+}$ OpenClip~\citep{OpenCLIP} & VQGAN~\citep{cao2023efficientvqgan} \\
Emu1~\citep{sun2023emu1} & 2023 & Image & \ding{51} & \ding{51} & I2T, T2I & Compositional & Continuous & EVA-CLIP~\cite{eva-clip} & Diffusion(SD$\text{1.5}$)~\cite{sd} \\
Emu2~\citep{EMU2} & 2023 & Image & \ding{51} & \ding{51} & I2T, T2I & Compositional & Continuous & EVA-CLIP~\cite{eva-clip} & Diffusion(SDXL)~\cite{SDXL} \\
LaVIT & 2023 & Image & \ding{51} & \ding{51} & I2T, T2I & Compositional & Continuous & EVA-CLIP~\cite{eva-clip} & Diffusion(SD1,5)~\cite{sd} \\
GSLM~\citep{lakhotia2021generative} & 2021 & Audio & \ding{55} & \ding{51} & A2A & Unified & Discrete & CPC~\citep{van2017neural} / HuBERT~\citep{hsu2021hubert} / wav2vec~\citep{baevski2020wav2vec} & Tacotron-2~\citep{shen2018natural} \\
SPEAR-TTS~\citep{kharitonov2023speak} & 2023 & Audio & \ding{55} & \ding{51} & T2A & Compositional & Discrete & w2v-BERT~\citep{chung2021w2v}, SoundStream~\citep{zeghidour2021soundstream} & SoundStream~\citep{zeghidour2021soundstream} \\
Make-A-Voice~\citep{huang2023make} & 2023 & Audio & \ding{55} & \ding{51} & T2A & Compositional & Discrete & HuBERT~\citep{hsu2021hubert}, SoundStream~\citep{zeghidour2021soundstream} & SoundStream~\citep{zeghidour2021soundstream} \\
MusicGen~\citep{copet2024simple} & 2023 & Audio & \ding{55} & \ding{51} & T2A & Unified & Discrete & Encodec~\citep{encodec} & Encodec~\citep{encodec} \\
VALL-E~\citep{wang2023neural} & 2023 & Audio & \ding{55} & \ding{51} & T2A & Compositional & Discrete & Encodec~\citep{encodec} & Encodec~\citep{encodec} \\
SpeechGen~\citep{wu2023speechgen} & 2023 & Audio & \ding{55} & \ding{51} & T2A & Unified & Discrete & Unit mBART~\citep{popuri2022enhanced} & Unit mBART~\citep{popuri2022enhanced} \\
MU-LLaMA~\citep{meng2024autoregressive} & 2023 & Audio & \ding{51} & \ding{55} & A2T & Compositional & Continuous & MERT~\citep{li2023mert} & \ding{55} \\
Pengi~\citep{deshmukh2023pengi} & 2023 & Audio & \ding{51} & \ding{55} & A2T & Compositional & Continuous & CLAP~\citep{elizalde2023clap} & \ding{55} \\
LTU~\citep{gong2023listen} & 2023 & Audio & \ding{51} & \ding{55} & A2T & Compositional & Continuous & AST~\citep{gong2021ast} & \ding{55} \\
SpeechLLaMA~\citep{wu2023decoder} & 2023 & Audio & \ding{51} & \ding{55} & A2T & Compositional & Continuous & Transformer & \ding{55} \\
SALMONN~\citep{tang2023salmonn} & 2023 & Audio & \ding{51} & \ding{55} & A2T & Compositional & Continuous & Whisper~\citep{radford2023robust}, BEATs~\citep{chen2022beats} & \ding{55} \\
Qwen-Audio~\citep{chu2023qwen} & 2023 & Audio & \ding{51} & \ding{55} & A2T & Compositional & Continuous & Whisper~\citep{radford2023robust} & \ding{55} \\
AudioPaLM~\citep{rubenstein2023audiopalm} & 2023 & Audio & \ding{51} & \ding{51} & A2T, T2A, A2A & Compositional & Discrete & w2v-BERT~\citep{chung2021w2v}, USM~\citep{usm}, SoundStream~\citep{zeghidour2021soundstream} & SoundStream~\citep{zeghidour2021soundstream} \\

VIOLA~\citep{wang2023viola} & 2023 & Audio & \ding{51} & \ding{51} & A2T, T2A, A2A, T2T & Compositional & Discrete & Encodec~\citep{encodec} & Encodec~\citep{encodec} \\
LauraGPT~\citep{du2023lauragpt} & 2023 & Audio & \ding{51} & \ding{51} & A2T, T2A, A2A, T2T & Unified & Discrete & Encodec~\citep{encodec} & Codec Vocoder~\citep{du2023lauragpt} \\
SpeechGPT~\citep{zhang2023speechgpt} & 2023 & Audio & \ding{51} & \ding{51} & A2T, T2A, A2A, T2T & Unified & Discrete & HuBERT~\citep{hsu2021hubert} & HiFi-GAN~\citep{kong2020hifi} \\
LlamaGen~\citep{llamagen} & 2024 & Image & \ding{55} & \ding{51} & T2I & Unified & Discrete & VQGAN~\citep{cao2023efficientvqgan} & VQGAN~\citep{cao2023efficientvqgan} \\
VAR~\citep{VAR} & 2024 & Image & \ding{55} & \ding{51} & T2I & Unified & Discrete & Multi-scale VQVAE~\citep{VAR} & Multi-scale VQVAE~\citep{VAR} \\
DnD-Transformer\citep{dnd-transformer} & 2024 & Image & \ding{55} & \ding{51} & T2I & Unified & Discrete & RQVAE~\citep{lee2022RQVAE} & RQVAE~\citep{lee2022RQVAE} \\
Mini-Genimi~\citep{Mini-Gemini} & 2024 & Image & \ding{51} & \ding{51} & I2T, T2I & Compositional & Continuous & CLIP~\citep{radford2021clip} $\text{+}$ ConvNeXt~\citep{liu2022convnet2020s} & SDXL\citep{SDXL} \\
Chameleon~\citep{chameleonteam2024chameleon} & 2024 & Image & \ding{51} & \ding{51} & I2T, T2I & Unified & Discrete & VQ-SEG~\citep{make-a-scene} & VQ-SEG~\citep{make-a-scene} \\
MAR~\citep{MAR} & 2024 & Image & \ding{55} & \ding{51} & I2T & Unified & Continuous & \ding{55} & Mask-VQ \\
Fluid~\citep{FLUID} & 2024 & Image & \ding{55} & \ding{51} & I2T & Unified & Discrete & \ding{55} & Random-VQ \\
Transfusion~\citep{Transfusion} & 2024 & Image & \ding{51} & \ding{51} & I2T, T2I & Unified & Discrete & CLIP~\citep{radford2021clip} & AR-Diffusion~\citep{Transfusion} \\
VILA-U~\citep{VILA-U} & 2024 & Image, Video & \ding{51} & \ding{51} & I2T, T2I, T2V & Unified & Discrete & SigLIP~\citep{zhai2023sigmoidlosslanguageimage}, RQVAE~\cite{lee2022RQVAE} & RQVAE~\cite{lee2022RQVAE} \\
Show-o~\citep{Show-o} & 2024 & Image, Video & \ding{51} & \ding{51} & I2T, T2I, T2V & Unified & Discrete & MAGVIT-v2 & MAGVIT-v2 \\
MIO~\citep{wang2024mio} & 2024 & Image, Video, Audio & \ding{51} & \ding{51} & T2T, \{I/V/A\}2T, T2\{I/V/A\}, \newline \{I/V\}2\{I/V\}, A2A & Unified & Discrete & Seed-Tokenizer~\citep{seed-tokenizer}, SpeechTokenizer~\citep{zhang2023speechtokenizer} & Seed-Tokenizer~\citep{seed-tokenizer}, SpeechTokenizer~\citep{zhang2023speechtokenizer} \\
Emu3~\citep{Emu3} & 2024 & Image, Video & \ding{51} & \ding{51} & I2T, T2I, T2V & Unified & Discrete & MoVQGAN~\citep{MoVQ}\footnote{https://github.com/ai-forever/MoVQGAN} & MoVQGAN~\citep{MoVQ}\footnote{https://github.com/ai-forever/MoVQGAN} \\
Janus~\citep{Janus} & 2024 & Image & \ding{51} & \ding{51} & I2T, T2I & Compositional & Continuous$\text{+}$Discrete & SigLIP~\citep{zhai2023sigmoidlosslanguageimage} & VQGAN~\citep{llamagen} \\
VoiceCraft~\citep{peng2024voicecraft} & 2024 & Audio & \ding{55} & \ding{51} & T2A & Unified & Discrete & Encodec~\citep{encodec} & Encodec~\citep{encodec} \\
BASE TTS~\citep{lajszczak2024base} & 2024 & Audio & \ding{55} & \ding{51} & T2A & Unified & Discrete & WavLM~\citep{chen2022wavlm}, CNN & CNN \\
UniAudio~\citep{yang2023uniaudio} & 2024 & Audio & \ding{55} & \ding{51} & T2A, A2A & Unified & Discrete & Universal Codec~\citep{yang2023uniaudio} & Universal Codec~\citep{yang2023uniaudio} \\
CosyVoice~\citep{du2024cosyvoice} & 2024 & Audio & \ding{55} & \ding{51} & T2A & Compositional & Discrete & Conformer~\citep{gulati2020conformer} & Transformer, ResNet \\
FireRedTTS~\citep{guo2024fireredtts} & 2024 & Audio & \ding{55} & \ding{51} & T2A & Compositional & Discrete & HuBERT~\citep{hsu2021hubert}, ECAPA-TDNN~\citep{desplanques2020ecapa} & Transformer, ResNet \\
Seed-TTS~\citep{anastassiou2024seed} & 2024 & Audio & \ding{55} & \ding{51} & T2A & Compositional & Discrete & Unknown & Unknown \\
MELLE~\citep{meng2024autoregressive} & 2024 & Audio & \ding{55} & \ding{51} & T2A & Unified & Continuous & Mel-Spectrogram & Mel-Spectrogram \\
WavLLM~\citep{hu2024wavllm} & 2024 & Audio & \ding{51} & \ding{55} & A2T & Compositional & Continuous & Whisper~\citep{radford2023robust}, WavLM~\citep{chen2022wavlm} & \ding{55} \\
AudioFlamingo~\citep{kong2024audio} & 2024 & Audio & \ding{51} & \ding{55} & A2T & Compositional & Continuous & Clapcap~\citep{elizalde2024natural} & \ding{55} \\
SpeechVerse~\citep{das2024speechverse} & 2024 & Audio & \ding{51} & \ding{55} & A2T & Compositional & Continuous & WavLM~\citep{chen2022wavlm}, Best-RQ~\citep{chiu2022self} & \ding{55} \\
VoxtLM~\citep{maiti2024voxtlm} & 2024 & Audio & \ding{51} & \ding{51} & A2T, T2A, A2A, T2T & Unified & Discrete & HuBERT~\citep{hsu2021hubert} & HiFi-GAN~\citep{kong2020hifi} \\
LLaMA-Omni~\citep{fang2024llama} & 2024 & Audio & \ding{51} & \ding{51} & A2A & Compositional & Continuous & Whisper~\citep{radford2023robust} & Transformer \\
Mini-Omni~\citep{xie2024mini} & 2024 & Audio & \ding{51} & \ding{51} & A2A & Compositional & Continuous & Whisper~\citep{radford2023robust} & Transformer \\
Moshi~\citep{defossez2024moshi} & 2024 & Audio & \ding{51} & \ding{51} & A2A & Unified & Discrete & Mimi~\citep{defossez2024moshi} & Mimi~\citep{defossez2024moshi} \\
Qwen2VL~\citep{Qwen2vl} & 2024 & Image & \ding{51} & \ding{55} & I2T & Compositional & Continuous & Qwen2-VL ViT~\citep{Qwen2vl} & \ding{55} \\
EVE~\citep{eve} & 2024 & Image & \ding{51} & \ding{55} & I2T & Unified & Discrete$\text{+}$Continuous & Image Patch~\citep{solo} $\text{+}$ CLIP~\cite{radford2021clip} & \ding{55} \\
SOLO~\citep{solo} & 2024 & Image & \ding{51} & \ding{55} & I2T & Unified & Discrete & Image Patches~\citep{solo} & \ding{55} \\
MonoInternVL~\citep{monointernvl} & 2024 & Image & \ding{51} & \ding{55} & I2T & Unified & Discrete & Image Patch~\citep{monointernvl} & \ding{55} \\
RAR~\citep{RAR} & 2024 & Image & \ding{55} & \ding{51} &T2I & Unified & Discrete &  MaskGit-VQGAN~\citep{MaskGIT} & MaskGit-VQGAN~\citep{MaskGIT}\\
Infinity~\citep{han2024infinityscalingbitwiseautoregressive} & 2024 & Image & \ding{55} & \ding{51} &T2I & Unified & Discrete &  VAR~\citep{VAR} & VAR~\citep{VAR}\\
LatentLM~\citep{sun2024multimodallatentlanguagemodeling} & 2024 & Image &\ding{55} & \ding{51} &T2I & Unified & Continuous & \ding{55} & \ding{55} \\

   \bottomrule
\end{tabular}}
\end{table}

After multimodal information is tokenized into sequential tokens, we need a model capable of handling multimodal information. In the literature, two classic MMNTP model structures are depicted in Fig.~\ref{fig:two type of MMNTP models}: 1) the Compositional Model and 2) the Unified Model. The key distinction lies in their design: the Compositional Model relies on heavily trained external encoders and decoders (such as ~\citep{radford2021clip}), and Diffusion models~\citep{ho2020denoising}, for understanding and generation tasks respectively. In contrast, the Unified Model features lightweight encoders and decoders, with multimodal understanding and generation tasks primarily occurring within the backbone model, typically a large transformer decoder. A categorization of current MMNTP models is shown in Table~\ref{table:mmntp_structure_summary}. We will introduce the general structure of MMNTP model in Section~\ref{sec: general structure}, the recent advances in compostional and unified models in Sections~\ref{sec:comp model} and~\ref{sec:unified model}, and compare them in Section~\ref{sec:comparision}.

\subsection{Basic Structure of MMNTP Model}
\label{sec: general structure}

\begin{figure}[t]
\centering
\includegraphics[width=0.9\textwidth]{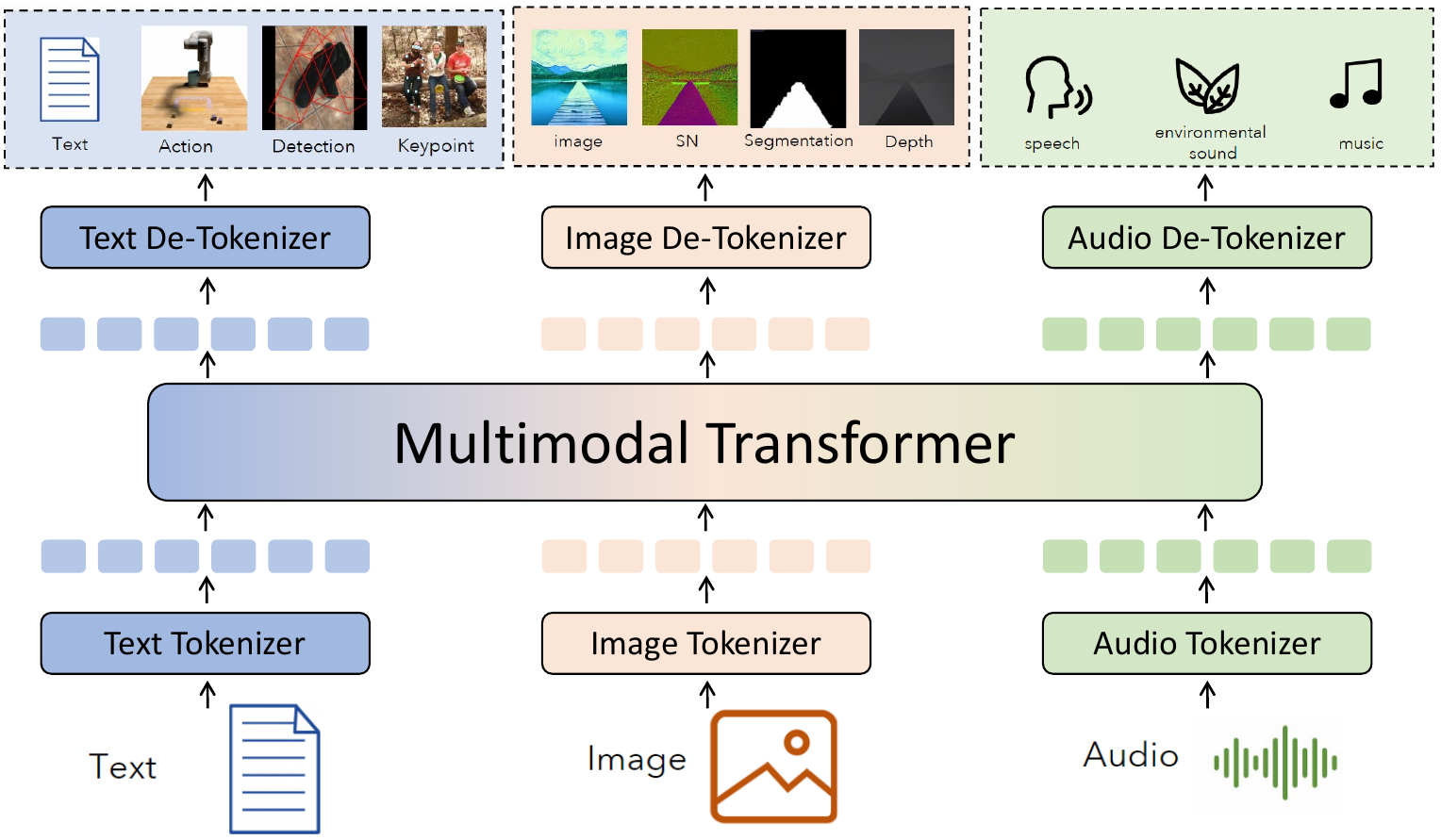}
\caption{Input text, images, audio, or image/audio history are encoded into sequences of tokens which are concatenated and used as input to an multi-modal transformer model. The transformer outputs discrete tokens that can be decoded into text, an image, or an audio clip. Some image examples are referenced from~\citep{lu2023unifiedio2}.
}
\label{fig:arch}
\end{figure}

\begin{figure}[t]
\centering
\includegraphics[width=\textwidth]{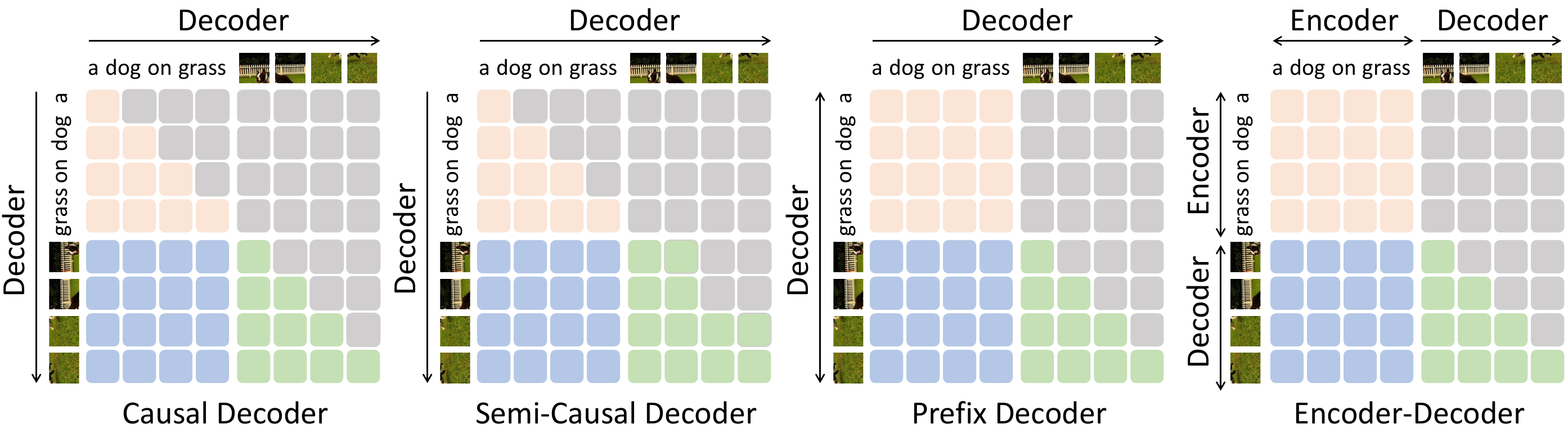}
\caption{Attention patterns in a causal decoder, non-causal decoder, and encoder-decoder
architecture. In a causal decoder, each token attends to the previous tokens only. In a semi-causal decoder, each token could attends to part of future token and all past tokens~\citep{VAR}. In both prefix-causal decoder and encoder-decoder, attention is allowed to be bidirectional on any conditioning information. For the encoder-decoder, that conditioning is fed into the encoder part of the model.
}
\label{fig:attn-mask}
\end{figure}

As shown in Fig.~\ref{fig:arch}, to implement multimodal understanding and generation as next token prediction, this typically involves three steps. 
\textbf{Step 1.} Encode various inputs – images, text, audio, action, boxes etc., into sequences of tokens in a shared representation space. \textbf{Step 2.} Use a multi-modal transformer to predict next token in an auto-regressive manner. \textbf{Step 3.} Decode the predicted tokens into the space of their respective modalities.

Fig.~\ref{fig:arch} also showcases the key modules of the NTP-based multimodal model, including tokenizers (encoders) and de-tokenizers (decoders) for each modality, as well as the multimodal Transformer. The tokenizer (encoder) and de-tokenizer (decoder) modules often appear together and are pretrained using unimodal data through techniques such as reconstruction. They have the capability to split the original input into tokens using the tokenizer (encoder) and restore the tokens back to their original form using the de-tokenizer (decoder). Once all the tokenizers (encoders) and de-tokenizers (decoders) for each modality are pretrained, we can activate the required tokenizer (encoder) separately for tokenization of input containing multiple modalities, enabling us to obtain a multimodal token sequence. Finally, these multimodal token sequences are fed into the multimodal Transformer for NTP training.

For multimodal Transformers, we can use different attention masks to control the flow of information from different modalities~\citep{Show-o,VAR}. As shown in Fig~\ref{fig:attn-mask}, a common attention mask is the causal mask, which requires each token to only depend on preceding context for generation. However, certain tasks require generating subsequent text conditioned on a content-rich input prefix, such as generating summaries based on a rich-text-format document. For such tasks, we can also utilize a non-causal mask, which applies a bidirectional attention to the prefix, allowing the context within the prefix to interdepend and provide better representation, while using causal attention for autoregressive generation of the content to be generated. In summary, we can flexibly select attention masks based on the requirements of the task.

\subsubsection{A Unified Structure for Vision Tasks}

\begin{figure}[t]
\centering
\includegraphics[width=\textwidth]{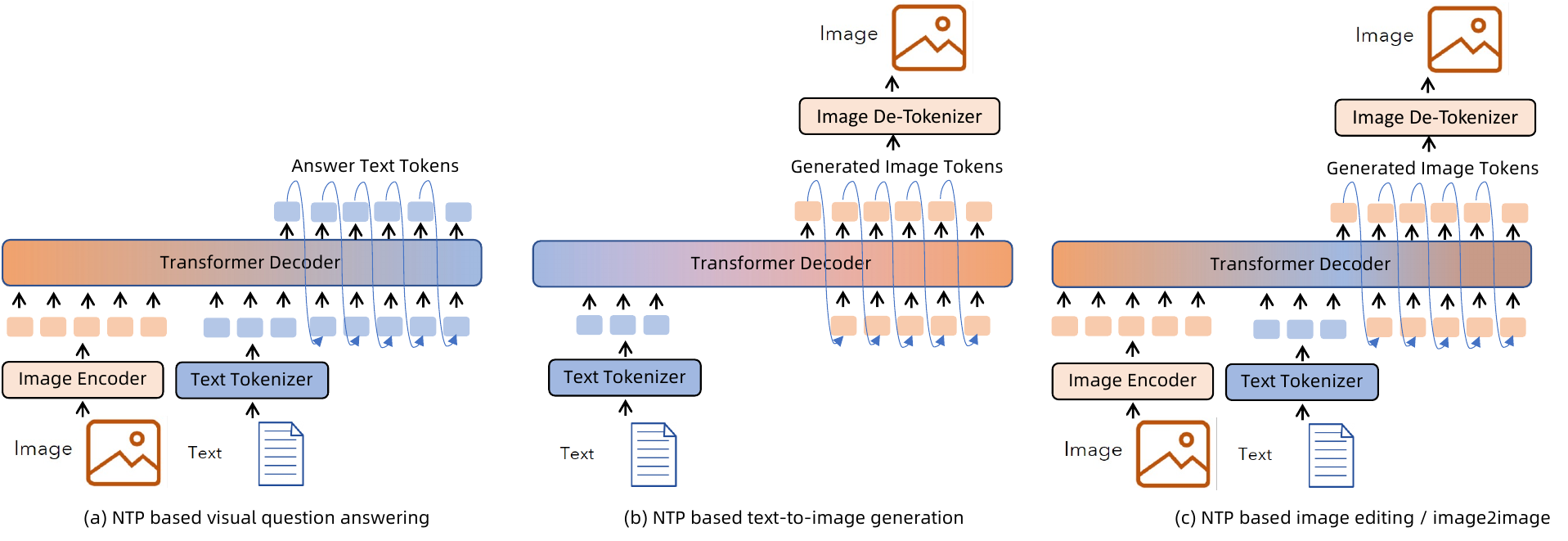}
\caption{Next token prediction based structures for (a) visual question answering, (b) text-to-image generation and (c) text guided image editing / image-to-image transform which require both image understanding and generation capabilities.
}
\label{fig:image_ntp}
\end{figure}

As illustrated in Fig.~\ref{fig:image_ntp}, various tasks in the vision modality can be encapsulated within the framework of MMNTP. Currently, a majority of large multimodal models (LMMs), such as LLaVA~\citep{liu2023llava} and the Qwen-VL~\citep{QwenVL,Qwen2vl} series, adhere to the NTP-based visual question answering paradigm. In this approach, images and text instructions are tokenized and sent to the transformer decoder to obtain the answer tokens. Another line of research, focusing on auto-regressive image generation, primarily adopts the NTP-based text-to-image generation paradigm, as seen in models like LlamaGen~\citep{llamagen}, VAR~\citep{VAR}, and DnD-Transformer~\citep{dnd-transformer}. Alternatively, the output image tokens can be generated in a non-causal order, as demonstrated by works like MaskGIT~\citep{MaskGIT} and RAR~\citep{RAR}. Additionally, these tokens can be continuous and later sent to a diffusion-based image de-tokenizer, as seen in recent developments like MAR~\citep{MAR} and Transfusion~\citep{Transfusion}. Some research combines the above paradigms to enable LMMs to perform both visual understanding and generation, as evidenced by models such as Show-o~\citep{Show-o}, Janus~\citep{Janus}, and Emu3~\citep{Emu3}. Specifically, the NTP paradigm also supports various image-to-image tasks, such as image editing and semantic segmentation, as distinguished by Unified-IO2 and LVM~\citep{bai2023sequential}.

\subsubsection{A Unified Structure for Audio Tasks}

\begin{figure}[t]
\centering
\includegraphics[width=\textwidth]{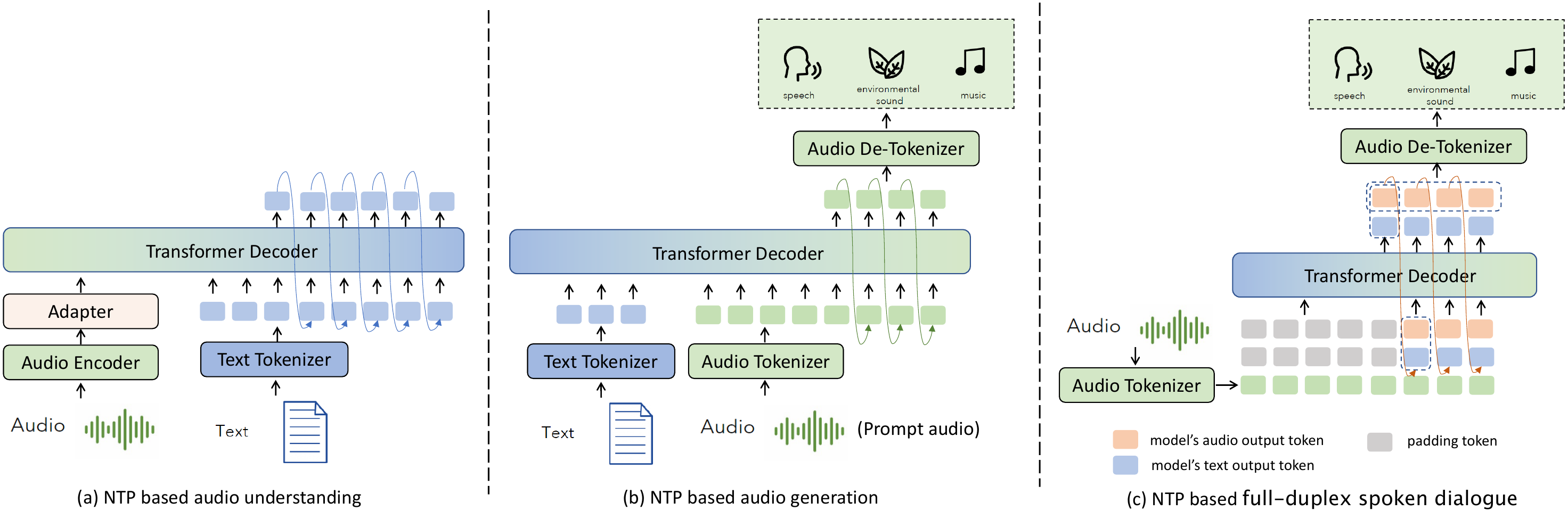}
\caption{Next token prediction based structures for (a) audio understanding, (b) audio generation and (c) full-duplex spoken dialogue, which requires both understanding and generation capabilities.
}
\label{fig:audio_ntp}
\end{figure}

As illustrated in Fig. \ref{fig:audio_ntp}, distinct NTP-based model architectures are required for various audio processing and generation tasks. For audio understanding \cite{hu2024wavllm, chu2023qwen, tang2023salmonn}, large-scale data pre-trained encoders demonstrate superior performance in extracting information from speech compared to discrete tokens. Additionally, an adapter is employed to facilitate the connection between the audio and text domains. Meanwhile, text instructions can specify specific audio processing tasks, such as automatic speech recognition, speech translation, and speech question answering. For audio generation, audio signals are typically transformed into discrete tokens \cite{wang2023neural, kreuk2022audiogen, lajszczak2024base} or continuous tokens \cite{meng2024autoregressive}. These tokens can subsequently be converted back into waveform format through the use of corresponding decoders or vocoders. The text serves as either the specific speech content to be synthesized,  or a detailed description of the audio. Leveraging the in-context learning capabilities and the scalability potential of the NTP based model, it achieves exceptional performance in zero-shot text-to-audio synthesis where a prompt audio is provided. Recently, the exploration of full-duplex real-time spoken dialogue \cite{defossez2024moshi, gpt4o} has been progressing at a rapid pace, which requires strong audio understanding and streaming speech generation capabilities. In Moshi~\cite{defossez2024moshi}, to address these requirements, multiple audio streams, encompassing both user inputs and model outputs, are modeled concurrently, and a novel streaming audio tokenizer is introduced. For these tasks, the parameters of the Transformer decoder can be effectively initialized using those derived from an LLM.

\subsection{Compositional Model}
\label{sec:comp model}

As shown in Fig.~\ref{fig:two type of MMNTP models}, the Compositional Model utilizes advanced external models to serve as the encoder and decoder for processing multimodal information. The section introduces the two components individually.

\subsubsection{Connecting External Encoders for Understanding}

A common architectural approach to enabling multimodal information understanding ability in LLM is using a robust external encoder to encode raw multimodal data to better representations. Pioneering work includes MiniGPT4~\citep{zhu2023minigpt4} and LLaVA~\citep{liu2023llava}, which combine a vision encoder, an alignment layer and an LLM for general-purpose visual and language understanding. The LLaVA-style structure~\citep{liu2023llava}, which uses CLIP~\citep{radford2021clip} as the encoder and an MLP as the alignment layer, has been utilized in numerous subsequent models. Recent studies reveal that scaling up the visual encoder~\citep{chen2023internvl,QwenVL} and allowing for more flexible input image resolutions~\citep{llava-uhd,QwenVL} can significantly improve the model's visual perception abilities. Similar architectural approaches are employed within the audio domain to equip LLMs with the ability to perceive and process speech signals, as exemplified by models such as SALMONN~\citep{tang2023salmonn}, Qwen-Audio~\citep{chu2023qwen}, and WavLLM~\citep{hu2024wavllm}. For a detailed discussion on encoder design, please refer to Section~\ref{sec: Tokenize Continuous Input}.

\subsubsection{Connecting External Decoders for Generation} To enable the LLM to generate multimodal outputs, including images, a straightforward approach is to connect it to a powerful image generation model such as a latent diffusion model~\citep{ldm}. In this context, it is crucial to ensure that the LLM generates continuous features beyond just language tokens, aligning the output with the input space of the diffusion models. Typical work includes Emu~\cite{sun2023emu1}, which adds a regression head on top of the LLM's output hidden state to predict the visual embedding for the diffusion model. For a detailed discussion on decoder design, please refer to Section~\ref{sec: De-tokenize Continuous Output}.

To enable both multimodal understanding and generation abilities of LLM in compositional manner, an external encoder and decoder can be attached to the backbone model simultaneously. A classic structure is exemplified by Emu1 and Emu2~\citep{sun2023emu1,sun2024emu}, which adopts EVA-CLIP~\citep{eva-clip} as the encoder and SDXL as the image decoder. For the audio domain, LLaMA-Omni~\citep{fang2024llama} utilizes Whisper-large-v3~\citep{radford2023robust} as the encoder and a Transformer based decoder.

\subsection{Unified Model}
\label{sec:unified model}

As shown in Fig.~\ref{fig:two type of MMNTP models}, the Unified Model leverages a light-weight encoder and decoder to process and generate multimodal information. The backbone model takes up most of the roles in understanding and generation tasks. This section will introduce two main structures of the unified model.

\subsubsection{Quantization-based Autoregression}

The quantization-based method is widely applied in building a unified model for multimodal understanding and generation due to its simplicity and similarity to the causal language modeling task. Typically, the encoder and decoder are derived from VQVAEs, trained to reconstruct the input from a discrete representation space. Focusing on generation, research explores generating images~\citep{DALLE,llamagen,VAR,dnd-transformer} and audio~\citep{kreuk2022audiogen, yang2023uniaudio, copet2024simple,lajszczak2024base} with higher quality in an autoregressive manner and integrating advanced techniques for optimizing LLMs. Another line of work focuses on both understanding and generating multimodal information using quantization-based methods. Notable examples include Unified-IO~\cite{lu2022unifiedio}, Chameleon~\cite{chameleonteam2024chameleon}, Emu-3~\cite{wang2024emu3nexttokenpredictionneed} and Moshi~\cite{defossez2024moshi}, which employ a unified NTP training objective for multimodal understanding and generation tasks.

\subsubsection{Autoregressive Diffusion}

The quantization-based method often faces criticism regarding generation quality. It typically produces images in a raster-scan order, which contradicts the intrinsic nature of 2D images. Additionally, the quantization process can lead to information loss. Several works aim to integrate the diffusion process into the NTP to enhance generation quality. Unlike compositional methods, the diffusion model is trained from scratch alongside the entire transformer model. Distinctive works such as Transfusion~\citep{Transfusion}, MAR~\citep{MAR}, CosyVoice~\citep{du2024cosyvoice} and Fluid~\citep{FLUID} demonstrate that diffusion models can be jointly trained with language modeling tasks, offering superior image generation quality compared to quantization-based methods. 

The debate between quantization-based and diffusion-based autoregressive  methods for image generation is on-going, highlighting the need for further research. For instance, while many diffusion-based AR methods~\citep{MAR,Transfusion} claim better generation quality compared to quantization method, Emu3~\citep{wang2024emu3nexttokenpredictionneed} significantly outperforms diffusion baselines like SDXL using a quantization-based AR approach. DnD-Transformer~\citep{dnd-transformer} showcased that quantization-based AR generation has superior performance in generating rich-text images than diffusion models. In summary, it is not concluded yet which modeling method has superior performance than another currently.

\subsection{Comparison Between Compositional and Unified Models}
\label{sec:comparision}

This subsection delves into a detailed comparison between compositional and unified models, evaluating their respective strengths and weaknesses in terms of general multimodal intelligence, training and deployment efficiency, and their potential to scale with increasing computational resources.

\paragraph{General Multimodal Intelligence.} Unified models handle multimodal understanding and reasoning within a single backbone model, whereas compositional models assign different tasks to specialized external models. Although NTP has transformed language intelligence, its impact on multimodal intelligence remains uncertain. Given this context, unified models are closer to a multimodal foundation model~\citep{li2023multimodalfoundationmodelsspecialists, Emu3} due to its end-to-end nature and it may hold more potential than their compositional counterparts, as they rely on a single NTP training objective, making them easier to scale compared to multi-module systems. We will discuss the scaling behavior of MMNTP models in Section~\ref{sec:challange_scaling_law}.

\paragraph{Training Efficiency.} Compositional models benefit from leveraging highly specialized external encoders and decoders, often resulting in reduced training time for new tasks since these components are pretrained separately. This modular approach allows for targeted updates, reusing existing powerful models without the need for extensive retraining of the entire system. In contrast, unified models leave most of the understanding and generation responsibility to one backbone model, leading to sub-optimal performance given the same amount of computation~\citep{Show-o}. This integrated training can be more resource-intensive, but it potentially facilitates a more coherent feature space across modalities within the LLM backbone, potentially enhancing overall performance on diverse multimodal tasks. 

\paragraph{Deployment Efficiency.} The unified model, particularly when using quantization-based methods, demonstrates significantly superior deployment efficiency compared to the compositional approach. A single unified transformer decoder backbone can effectively leverage the advanced techniques developed by the LLM community for accelerating both training and inference, such as Flash-Attention~\citep{flash-attention} and vLLM~\citep{vllm}. This capability is frequently cited as a key advantage of unified models, as highlighted by works like~\citep{Emu3,llamagen}.

\section{Training with Unified Multimodal Task Representation}
\label{sec:training}
Once content from various modalities has been tokenized into a sequence of tokens, with a unified backbone model, typically a decoder-only transformer model~\cite{vaswani2017attention}, we can undergo training to tackle a wide array of downstream understanding and generation tasks following different training objectives (refer to Section~\ref{sub:training_obj}). The training tasks are primarily divided into two categories, which resemble the training of large language models: Pretraining (refer to Section~\ref{subsub-ssl}) and Finetuning (refer to Section~\ref{subsub-sl}).


For a sequence of input tokens $x_{1\sim i-1} = \{ x_1, x_2, \ldots, x_{i-1} \}$ , the model predicts the next token $x_i \in V$. The general loss function $f$ for a single prediction could be written as:
\begin{equation}
    L(\theta) = f\left( y_i , p_{\theta}\left(x_i \mid x_{1\sim i-1} \right)\right),
\end{equation}
where:
\begin{itemize}
\item $L(\theta)$ is the loss, parameterized by the model parameters $\theta$ and loss function $f$.
  \item $V$ is the total vocabulary. We use $V_T$, $V_M$ to denote text split and multimdoal split of the full vocabulary, $V_S$ to denote the continuous tokens which are continuous vectors.
  \item $y_i$ represents the target output for the next token. In supervised training, $y_i$ is typically derived from labeled data, whereas in self-supervised training, $y_i$ can be constructed from the data itself without explicit labels, often using the true next token from the input sequence. In special cases, $y_i$ could involve multiple tokens, enabling parallel prediction of next tokens.

  \item $f$ is cross-entropy loss when $y_i$ is the discrete token distribution. $f$ can also have different forms like mean-square error if $y_i$ belongs to continuous tokens.

\end{itemize}

Different training tasks differ in the organization of given sequence $x_{1\sim i-1}$ and target label $y_i$. For self-supervised training, the sequence itself provides the target $y_i$, with the correct next token being used as the label. This allows the model to learn from the vast amounts of unlabeled multimodal data available, which consumes larger training resources. Supervised training would require explicit labeling of the next tokens, which can improve more specific downstream tasks at the cost of being more labor-intensive in the data collection period.

\subsection{Training Objectives}
\label{sec:training_objectives}

\begin{figure}[h]
    \centering
    \includegraphics[width=\linewidth]{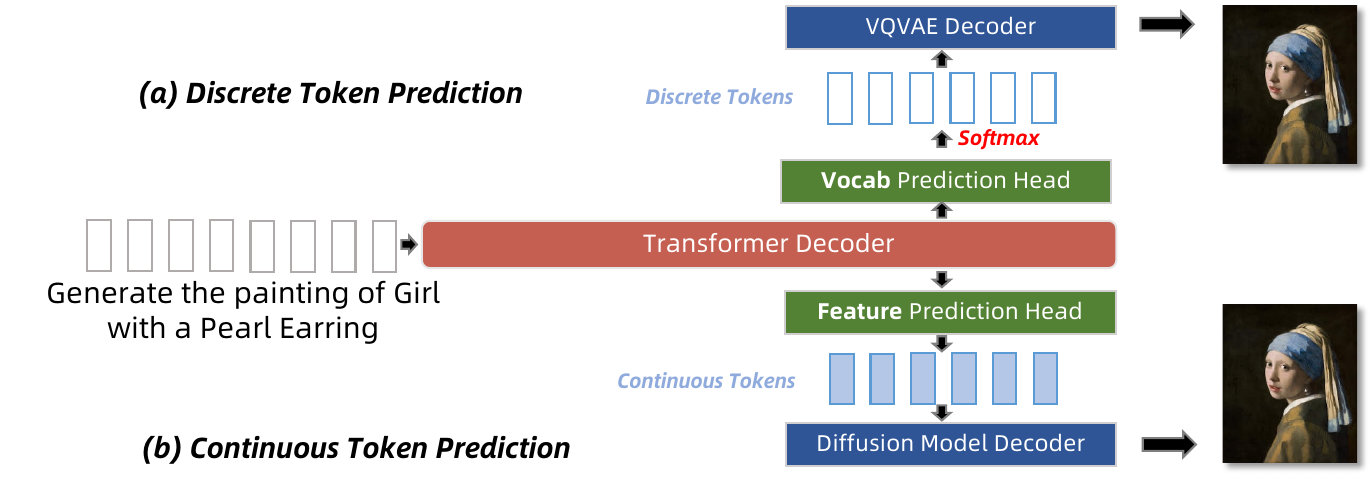}
    \caption{Training objectives example for text to image generation. (a) For the discrete token prediction task, the backbone model, typically a transformer decoder, processes text input to generate discrete image tokens. This is achieved using a vocabulary prediction head and a Softmax function. The resulting tokens are then passed to the VQVAE decoder to construct the final image. (b) For the continuous token prediction task, there is no application of the Softmax function to map the output hidden state to a fixed-size vocabulary within the feature prediction head. Instead, the output is directly processed by a vision generative module, such as a diffusion model. }
    \label{fig:training_obj}
\end{figure}

\label{sub:training_obj}
Based on what kind of target token $y_i$ to predict, the NTP training objectives could be further categorized into two classes: \textbf{ Discrete Token Prediction}, \textbf{Continuous Token Prediction} or a combination of them.

\begin{equation}
    y_i \in \left\{
             \begin{array}{lr}
             V_D, &\text{Discrete Token Prediction}  \\
             V_S, &\text{Continuous Token Prediction}   
             \end{array}
\right.
\end{equation}

In Fig~\ref{fig:training_obj}, we give an example using the task of text-to-image generation to show the difference between the two training objectives.

\subsubsection{Discrete-Tokens Prediction (DTP)} 
Discrete token Prediction (DTP) refers learn to predict the next discrete token given the context. The next token could belong to text or different modalities. This approach extends the conventional Causal Language Modeling (CLM), which typically deals with a unimodal text sequence, to accommodate inputs and outputs that interleave text with other data modalities, such as images. DTM enables the model to understand and generate different content from different modalities in a unified way. The training objective is to minimize the average cross-entropy loss among tokens.

Focusing on multimodal understanding ability, a majority of multimodal LLMs (e,g. Flamingo~\citep{madureira-2021-flamingos}, GPT4V~\citep{gpt4v}, MiniGPT4~\citep{zhu2023minigpt4}, Qwen-VL~\citep{QwenVL} and LLaVA~\citep{liu2023llava}) only predict language tokens $V_T$ given multimodal inputs. It leverages the powerful reasoning ability and world knowledge of LLMs to support various multimodal understanding tasks without re-pretraining the model.

Enlarging the output token space to discrete multimodal tokens $V_M$ like quantization codes would enable multimodal generation ability.  In this approach, multimodal contents are first converted into discrete tokens, utilizing cross-entropy loss as the loss function. A major line of works is auto-regressive multimodal information generation, such as DALLE~\cite{ramesh2021zeroshot}, CogView~\cite{CogView}, Unified-IO~\cite{lu2022unifiedio}, LVM~\cite{bai2023sequential} and Video-Poet~\cite{kondratyuk2023videopoet}.

Merging the two output spaces ($V_T$ and $V_M$) into one model is an intriguing direction~\cite{lu2022unifiedio,lu2023unifiedio2,liu2023world}, which naturally unifies multimodal understanding and generation tasks. However, some related research~\cite{zhang2023pretrained} shows that learning to predict text tokens have no benifit for predicting multimdoal tokens and sometimes lead to strong conflict. Under the NTP training framework, whether multimodal generation helps understanding ability also remains unclear. Consequently, effectively integrating the output spaces of text and multimodal tokens presents itself as one of the main challenges in the domain, underscoring the need for innovative and scalable approaches to harness the full potential of NTP models in the realm of multimodal learning.

A variant of standard next token prediction is to predict multiple tokens at one time, disobeying the causal order. Recent researches~\cite{MaskGIT,magvit2,tian2024VAR} have found that parallel prediction is more effective for visual domains such as images and videos than simple raster-based prediction, which predicts the image tokens from left to right and top to down. MaskGIT~\citep{MaskGIT} and MAGVIT~\citep{magvit2} predict a portion of tokens at each prediction step according to a dynamic confidence threshold. VAR~\citep{tian2024VAR} predicts the visual tokens in a resolution-autoregressive manner, which predicts tokens in the same resolution in parallel and predict low-to-high images in sequential. Those approaches inject different inductive bias for different modality during NTP modeling, which is also an important challenge when unifying multiple modalities in multimodal NTP framework.

\subsubsection{Continuous Token Prediction (CTP)} In addition to discrete multimodal tokens, the multimodal information can also be represented as continuous vectors, referred to as Continuous-tokens. The Continuous-tokens can be viewed as conditions for external model such as stable diffusion model for better generation quality. The continuous tokens are usually predicted auto-regressively with MSE loss~\cite{sun2023emu1,sun2023generative,zheng2023minigpt5,koh2023GILL,tang2023codi2}. For example, Emu-1 and Emu-2~\citep{sun2023emu1,sun2023generative} leverage a large language model to generate continuous tokens, which are used as condition for a pretrained diffusion model to generate images. The language model and diffusion model are trained simultaneously during the text-to-image instruction tuning stage. This method utilizes the powerful image generation ability of open-source diffusion model and unlocks the multimodal generation ability of large language model with modest additional cost. 

Beyond utilizing continuous tokens as conditions for external models, some researches explored using continuous tokens to directly generate images, replacing discrete tokens with continuous tokens throughout the NTP training paradigm.  \citet{AIM} reveals that when trained with L2 loss, a patch-based image Transformer exhibits scaling properties akin to those of LLMs. \citep{li2024denoising} represents image with continuous tokens and involves diffusion loss during training the causal transformer model. However, these models are trained solely on single modality such as image. Whether different training objectives for different modalities can coexist harmoniously in one NTP model remains under-explored.

\subsection{Pretraining: Modality Alignment}
\label{subsub-ssl}
Large Language Models have demonstrated their effectiveness and scalability in the pure language domain. In a similar vein, pioneering research is exploring the use of the abundant supply of multimodal data in training large multimodal models in NTP framework. The major focus of pretraining in LMM is to align the representation space of different modality with language space, which could be categorized into alignment in understanding (Section~\ref{sec: alignment understand}) and generation (Section~\ref{sec: alignment generation}) task.

\subsubsection{Modality Alignment in Understanding}
\label{sec: alignment understand}

\begin{figure*}
    \centering
    \includegraphics[width=\linewidth]{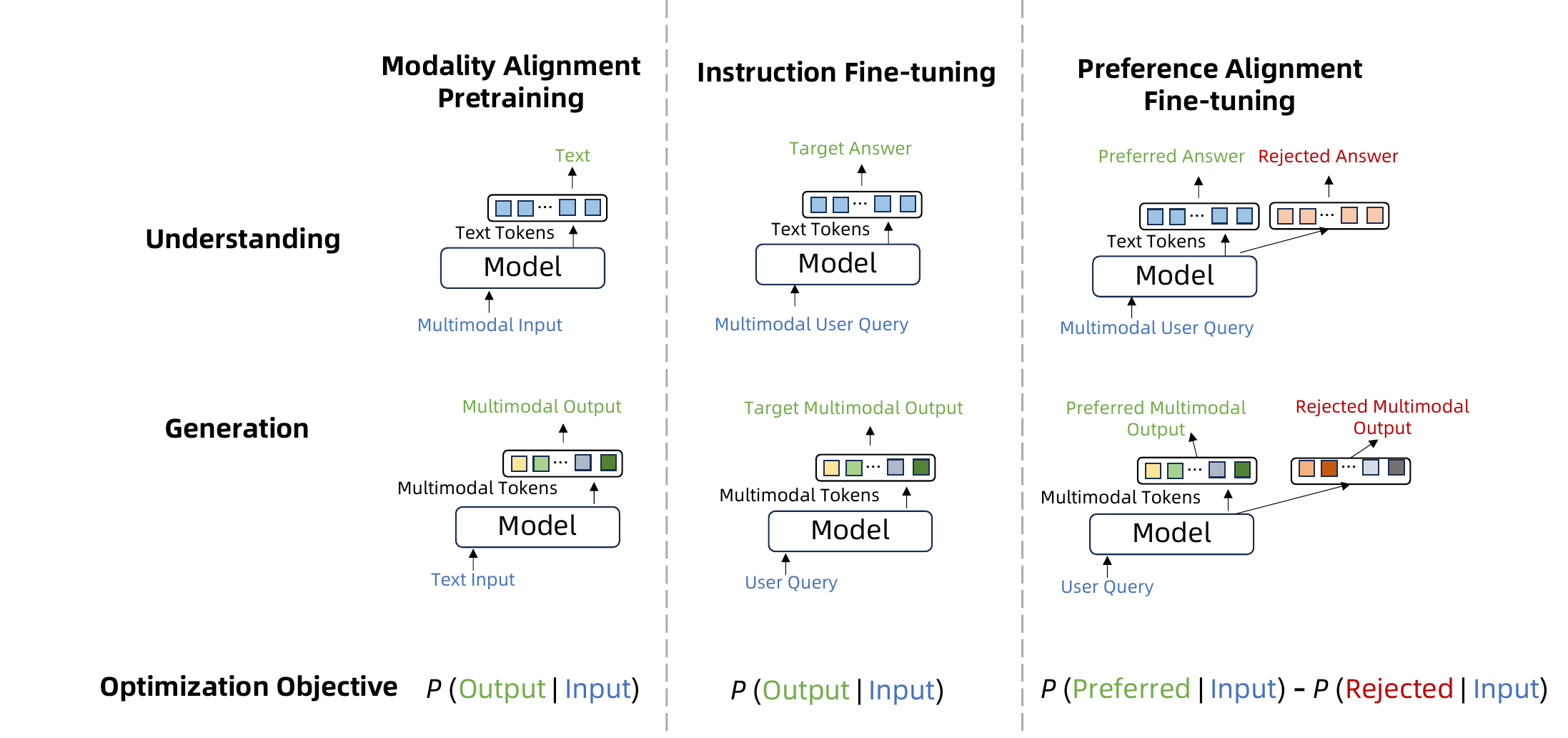}
    \caption{Training stage overview. }
    \label{fig:training_stage}
\end{figure*}
Modality alignment is a critical process that endeavors to represent inputs from diverse modalities within a shared space for subsequent processing. 
Given the inherent differences in the nature of various modalities, dedicated encoders tailored to each modality transform raw inputs into a vector representation, which is then aligned in the shared space. 
For instance, the alignment training of vision-language models typically occurs on a large-scale corpus $\mathcal{C} = \{(C, I)\}$ comprising image-text pairs with the image denoted as $I$ and its corresponding caption as $C$.
The modality alignment objective typically adheres to a conditional language modeling format, expressed as:
\begin{equation}
    L(\theta_\mathcal{M}) = f\left( y_i, p_{\theta}\left(x_i \mid x_{1\sim i-1}, I\right)\right),
\end{equation}
where the parameter of the modality encoder module $\theta_\mathcal{M}$—such as a CLIP vision encoder responsible for mapping multi-modal inputs into vectors in the shared space—is exclusively trained to enhance stability.

It is noteworthy that the modality condition $I$ for images can be seamlessly adapted to other modalities, such as videos and audios, with corresponding training corpora like WebVid~\citep{webvid} for video-text alignment and Clotho~\citep{drossos2020clotho} for audio-text alignment, CroCo~\citep{croco} for 3D views and embodiment Habitat~\citep{habitat}.
Besides, it is also possible that the text and the image are interleaved with each other, and the objective can be adjusted accordingly~\citep{awadalla2023openflamingo,laurencon2023obelics}.
We provide a comprehensive list of modality alignment training in the later section~(\S~\ref{subsec:pretrain_dataset}).
\subsubsection{Modality Alignment in Generation}
\label{sec: alignment generation}

The alignment objective can be easily adapted to the generative scenarios by replacing the one-hot word index $y_i$ with corresponding modality tokens, which might be learned via a pre-defined codebook or optimized via regression. 
Take the traditional text-to-image task as an example, given a description-image pair  $(C, I)$, 
the alignment objective becomes:
\begin{equation}
        L(\theta_\mathcal{M}) = f\left( y_i, p_{\theta}\left(t_i \mid t_{1\sim i-1}, C\right)\right).
\end{equation}
In DTM,
the $y_i$ could be a targeted discrete visual token learned via an off-shelf model such as VQGAN, and the image content would be reconstructed by mapping the token back to the image space via the codebook.
In CSM, the $y_i$ is instead a contiguous modality vector that can be further decoded by a decoder to produce the image pixels~\citep{sun2024emu}.
Besides, the objective can also be implemented in a span corruption style for a better reconstruction of specific modalities~\citep{lu2022unifiedio}.

Given that a primary objective in the alignment stage is to harmonize the semantics of concepts expressed across different modalities, comprehensive coverage of the training corpus becomes imperative. Consequently, the alignment training is often performed on web-scale datasets. For example, the visual-text alignment is usually conducted on up to millions and even billions of pairs on Laion400M~\citep{laion400m} and Laion5B~\citep{laion5b}.



\subsection{Finetuning: Instruction and Preference}
\label{subsub-sl}
After modality alignment training, LMMs acquire a foundational understanding of the semantics associated with various modalities in a unified semantic space. 
To further enhance LMMs' ability to comprehend and perform complex user queries, such as image understanding and generation, researchers employ \emph{instruction tuning} on meticulously curated datasets. Subsequently, \emph{preference alignment training} is utilized to refine model behaviors with implicit human preferences and address potential issues that may have emerged during earlier training phases. 
In the following discussion, we will discuss recent advancements in instruction tuning (\S\ref{subsec:sft_understanding} and \S\ref{subsec:sft_generation}) and alignment training (\S\ref{subsec:alignment_understanding} and \S\ref{subsec:alignment_gen}), as well as explore promising avenues for future research in these domains.

\subsubsection{Instruction Tuning in Understanding}
\label{subsec:sft_understanding}
After the modality alignment training, different modality inputs now can be represented in a unified embedding space for the backbone LLM to perform complex tasks. Instruction tuning~(\emph{alias} supervised fine-tuning) plays a crucial role in activating this potential of multi-modal language models.
Specifically, the instruction tuning aims to improve the model's ability to satisfy user queries.
Again, take the vision language models as an example.
The visual instruction tuning involves training the model on a 
dataset that usually consists of a multi-modal triplet $(I, Q, A)$ of an image $I$, a
user query $Q$, and a desired response $A$. This still can be achieved by the previous training object:
\begin{equation}
    L(\theta) = f\left( A, p_{\theta}\left(x_i \mid x_{1\sim i-1}, I\right)\right).
\end{equation}
Different from the previous alignment training, the instruction tuning stage involves a more challenging objective to reason over the modalities, motivating the model to explore the inner interaction between different modalities to increase the likelihood of the preferred answers.
It has been shown that the quality of the instruction tuning is the key to the ability~\citep{liu2023llava15}. Pilot studies explore various methods for constructing high-quality instruction tuning datasets such as 
adapting publicly available multi-modal benchmarks~\citep{li2023m3it,visionFlan2023,xu2022multiinstruct}, synthesizing datasets using self-instruction with ChatGPT/GPT-4~\citep{liu2023llava,chen2023sharegpt4v,zhao2023svit,zhao2023chatbridge}. Furthermore, mixing the multi-modal instruction dataset with 
text-only query-response pairs is also shown to be effective for improving the instruction following ability~\citep{xu2022multiinstruct,liu2023llava}.
For 
A curated list of these instruction tuning dataset can also be found in later section.

%



\subsubsection{Instruction Tuning in Generation}
\label{subsec:sft_generation}

Similar to the practice in understanding, the key to improving the generation ability after alignment is collecting high-quality and diverse task datasets, where the reconstruction targets vary according to the task requirements. 
However, most training objectives still fall into the token modeling paradigm with different tokenization schemas. The desired output such as textual sentences, images/videos and audios, is represented in a sequence of $N$ tokens $S = (s_0, \ldots, s_N)$, given the conditioned user queries $Q$ specifying the requirements on the target outputs. During the instruction tuning stage, the following objective is optimized:
\begin{equation}
    L(\theta) = f\left( y_i, p_{\theta}\left(s_i \mid s_{1\sim i-1}, Q\right)\right),
\end{equation}
where $y_i$ would be the corresponding discrete token or contiguous vector processed as in the alignment training objective.
To provide wide coverage of the generation ability, previous work~\citep{lu2022unifiedio} ensembles a massive multi-tasking dataset and the sampling ratio during training would be balanced to better expose the model to underrepresented tasks. AnyGPT~\citep{Zhan2024AnyGPT}
utilizes commercial image-generation and music-generation systems to construct a large-scale high-quality text-to-multimodal instruction tuning datasets.

\subsubsection{Preference Alignment Training in Understanding}
\label{subsec:alignment_understanding}

Despite the progress made by previous training stages, misalignment issues that pose a potential risk of generating misleading content without anchoring to the provided visual context~\citep{li2023hallucinate,2023llavarlhf}, or biased responses against minority groups~\citep{gpt4v}, still exist.
To further align with human preference for LMMs, pilot studies draw insights from LLMs and 
apply alignment techniques such as Reinforcement Learning with Human Feedback (RLHF)~\citep{RLHF} and Direct Preference Optimization (DPO)~\citep{DPO} for LMMs.
LLaVA-RLHF~\citep{2023llavarlhf} first explores the RLHF for VLM, by training a factuality-oriented reward model on a synthesized dataset to guide the VLM to produce outputs that anchor with the visual context better.
Formally, let $x$ be a prompt containing both images and text inputs, and $y_i$ denotes the corresponding response generated by model $\pi_i$. The RLHF process can be formulated as: 
\begin{equation*}
    \max _{\pi_\theta} \mathbb{E}_{x \sim \mathcal{D}, y \sim \pi_\theta(y \mid x)}\left[r(x, y)\right]-\beta \mathbb{D}_{\mathrm{KL}}\left[\pi_\theta(y \mid x) \| \pi_{\mathrm{ref}}(y \mid x)\right],
\end{equation*}
where $r$ is the reward model and the KL term penalizes deviations of the current model $\pi_{\theta}$ from the initial model $\pi_{\mathrm{ref}}$. $\beta$ is a hyper-parameter. The RLHF process aims to finetune the model to achieve higher rewards from the reward model, all while preserving the majority of its original knowledge.

As training the reward model can be difficult due to the stability issue, there has been a DPO method to tackle these challenges.
The key insight behind DPO is that the optimal policy $\pi^*$ has a closed-form solution with regard to a reward function $r$ and initial policy $\pi_{\mathrm{ref}}$:
\begin{equation*}
    r(x, y)=\beta \frac{\pi^*(y \mid x)}{\pi_{\mathrm{ref}}(y \mid x)}+\beta \log Z(x),
\end{equation*}
where $Z$ is the partition function.

Under the  Bradley-Terry (BT) preference model~\citep{Bradley1952RankAO}, the objective becomes:
\begin{equation}
\label{eq:dpo}
    \max _{\pi_\theta} \mathbb{E}_{\left(x, y_w, y_l\right) \sim \mathcal{D}} \log \sigma\left(\beta \log \frac{\pi_{\theta}\left(y_w \mid x\right)}{\pi_{\mathrm{ref}}\left(y_w \mid x\right)}-\beta \log \frac{\pi_{\theta}\left(y_l \mid x\right)}{\pi_{\mathrm{ref}}\left(y_l \mid x\right)}\right),
\end{equation}
where $\sigma$ denotes the sigmoid function.
RLHF-V\citep{Yu2023RLHFVTT}  collects human preference in the form of segment-level corrections on hallucinations, and performs dense direct preference optimization
over the human feedback.
\citet{2023vlfeedback} build VLFeedback by annotating the preference with GPT-4V models and applies DPO on Qwen-VL-Chat showing clear advantages.

\subsubsection{Preference Alignment Training in Generation}
\label{subsec:alignment_gen}

Due to the computation cost and the difficulty of collecting large-scale comparison datasets (i.e., creating slightly different images), there are few explorations on preference alignment in generative unified multimodal models.
There are pilot studies investigating preference alignment for diffusion models, where the expected reward of a generated sequence $\boldsymbol{x}_{1:T}$ given a condition $\boldsymbol{c}$ and initial latent $\boldsymbol{x}_0$ is:
\begin{equation}
r\left(\boldsymbol{c}, \boldsymbol{x}_0\right)=\mathbb{E}_{p_\theta\left(\boldsymbol{x}_{1: T} \mid \boldsymbol{x}_0, \boldsymbol{c}\right)}\left[R\left(\boldsymbol{c}, \boldsymbol{x}_{0: T}\right)\right]
\end{equation}
Similar to the alignment training in understanding tasks, the objective is to maximize the expected reward while minimizing the KL divergence between the learned distribution $p_\theta$ and a reference distribution $p_\text{ref}$:
\begin{equation}
\begin{aligned}
& \max _{p_\theta} \mathbb{E}_{\boldsymbol{c} \sim \mathcal{D}_c, \boldsymbol{x}_{0: T} \sim p_\theta\left(\boldsymbol{x}_{0: T} \mid \boldsymbol{c}\right)}\left[r\left(\boldsymbol{c}, \boldsymbol{x}_0\right)\right] \\
&-\beta \mathbb{D}_{\mathrm{KL}}\left[p_\theta\left(\boldsymbol{x}_{0: T} \mid \boldsymbol{c}\right) \| p_{\mathrm{ref}}\left(\boldsymbol{x}_{0: T} \mid \boldsymbol{c}\right)\right]
\end{aligned}
\end{equation}
Current methods for aligning image generative models mainly adopt DPO to bypass the cumbersome reward modeling process. 
\citet{diffusion_dpo1} re-formulate DPO to account for the intractable likelihood in diffusion models, where the evidence lower bound (ELBO) is employed to derive a differentiable objective function for optimization.
The final DPO-Diffusion loss function encourages the model to improve the denoising process for preferred images more than for non-preferred images.  
\begin{equation}
\begin{aligned}
L_{\text {DPO-Diffusion }}(\theta) &= -\mathbb{E}_{\left(\boldsymbol{x}_0^w, \boldsymbol{x}_0^l\right) \sim \mathcal{D}, t \sim \mathcal{U}(0, T),
\boldsymbol{x}_{t-1, t}^w \sim p_\theta\left(\boldsymbol{x}_{t-1, t}^w \mid \boldsymbol{x}_0^w\right), 
\boldsymbol{x}_{t-1, t}^l \sim p_\theta\left(\boldsymbol{x}_{t-1, t}^l \mid \boldsymbol{x}_0^l\right)} \\
& \log \sigma\left(\beta T \log \frac{p_\theta\left(\boldsymbol{x}_{t-1}^w \mid \boldsymbol{x}_t^w\right)}{p_{\mathrm{ref}}\left(\boldsymbol{x}_{t-1}^w \mid \boldsymbol{x}_t^w\right)}-\beta T \log \frac{p_\theta\left(\boldsymbol{x}_{t-1}^l \mid \boldsymbol{x}_t^l\right)}{p_{\mathrm{ref}}\left(\boldsymbol{x}_{t-1}^l \mid \boldsymbol{x}_t^l\right)}\right),
\end{aligned}
\end{equation}
where condition $\mathbf{c}$ is omitted for brevity.
The models are trained  on the Pick-a-Pic~\citep{kirstain2024pick} dataset, which contains pairwise preferences for images generated by SDXL-beta and
Dreamlike, a fine-tuned version of Stable Diffusion 1.5.
D3PO~\citep{yang2023d3po} instead treats diffusion generation as the multi-step decision problem. Under mild assumptions, the model is trained by the preference objective at the image segment level. 
The human annotators are asked about the final image quality and D3PO assumes that any state-action
pair of the preferred image is better than that of the rejected image.

\subsection{Inference: Enhancing Multimodal Task Performance via Prompt Engineering}
\label{sub:inference}

\begin{figure}[h]
\centering
\includegraphics[width=\textwidth]{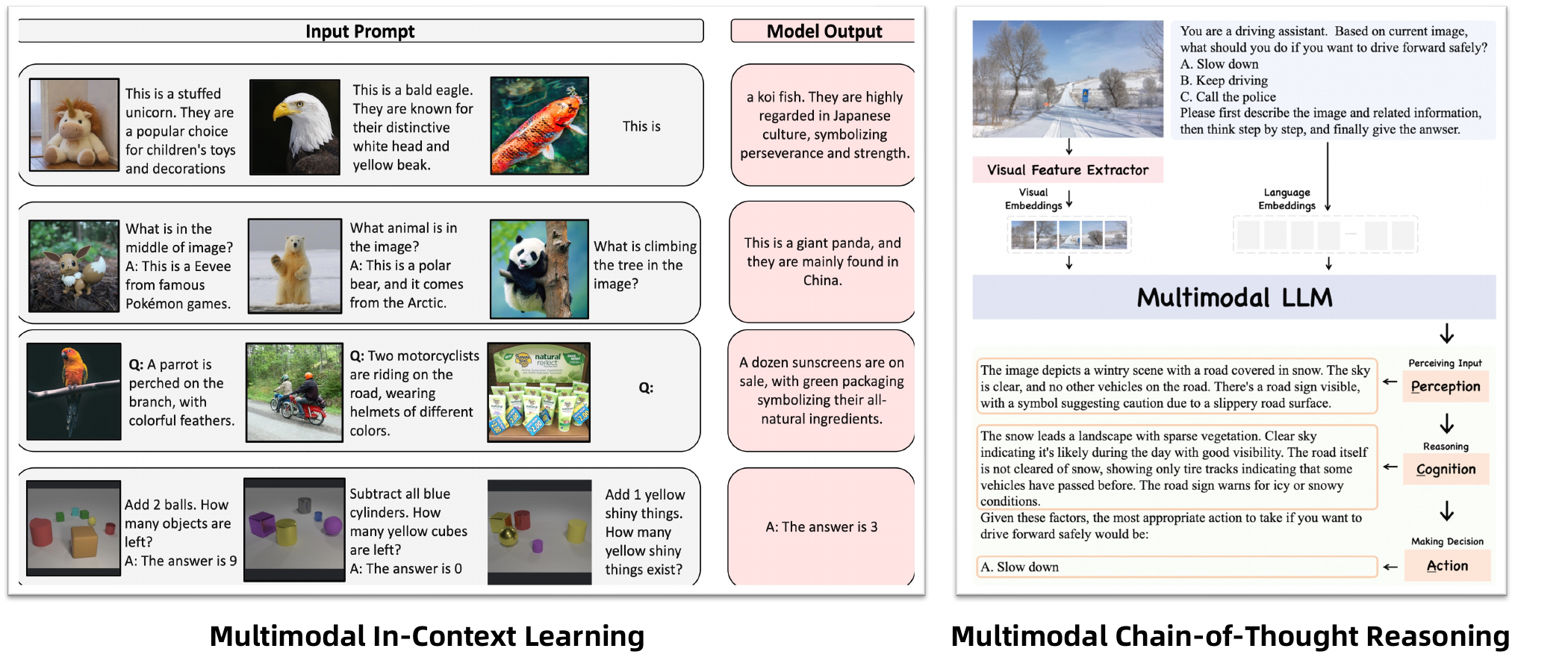}
\caption{Examples of Multimodal In-Context Learning and Chain-of-Thought reasoning.}
\label{fig:inference_examples}
\end{figure}

\begin{figure}[h]
\centering
\includegraphics[width=\textwidth]{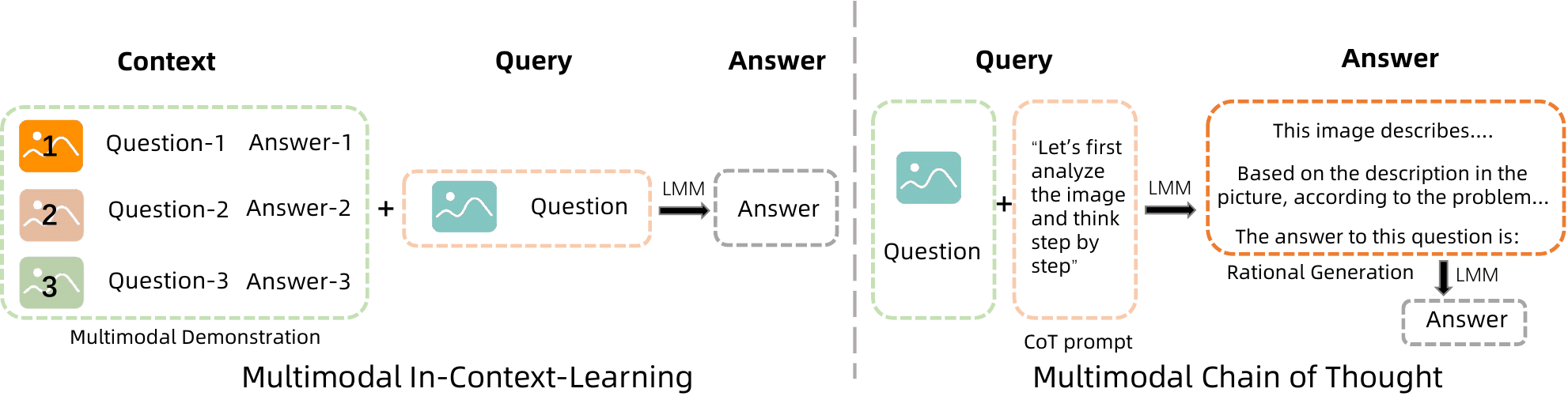}
\caption{Multimodal inference enhancement methods.}
\label{fig:inference_methods}
\end{figure}

\begin{table}[h!]
\centering
\caption{Summary of Multimodal Prompt Engineering research. }
\label{table:multimodal_ICL_summary}
\resizebox{1\textwidth}{!}{
\begin{tabular}{cc|ccc}
\toprule
 Method & Year & Modality & Backbone Model & Task \\
\midrule
\multicolumn{5}{c}{\textbf{\textit{Multimodal ICL}}}\\
\midrule
Frozen~\cite{tsimpoukelli2021multimodal}  & 2021 & Image & GPT2 Architecture~\citep{gpt2} &  Understanding  \\
Flamingo~\cite{alayrac2022flamingo}  & 2022 & Image  & GPT2 Architecture~\citep{gpt2} &  Understanding  \\
MMICL~\citep{zhao2023mmicl} & 2023 & Image  & InstructBLIP~\citep{dai2023instructblip} &Understanding\\
EILeV~\cite{yu2023efficient} & 2023 & Image  & - &  Understanding \\
Open-Flamingo~\cite{awadalla2023openflamingo}  & 2023 & Image  & Flamingo Architecture~\cite{alayrac2022flamingo} &  Understanding \\
LCL~\cite{tai2023linkcontext} & 2023 &  Image & Otter~\citep{li2023mimicit}, OpenFlamingo~\cite{awadalla2023openflamingo} &  Understanding \\
Med-Flamingo~\cite{moor2023medflamingo}  & 2023 & Image  & Open-Flamingo~\cite{awadalla2023openflamingo} &  Understanding \\
MIMIC-IT~\cite{li2023mimicit}  & 2023 & Image & OpenFlamingo~\cite{awadalla2023openflamingo} &  Understanding \\
LVM~\cite{bai2023sequential} & 2023 & Image & LLaMA Architecture~\cite{touvron2023llama} & Understanding\& Generation  \\
LWM~\cite{liu2023world}  & 2023 & Image, Video & LLaMA Architecture~\cite{touvron2023llama} & Understanding \& Generation \\
~\citet{yang2024exploring} & 2024 & Image  & Open-
Flamingo~\cite{awadalla2023openflamingo}  &  Understanding \\
VisualICL~\cite{zhou2024visual} & 2024 & Image  & LLaVA~\citep{liu2023llava} & Understanding \\
Many-Shots ICL~\citep{jiang2024manyshotincontextlearningmultimodal} & 2024 & Image & GPT4-o~\citep{gpt4o}, Gemini1.5~\citep{team2024gemini} & Understanding \\

CoBSAT~\cite{zeng2024mllms}  & 2024 & Image & Emu~\citep{sun2023emu1} & Generation \\
Video ICL~\citep{zhang2024videoincontextlearning} & 2024 & Video & LLaMA Architecture~\citep{touvron2023llama} & Generation \\

Emu~\cite{sun2024emu}  & 2024 &  Image, Video & LLaMA~\cite{touvron2023llama} & Understanding \& Generation \\
Emu2~\cite{sun2024generative} & 2024 & Image, Video &  LLaMA-33B~\cite{touvron2023llama} & Understanding \& Generation  \\

Yang et al.~\citep{sheng2024unified} & 2024 & Image & GPT2 Architecture~\citep{gpt2} & Understanding \& Generation  \\

VALL-E~\cite{wang2023neural} & 2023 & Audio &-& Generation \\

MELLE~\cite{meng2024autoregressive} & 2024 & Audio &-& Generation\\

Seed-TTS~\cite{anastassiou2024seed} & 2024 & Audio &-& Generation\\

Audio Flamingo~\cite{kong2024audio} & 2024 & Audio &OPT-IML-MAX-1.3B~\cite{iyer2022opt}& Understanding\\ 

Moshi~\cite{defossez2024moshi} & 2024 & Audio &Helium~\cite{defossez2024moshi}& Understanding \& Generation \\

\midrule
\multicolumn{5}{c}{\textbf{\textit{Multimodal CoT}}}\\
\midrule
MM-CoT~\citep{zhang2023multimodal}  & 2023 & Image & T5-770M~\citep{raffel2020exploring} &  Understanding  \\
DDCoT~\citep{zheng2023ddcot}  & 2023 & Image & ChatGPT~\citep{ouyang2022training}/GPT-3~\citep{gpt3} &  Understanding  \\
VCDM~\citep{harvey2023visual}  & 2023 & Image &  Stable Diffusion~\citep{diffusion_dpo1} &   Generation   \\
V*~\citep{wu2024v}  & 2024 & Image & Vicuna-7B~\citep{vicuna2023} &  Understanding  \\
CogCoM~\citep{qi2024cogcom}  & 2024 & Image & Vicuna-7B~\citep{vicuna2023} &  Understanding  \\
VisualCoT~\citep{shao2024visual}  & 2024 & Image & Vicuna-7B/13B~\citep{vicuna2023} &  Understanding  \\
CCoT~\citep{Mitra_2024_CVPR}  & 2024 & Image & - &  Understanding  \\
VideoCoT~\citep{wang2024videocot}  & 2024 & Video & - &  Understanding  \\
VoT~\citep{fei2024videoofthought}  & 2024 & Video & Vicuna-7B~\citep{vicuna2023} &  Understanding  \\
WavLLM~\citep{hu2024wavllm}  & 2024 & Audio & LLaMA Architecture~\citep{touvron2023llama2} & Understanding\\
SpeechVerse~\citep{das2024speechverse}  & 2024 & Audio &  Flan-T5-XL~\citep{JMLR:v25:23-0870} & Understanding\\
CoT-ST~\citep{du2024cot}  & 2024 & Audio & - &  Understanding\\
AST-CoT~\citep{hu2024chain11}  & 2024 & Audio & T5~\citep{raffel2020exploring} &  Understanding\\
\bottomrule
\end{tabular}}
\end{table}

After the pretraining and finetuning stages, MMNTP models can also benefit from prompt engineering techniques, much like LLMs. Stemming from research in prompt engineering \citep{vatsal2024surveypromptengineeringmethods}, In-Context Learning (ICL) \citep{icl_survey} and Chain-of-Thought reasoning (CoT) \citep{wei2022chain} are key methods that significantly enhance the performance of LLMs on complex tasks, such as mathematical reasoning \citep{cobbe2021trainingverifierssolvemath}. 
As illustrated in Fig.~\ref{fig:inference_methods}, ICL adds few-shot examples in the prompt of LMM to guide and improve models' performance on unseen examples. 
CoT guides the model to articulate step-by-step reasoning processes. 

Although prompt engineering techniques have had huge success in LLMs~\citep{vatsal2024surveypromptengineeringmethods}, their application in multimodal remains largely underexplored so far. Table~\ref{table:multimodal_ICL_summary} lists the related work on multimodal ICL and CoT research.

\subsubsection{Multimodal In-Context Learning}
Multimodal In-Context Learning (ICL) is an emerging paradigm in which models leverage a few demonstration examples incorporating visual, textual, and other optional modalities to perform multimodal tasks. In this learning paradigm, the input processed by the Large Multimodal Model is divided into two components: the query \( x_q \) and the context \( C \). The LMM needs to generate a sequence of tokens as outputs \(y_q\) based on these two parts:
\begin{equation}
y_q = LMM(x_q, C)
\end{equation}
The context \( C \) consists of a set of input-output ICL examples:
\begin{equation}
C = \{(x_i, y_i)\}_{i=1}^n 
\end{equation}
Adopting the notation from Todd et al. \cite{todd2024function}, we represent the generic template for organizing the context \( C \) as follows:  
\begin{equation}
 Q:\{x_1\}\textbackslash n \textbf{ }A:\{y_1\} \textbackslash n \textbackslash n \textbf{ }\dots Q:\{x_n\} \textbackslash n \textbf{ }A:\{ y_n\},
\end{equation}
where \( Q \) and \( A \) symbolize the question and answer template structures respectively, and \( x_i \) and \( y_i \) denote the question and answer of the \( i \)-th demonstration respectively.

Multimodal ICL introduces unique challenges compared to unimodal ICL, particularly in integrating and aligning diverse modalities such as text, images, and videos~\citep{shukor2023beyond}~\citep{zhao2023mmicl}\citep{baldassini2024makes}. In multimodal ICL, both the query \( x_q \) and context \( x_i \) may vary in modality, conveying complementary yet distinct information that can lead to imbalanced or inefficient learning. A primary challenge, as noted in recent studies \citep{awadalla2023openflamingo} \citep{baldassini2024makes}, is that performance in many multimodal ICL systems remains largely text-driven, with other modalities—such as images or videos—contributing minimally to overall task performance.

To address this challenge, several approaches~\citep{awadalla2023openflamingo,yu2023efficient,yu2023efficient,zhao2023mmicl} focus on enhancing the model's ability to generalize across diverse multimodal tasks. EILEV~\citep{yu2023efficient} proposes new training methods for video understanding. MMICL~\citep{zhao2023mmicl} and CoBSAT~\citep{zeng2024mllms} use specialized datasets and prompt engineering to enhance multimodal reasoning. Recent work further extends these efforts by exploring large-scale models for more effective in-context learning with interleaved multimodal inputs, ~\citep{EMU2,liu2023world,EMU2,laurenccon2024obelics}.

%
%
%
%

\subsubsection{Multimodal Chain-of-Thought Prompting}
Multimodal Chain-of-Thought (CoT) is a method that enables models to perform complex reasoning and decision-making in a multimodal setting through step-by-step derivation and coherent thinking. Pioneered by~\citet{zhang2023multimodal}, MM-CoT introduces Chain-of-Thought prompting into visual domains, raising the challenge of labor-intensive annotation, as multimodal data often demands expensive and complex human-labeled information. MM-CoT employs ScienceQA~\citep{lu2022scienceqa}, a dataset focused on scientific questions involving multiple modalities with annotated rationales, while VoT~\citep{fei2024videoofthought} tackles the annotation challenge in video tasks by combining machine and human expertise through active learning.

Another challenge lies in mitigating language hallucinations~\citep{alayrac2022flamingo,ji2023survey,maynez2020faithfulness,rawte2023survey,zhang2023sirenssongaiocean,chen2024pcabench,zhao2024lookingtextreducinglanguage}, which are exacerbated due to the lack of necessary and fine-grained visual context when multimodal information is provided simultaneously. To better inject visual information, V*~\citep{wu2024v} addresses this by dynamically focusing on key visual regions, ensuring that visual details are accurately attended to, particularly in high-resolution images. CCoT~\citep{Mitra_2024_CVPR} generates scene graphs instead of simple captions, explicitly reasoning over visual features to avoid misinterpretation. Moreover, DDCoT ~\citep{zheng2023ddcot} introduces a new CoT prompting method that divides the roles of reasoning and visual recognition between language and visual models, thereby enhancing reasoning clarity and reducing hallucinations.

Subsequent work~\citep{wang2024videocot}~\citep{fei2024videoofthought}~\citep{du2024cot}~\citep{raffel2020exploring} has extended the method beyond images to include video and audio. For instance, the CoT-ST~\citep{du2024cot}framework adapts chain-of-thought reasoning for speech translation, breaking the process into distinct steps to improve accuracy and fluency. Video-CoT~\citep{wang2024videocot}  focus on complex video reasoning, aiming to achieve human-level video comprehension.

\section{Datasets and Evaluation}
\label{sec:datasets}
In this section, we delve into several crucial aspects of training and evaluating MMNTP models. The subdivision begins with an exploration of the training datasets (Section~\ref{sec: training_dataset}), categorized into pre-training and fine-tuning datasets. The pre-training datasets are further divided based on modality into text-only, image-based, video-based, and audio-based data, which are essential for modality alignment and the establishment of a unified multimodal representation. Following this, fine-tuning datasets are described, focusing on their specific applications in multimodal understanding and multimodal generation tasks.

Additionally, we discuss the evaluation of MMMNTP models (Section~\ref{sec: evaluation_dataset}), which is pivotal in measuring their effectiveness and capability across various modalities. This aspect is divided into holistic evaluation and emerging evaluation benchmarks. Holistic evaluation benchmarks, such as MME~\citep{fu2023mme} and SEED-Bench~\citep{li2023seedbench}, comprehensively assess the integration and interplay between different modalities like image, text, and video. Emergent benchmarks, including SparklesEval~\citep{huang2023sparkles} and HallusionBench~\citep{guan2024hallusionbench}, push the boundaries further by testing specialized capabilities like conversational competence, mathematical reasoning, and mitigation of hallucinations in model outputs. 

\subsection{Training Datasets}
\label{sec: training_dataset}
Depending on the stage of training, we categorize data into pre-training data and fine-tuning data. Pre-training data can be classified into uni-modal data and multimodal data based on modality. Fine-tuning data is categorized based on its usage scenario into multimodal understanding data and multimodal generation data.\subsubsection{Pre-training Datasets}
\label{subsec:pretrain_dataset}
Unlike large language models that are pre-trained only on pure text data, multimodal models require pre-training on a variety of different modalities of data, which demands a significant quantity and diversity of multimodal data. In this section, we briefly summarize several multimodal datasets widely used for training multimodal models. Based on the type of modality, we categorize these data into four groups: Text-Only, Image-Based, Video-Based, and Audio-Based.
\paragraph{Text-Only} Although pure text data is commonly utilized in language models, it also plays a crucial role in enhancing the language expression and reasoning abilities of multimodal models. For this purpose, pure text data is integrated into the pre-training corpus. One of the most extensively used datasets in this context is C4~\cite{habernal2016c4corpus}, a filtered open-source dataset derived from web crawls. Its multilingual variant, mC4~\cite{xue2021mt5}, encompasses natural text in 101 languages, sourced from the public Common Crawl web archive. Additionally, the Wikipedia dataset~\cite{guo2020wiki}, which consists of cleaned articles in multiple languages, is created from language-specific segments of the Wikipedia dump. Another significant contribution to this field is The Pile, an expansive and diverse open-source dataset for language modeling. Amassing a total of 825 GiB, The Pile~\cite{gao2020pile} is an amalgamation of 22 distinct, high-quality smaller datasets, providing a rich resource for language model pretraining. Recently, RedPajama~\cite{together2023redpajama}, an open dataset with 30 trillion tokens for training large language models, has also been introduced, contributing significantly to the resources available for developing advanced language models. Furthermore, FineWeb~\cite{penedo2024finewebdatasetsdecantingweb} release a new, large-scale (15-trillion tokens) dataset for LLM pretraining. FineWeb is derived from 96 CommonCrawl snapshots and produces better-performing LLMs. Dolma~\citep{soldaini2024dolmaopencorpustrillion} is a high-quality open dataset from a diverse mix of web content, academic publications, code, books, and encyclopedic materials, covering 3T tokens.

\paragraph{Image-Based} Multimodal data is key for models to perform modality alignment, that is, to map different modal representations into a unified space. CLIP~\citep{radford2021clip} was developed using 400 million image-text pairs sourced from the internet. Subsequent models like ALIGN~\citep{Jia2021ALIGN}, BASIC~\citep{pham2023combined}, and Florence~\citep{yuan2021florence} were trained on even larger and more diverse datasets with noisier image-text pairs. However, the majority of these extensive datasets remain inaccessible to the public. In the academic community, researchers recommend using several million image-text pairs for multimodal model pre-training, including CC12M~\citep{changpinyo2021conceptual}, RedCaps~\citep{desai2021redcaps}, YFCC~\citep{thomee2016yfcc100m}, WIT~\citep{srinivasan2021wit}, and Capsfusion~\cite{yu2023capsfusion}. Publicly accessible datasets of a relatively smaller scale include SBU~\cite{ordonez2011im2text}, MSCOCO~\cite{lin2014microsoft}, VG~\cite{krishna2017visual_genome}, and CC3M~\citep{sharma2018conceptual}. Among the larger-scale image-text datasets available to the public are FILIP~\citep{filip}, LAION-400M~\citep{laion400m}, COYO-700M~\citep{kakaobrain2022coyo-700m}, SA-1B~\citep{wang2023all}, and LAION-5B~\citep{laion5b}, among others. Additionally, some studies have emphasized the importance of data quality in building robust multimodal models, such as DataComp~\citep{gadre2023datacomp}, Shutterstock~\citep{nguyen2022quality}, and ShareGPT4V~\citep{chen2023sharegpt4v}. 
Beyond sourcing image-text data from the web, there has been a growing interest in compiling datasets that interleave images and text, a concept pioneered by the M3W~\citep{alayrac2022flamingo} dataset featured in Flamingo~\citep{alayrac2022flamingo}. Notable examples of such datasets are MMC4~\citep{zhu2023multimodal} and OBELISC~\cite{laurenccon2023obelisc}. Additionally, there's an emerging trend in research to focus on the extraction and association of text segments in captions with specific areas in images, leading to the formation of grounded image-text pairs. Datasets like GRIT-20M~\citep{peng2023kosmos} and CapsFusion-grounded~\citep{sun2023generative} exemplify this methodology.

\paragraph{Video-Based} 
MSR-VTT~\citep{xu2016msrvtt} features 10K diverse web video clips and 200K clip-sentence pairs spanning a wide range of categories. HowTo100M~\citep{miech2019howto100m} expands this landscape with 1.22 million YouTube videos on topics like cooking and crafting, enriched with subtitles from ASR systems or manual input. ACAV100M~\citep{lee2021acav100m} provides a vast 100 million video library, ideal for self-supervised learning with high audio-visual correspondence. WebVid~\citep{webvid} enhances video data with manually crafted, accurate captions. Ego4D~\citep{grauman2022ego4d} offers an extensive collection of diverse egocentric video footage for research. HD-VILA~\citep{xue2022advancing} introduces a high-resolution video-language dataset with varied content. YT-Temporal~\cite{zellers2022merlot}, sourced from public YouTube videos, focuses on broadening understanding of objects, actions, and scenes. VideoCC3M~\cite{nagrani2022learning} utilizes a new pipeline to transfer image captions to videos without extra manual labor. Youku-mPLUG~\citep{xu2023youku} has released the largest public Chinese video-language dataset, prioritizing safety, diversity, and quality. Most recently, InternVid~\citep{wang2023internvid} demonstrates a scalable method for building high-quality video-text datasets using large language models, effectively enhancing video language representation learning.
\paragraph{Audio-Based}
Audio-based pretraining datasets can be primarily categorized into three types: speech pretraining datasets, music pretraining datasets, and general audio pretraining datasets.
Librilight~\cite{kahn2020libri} includes more than 60k hours unlabeled speech data and is widely used by audio pretraining~\cite{wang2023neural,zhang2024speechlm}.
Libriheavy~\cite{kang2024libriheavy} introduces a refined pipeline for audio alignment and segmentation and detailed annotations with punctuation and capitalization, reflecting more natural speech patterns, to the mostly unlabeled Librilight.
Wenetspeech~\cite{zhang2022wenetspeech} is the largest Mandarin speech pretraining corpus, collecting over 22,400 hours of audio, with 10,000+ hours of high-quality labeled speech, 2,400+ hours of weakly labeled speech, and roughly 10,000 hours of unlabeled speech from diverse sources such as YouTube and podcasts. 
Yodas~\cite{li2023yodas} offer over 500,000 hours of speech data in more than 100 languages, significantly benefiting the multilingual nature of the audio pretrain community.
Other widely-used speech pretraining datasets include librispeech~\cite{panayotov2015librispeech}, libritts~\cite{zen2019libritts} and gigaspeech~\cite{chen2021gigaspeech}. 
Music pretraining is a growing research area~\cite{dhariwal2020jukebox,zhu2021musicbert,li2022map,li2023mert,lu2023musecoco,hussain2023m,qu2024mupt}.
Million Song Dataset (MSD)~\cite{bertin2011million} is one of the largest publicly available collections of audio features and metadata for a million contemporary popular music tracks.
FMA (Free Music Archive) Dataset~\cite{defferrard2016fma} is a well-curated collection of over 100,000 tracks from various artists and genres available under Creative Commons licenses.
Other widely-used music pretraining datasets include disco10m~\cite{lanzendorfer2024disco}, mtg-jamendo~\cite{bogdanov2019mtg}, and Lp-musiccaps~\cite{doh2023lp}.
General audio pretraining datasets, including wavcaps~\cite{mei2023wavcaps}, audioset~\cite{gemmeke2017audio}, vggsound~\cite{chen2020vggsound}, and clotho~\citep{drossos2020clotho}, mainly focus on boosting the performance of localizing audio-visual correspondence and audio-text intermodal translation tasks (not speech-to-text).
\begin{table*}[htb]
    
    \caption{Statistics of commonly-used Pre-training data.}
    \resizebox{0.85\textwidth}{!}{
    \begin{tabular}{lcccr}
    \toprule
    \textbf{Datasets} & \textbf{Tags}  & \textbf{Doc/Img/Vid/Aud} & \textbf{Source} & \textbf{Time}  \\
    \midrule
    C4~\cite{habernal2016c4corpus} & Text-Only & 8.2M/-/-/- & CommonCrawl & Apr-2019  \\
    mC4~\cite{xue2021mt5}& Text-Only & 2.1M/-/-/- & CommonCrawl & Oct-2020 \\
    Pile~\cite{gao2020pile} & Text-Only  & 211M/-/-/- & Other & Dec-2020 \\
     Wikipedia~\cite{guo2020wiki} &  Text-Only  & 13.4M/-/-/- & Wikipedia & Mar-2023  \\
     RedPajama~\cite{together2023redpajama} & Text-Only & 100B/-/-/- & CommonCrawl  & Oct-2023 \\

     Dolma~\citep{soldaini2024dolmaopencorpustrillion} &  Text-Only & 4.4B/-/-/- & Common Crawl, GitHub, Reddit, ...  & Jan-2024 \\
     FineWeb~\cite{penedo2024finewebdatasetsdecantingweb} & Text-Only & 22.7B/-/-/- & CommonCrawl & May-2024 \\
    \midrule 
    SBU~\cite{ordonez2011im2text}  & Image-Based   & 1M/1M/-/-  & Flickr & Dec-2011  \\
    YFCC~\cite{thomee2016yfcc100m}  & Image-Based   &  100M/99.2M/0.8M/-  & Flickr & Jan-2016  \\
    MS-COCO~\cite{lin2014mscoco}   & Image-Based   & 1M/200K/-/-  & HumanCurated & Jul-2018 \\
    VG~\cite{krishna2017visual_genome}  & Image-Based & 5.4M/108K/-/- & HumanCurated & Feb-2016 \\
    CC3M~\cite{sharma2018conceptual}  & Image-Based   & 3.3M/3.3M/-/- & Web Crawl & Jul-2018 \\
    CC12M~\cite{changpinyo2021conceptual} & Image-Based & 12M/12M/-/-  & Web Crawl & Feb-2021 \\
    WIT~\cite{srinivasan2021wit}  & Image-Based  &  37.6M/11.5M/-/- & Wikipedia & Jul-2021 \\
    RedCaps~\cite{desai2021redcaps}  & Image-Based  & 12M/12M/-/- & Reddit links & Nov-2021 \\
    FILIP300M~\cite{yao2021filip} &	Image-Based & 300M/300M/-/-	& Web Crawl & Nov-2021 \\
    LAION-400M~\cite{laion400m}  & Image-Based  & 400M/400M/-/- &  CommonCrawl  & Nov-2021 \\
    Shutterstock~\cite{nguyen2022quality}  & Image-Based  & 15M/15M/-/-& Shutterstock  & Aug-2022\\
    Coyo-700M~\cite{kakaobrain2022coyo-700m}  & Image-Based  & 747M/747M/-/- & CommonCrawl  & Aug-2022 \\
    Laion-5B~\cite{laion5b}  & Image-Based   & 5B/5B/-/-  & CommonCrawl & Oct-2022 \\
    DataComp~\cite{gadre2023datacomp}  & Image-Based   & 1.4B/1.4B/-/- & Web Crawl & Apr-2023 \\
    SA-1B~\cite{wang2023all}  & Image-Based   & 1.1B/11M/-/-  & Photo Company & Aug-2023  \\
    Capsfusion~\cite{yu2023capsfusion}  & Image-Based & 120M/120M/-/- & Other & Oct-2023 \\
    ShareGPT4V~\cite{chen2023sharegpt4v}  & Image-Based  & 1.2M/1.2M/-/- & Other & Nov-2023 \\
    M3W~\cite{alayrac2022flamingo}  & Image-Based (Interleaved)   & -/185M/-/- & Web Crawl & Apr-2022\\
    MMC4~\cite{zhu2023multimodal}  & Image-Based (Interleaved)  & 103M/585M/-/- & Other & Apr-2023 \\
    Obelisc~\cite{laurenccon2023obelisc}  & Image-Based  (Interleaved)  & 141M/353M/-/- & Web Crawl& Jun-2023 \\
    GRIT-20M~\cite{peng2023kosmos} &  Image-Based (Grounded)  & 20M/20/-/- & Other & Jun-2023 \\
    CapsFusion-grounded~\cite{sun2023generative} & Image-Based  (Grounded)  & 100M/100M/-/- & Other & Dec-2023 \\
    \midrule
    MSR-VTT~\cite{xu2016msr}  &  Video-Based  & 200K/-/10K/-  & HumanCurated & Jun-2016 \\
    HowTo100M~\cite{miech2019howto100m} &  Video-Based  &  136M/-/1.2M/-  & Youtube & Jun-2019 \\
    ACAV~\cite{lee2021acav100m}  &  Video-Based  & -/-/100M/100M & Web Crawl & Jan-2021  \\
    WebVid~\cite{webvid}  &  Video-Based  &  10M/-/10M/-  & Stock Footage & Jan-2021 \\
    Ego4D~\cite{grauman2022ego4d}  &  Video-Based  & -/-/-/-  & HumanCurated & Oct-2021  \\
    HD-VILA~\cite{xue2022advancing}  &  Video-Based  &  100M/-/3.3M/-  & YouTube & Nov-2021 \\
    YT-Temporal~\cite{zellers2022merlot}  &  Video-Based  & 1B/-/20M/-  & YouTube & Jan-2022 \\
    VideoCC3M~\cite{nagrani2022learning} & Video-Based  & 10.3M/-/6.3M/- & Other & Apr-2022 \\
    Youku-mPLUG~\cite{xu2023youku} &  Video-Based &  10M/-/10M/-  & Youku & Jun-2023  \\
    InternVid~\cite{wang2023internvid}  &  Video-Based  &  234M/-/7.1M/-  & YouTube  & Jul-2023 \\
    
    \midrule
    Million Song Dataset~\cite{bertin2011million} & Audio-Based & -/-/-/1M & The Echo Nest & Feb 2011\\
    MTT~\cite{law2009evaluation} & Audio-Based & -/-/-/25.8k & Web Crawl & June 2013\\
    LibriSpeech~\cite{panayotov2015librispeech} & Audio-Based & 155.8k/-/-/1k hours & Audio Books & Jun 2015\\
    FMA~\cite{defferrard2016fma} & Audio-Based & -/-/-/106k & Free Music Archive & Dec 2016\\
    Audio Set~\cite{gemmeke2017audio}  & Audio-Based & 2.1M/-/-/2.1M & YouTube  & Mar-2017\\
    LibriTTS~\cite{zen2019libritts} & Audio-Based & 2.4k/-/-/0.58k hours & Audio Books & Apr 2019\\
    MTG-Jamendo~\cite{bogdanov2019mtg} & Audio-Based & -/-/-/55k & Jamendo & Jun 2019\\
    Clotho~\cite{drossos2020clotho} & Audio-Based & 25k/-/-/5k & FreeSound Platform & Oct 2019\\
    Librilight~\cite{kahn2020libri} & Audio-Based & -/-/-/60k hours & Audio Books & Dec 2019\\
    VGGSound~\cite{chen2020vggsound} & Video-Based & 309/-/200k/200k & Web Crawl & Apr 2020\\
    Gigaspeech~\cite{chen2021gigaspeech} & Audio-Based & -/-/-/40k hours & Audio Books & Jun 2021\\
    LAION-Audio-630k~\cite{wu2023large} & Audio-Based & 630k/-/-/630k & Web Crawl & Nov 2021\\
    wenetspeech~\cite{zhang2022wenetspeech} & Audio-Based & -/-/-/22.4k hours & Youtube & Feb 2022\\
    WavCaps~\cite{mei2023wavcaps} & Audio-Based & 400k/-/-/400k & Other & Mar-2023\\
    LP-MusicCaps~\cite{doh2023lp} & Audio-Based & 2.2M/-/-/0.5M & Web Crawl & Jul 2023\\
    LibriHeavy~\cite{kang2024libriheavy} & Audio-Based & 9M/-/-/50k hours & Audio Books & Sep 2023\\
    disco-10m~\cite{lanzendorfer2024disco} & Audio-Based & -/-/-/15.2M & Youtube & 2023\\
    yodas~\cite{li2023yodas} & Audio-Based & -/-/-/500k hours & Youtube & Dec 2023\\
    \bottomrule 
    \end{tabular}}

    \label{tab:data_pt}
\end{table*}
\begin{table*}
    \centering
    \caption{Statistics of commonly-used instruction tuning data.}
    \resizebox{0.85\textwidth}{!}{
    \begin{tabular}{lcccr}
    \toprule
    \textbf{Datasets} & \textbf{Tags} & \textbf{Nums} & \textbf{Source} & \textbf{Time}  \\
    \midrule
    MultiInstruct~\citep{xu2022multiinstruct} & Image+Text & 235K& Existing datasets + Human & Dec-2022 \\
    LLaVA~\citep{liu2023llava} & Image+Text & 158K & COCO + GPT & April-2023 \\
    Mini-GPT4~\citep{zhu2023minigpt4} & Image+Text & 3.5K & CC3M + GPT & April-2023 \\
    LMeye~\citep{li2023lmeye} & Image+Text & 7.3M & Existing datasets + GPT & May-2023 \\
    X-LLM~\citep{chen2023x} & Image+Video+Audio+Text & 10K & Existing datasets + GPT & May-2023 \\
    Video-Chat~\citep{li2023videochat} & Video+Audio+Text & 11K &  WebVid-10M + GPT & May-2023 \\
    PMC-VQA~\citep{zhang2023pmc} & Image+Text & 227K & PMC-OA + GPT & May-2023 \\
    DetGPT~\citep{pi2023detgpt} & Image+Text & 30K & COCO+GPT & May-2023 \\
    GPT4Tools~\citep{yang2024gpt4tools} & Image+Text & 71K & Visual ChatGPT & May-2023 \\
    LLaVA-Med~\citep{li2024llava} & Image+Text & 60K & PubMed +GPT & June-2023 \\
    M$^3$IT~\citep{li2023m3it} & Image+Text & 2.4M & Existing datasets + GPT & June-2023 \\
    MIMIC-IT~\citep{li2023mimicit} & Image+Video+Text & 2.8M & Existing datasets + GPT & June-2023 \\
    Video-ChatGPT~\citep{maaz2023video} & Video+Text & 100K & ActivityNet-200+Human & June-2023 \\
    LAMM~\citep{yin2024lamm} & Image+Text & 196K & Existing datasets + GPT & June-2023 \\
    LLaVAR~\citep{zhang2023llavar} & Image+Text & 422K & LAION-5B + GPT & June-2023 \\
    Macaw-LLM~\citep{lyu2023macaw} & Image+Video+Audio+Text & 119K & Existing datasets + GPT & June-2023 \\
    GAVIE~\citep{liu2023mitigating} & Image+Text & 400K & Existing datasets + GPT & June-2023 \\
    MotionGPT~\citep{jiang2024motiongpt} & Motion+Text & 50K & Existing datasets+Human & July-2023 \\
    PF-1M~\citep{chen2023visual} &  Image+Text & 1M & Existing datasets+GPT & July-2023 \\
    SVIT~\citep{zhao2023svit} & Image+Text & 4.2M &  Existing datasets + GPT & July-2023 \\
    BuboGPT~\citep{zhao2023bubogpt} & Image+Audio+Text & 170K & Existing datasets + GPT & July-2023 \\
    MGVLID~\citep{zhao2023chatspot} & Image+Text & 108K &  Existing datasets + GPT & July-2023 \\
    HalDetect~\citep{gunjal2023detecting} & Image+Text & 16K & COCO+Human & Aug-2023 \\
    StableLLaVA~\citep{li2023stablellava} & Image+Text & 126K & SD+GPT & Aug-2023 \\ 
    Sparkles~\citep{huang2023sparkles} & Image+Text & 6.5K & Existing datasets + GPT & Aug-2023 \\
    LVIS-INSTRUCT4V~\citep{wang2023see} & Image+Text & 220K & Existing dataset+GPT & Nov-2023 \\
    M3DBench~\citep{li2023m3dbench} & Image+Text & 320K & Existing datasets + GPT & Dec-2023 \\
    MMEvol~\citep{luo2024mmevol} & Image+Text & 480K & Existing datasets + GPT & Sept-2024 \\
    InstructPix2Pix~\citep{brooks2023instructpix2pix} & Image Editing & 313K & SD+GPT & Jan-2023 \\
    HIVE~\citep{zhang2023hive} & Image Editing & 1.1M & Existing datasets + SD + GPT & Mar-2023 \\
    MagicBrush~\citep{zhang2024magicbrush} & Image Editing & 10K & Existing datasets + SD + Human & Nov-2023 \\
    HQ-Edit~\citep{hui2024hq} & Image Editing &  200K & SD+GPT & Apr-2024 \\
    UltraEdit~\citep{ultraEdit} & Image Editing &  4.1M & Existing datasets +SD+GPT & June-2024 \\
    \bottomrule
    \end{tabular}}
    \label{tab:data_ins}
\end{table*}

\subsubsection{Fine-tuning Datasets}
\paragraph{Multimodal Understanding}
The inaugural work in applying instruction tuning to the multi-modal domain was presented by MultiInstruct~\citep{xu2022multiinstruct}, which successfully combined multi-modal learning into a single-format benchmark dataset incorporating 62 diverse tasks. Concurrently, LLaVA~\citep{liu2023llava} harnessed the capabilities of the language-centric GPT-4 to generate datasets for multi-modal, instruction-based tasks involving both text and images. MiniGPT-4~\citep{zhu2023minigpt4} precisely assembled a dataset rich in detailed image descriptions to facilitate the convergence of visual and linguistic elements.

Further advancements were marked by LMeye~\citep{li2023lmeye}, MMEvol~\citep{luo2024mmevol}, PF-1M~\citep{chen2023visual}, and SVIT~\citep{zhao2023svit}, which scaled up the magnitude of instruction tuning. The domains of video content were also explored by Video-Chat~\citep{li2023videochat} and Video-ChatGPT~\citep{maaz2023video}, which adapted instruction tuning to this dynamic format. In the specialized medical sector, PMC-VQA~\citep{zhang2023pmc} and LLaVA-Med~\citep{li2024llava} crafted datasets for instruction tuning by leveraging existing medical data repositories.
Object detection tasks were ingeniously integrated into instruction tuning through the efforts of DetGPT~\citep{pi2023detgpt} and MGVLID~\citep{zhao2023chatspot}. GPT4Tools~\citep{yang2024gpt4tools} was developed to enhance open-source large language models (LLMs) by equipping them with the versatility to utilize an array of tools effectively, while M$^3$IT expanded the reach of multi-modal instruction tuning across multiple languages.
Expanding the horizon further, X-LLM~\citep{chen2023x}, MIMIC-IT~\citep{li2023mimicit}, MotionGPT~\citep{jiang2024motiongpt}, Macaw-LLM~\citep{lyu2023macaw}, and BuboGPT~\citep{zhao2023bubogpt} ventured into new modalities, enhancing the scope of instruction tuning. The integration of 3D tasks into this domain was initiated by LAMM~\citep{yin2024lamm} and M3DBench~\citep{li2023m3dbench}, enriching the complexity and applicability of instruction tuning.
Meanwhile, LLaVAR~\citep{zhang2023llavar} leveraged publicly accessible OCR tools to harvest text-rich images from the LAION~\citep{laion400m} dataset, thus enhancing visual instruction tuning processes. To address the phenomenon of hallucinations, HalDetect~\citep{gunjal2023detecting} developed a pioneering multi-modal dataset focused on accurate image descriptions. In the pursuit of robustness, GAVIE~\citep{liu2023mitigating} introduced a mix of positive and negative instructions, fortifying the training for visual instruction tuning.
StableLLaVA~\citep{li2023stablellava} combined the generative prowess of ChatGPT with text-to-image models to produce a versatile and diversified dataset featuring a wide range of image content. Sparkles~\citep{huang2023sparkles} introduced the first machine-generated dialogue dataset tailored for word-level interleaved multi-image and text interactions. The project LVIS-INSTRUCT4V~\citep{wang2023see} capitalized on the improved visual processing strengths of GPT-4 to achieve a higher precision in image detail capture and instruction annotation accuracy. 
\paragraph{Multimodal Generation} Additionally, some instruction-based image editing datasets focus on image generation. A typical dataset is InstructPix2Pix~\citep{brooks2023instructpix2pix}, which initially uses GPT-3~\citep{gpt3} to generate the text for edits, and then utilizes Stable Diffusion~\citep{Rombach_Blattmann_Lorenz_Esser_Ommer_2022} along with Prompt2Prompt~\citep{Hertz2022PrompttoPromptIE} technology to generate the corresponding edited images to construct the dataset. Furthermore, HIVE~\citep{zhang2023hive} introduces a larger number of training triplets and incorporates human ranking results, providing stronger supervision signals for more effective model training. Building on these advancements, MagicBrush~\citep{zhang2024magicbrush} introduces the first large-scale, manually annotated dataset specifically designed for instruction-guided real image editing. Expanding further, HQ-Edit~\citep{hui2024hq} provides a high-quality instruction-based image editing dataset consisting of approximately 200,000 edits. Unlike previous methods that relied on attribute guidance or human feedback to build datasets, HQ-Edit employs a scalable data collection pipeline that leverages advanced foundation models, specifically GPT-4V and DALL-E 3.

\definecolor{benchmark_table_row_color}{HTML}{f7efca}
\begin{table}[t]
  \caption{
    Statistics of Benchmark.
  }
  \centering
  \resizebox{0.85\textwidth}{!}{
    \begin{tabular}{llrrc}
      \toprule
      \textbf{Benchmark}                            & \textbf{Modalities} & \textbf{Samples} & \textbf{Span / Feature} & \textbf{Release Date} \\
      \midrule
      \multicolumn{5}{c}{\textbf{Holistic Evaluation}}                                                                                            \\
      \midrule
      MME \citep{fu2023mme}                         & text, image         & 2,374               & 14 Subtasks             & 2023-06-23            \\
      MMBench \citep{liu2023mmbench}                & text, image         & 4,377               & 20 Dimensions           & 2023-07-12            \\
      SEED-Bench \citep{li2023seedbench}            & text, image         & 19,242              & 12 Dimensions           & 2023-07-30            \\
      MLLM-Bench \citep{ge2023mllmbench}            & text, image         & 420                 & 6 Dimensions            & 2023-11-23            \\
      MMMU \citep{yue2023mmmu}                      & text, image         & 11,550              & 183 Subfields           & 2023-11-27            \\
      MVBench \citep{li2024mvbench}                 & text, video         & 4,000               & 20 Subtasks             & 2023-11-28            \\
      SEED-Bench-2 \citep{li2023seedbench2}         & text, image, video  & 24,000              & 27 Dimensions           & 2023-11-28            \\
      VBench \citep{huang2023vbench}                & text, video         & 1,600               & 16 Dimensions           & 2023-11-29            \\
      CMMMU \citep{zhang2024cmmmu}                  & text, image         & 12,000              & 30 Subjects             & 2024-01-22            \\
      \midrule
      \multicolumn{5}{c}{\textbf{Emerging Benchmarks}}   \\
      \midrule
      SparklesEval \citep{huang2023sparkles}        & text, image         & 1,967               & Multi-modal Dialogue    & 2023-08-31            \\
      MathVista \citep{lu2024mathvista}             & text, image         & 6,141               & Math Reasoning          & 2023-10-03            \\
      HallusionBench \citep{guan2024hallusionbench} & text, image         & 1,129               & Hallucination           & 2023-10-23            \\
      Bingo \citep{cui2023holistic}                 & text, image         & 370                 & Hallucination           & 2023-11-06            \\
      MMC-Benchmark \citep{liu2023mmc}              & text, image         & 2,126               & Chart Reasoning         & 2023-11-15            \\
      BenchLMM \citep{cai2023benchlmm}              & text, image         & 1,967               & Style Robustness        & 2023-12-05            \\
      TVGE \citep{wu2024better}                     & text, video         & 2,543               & New Metric              & 2024-01-15            \\
      MMCBench \citep{zhang2024benchmarking}        & text, image, speech & 4,000               & Self-consistency        & 2024-01-22
      \\

      VQAv2-IDK \citep{cha2024visually}             & text, image         & 6,624               & Hallucination           & 2024-02-15            \\

    PCA-Bench \citep{chen2024pcabench,chen2023endtoend}        & text, image& 1200               & Embodied-AI        & 2024-02-21\\
      MATH-Vision \citep{wang2024measuring}         & text, image         & 3,040               & Math Reasoning          & 2024-02-22            \\
      TempCompass \cite{liu2024tempcompass}         & text, video         & 7,540               & Video Understanding     & 2024-03-01            \\
      MMEvalPro \cite{huang2024mmevalprocalibratingmultimodalbenchmarks} & text, image & 2,138 & Reasoning, Calibration & 2024-06-29\\
      \bottomrule
    \end{tabular}
  }

  \label{tab:benchmarks_tab}
\end{table}

\subsection{Evaluation}
\label{sec: evaluation_dataset}

The evaluation MMNTP models is crucial to understand their capabilities, limitations, and potentials across different dimensions.  This section delves into the different facets of evaluating such models, outlining both established holistic benchmarks and emerging evaluation practices.

\subsubsection{Holistic Evaluation}
In the assessment of multi-modal large language models, holistic benchmarks serve as foundational tools for evaluating the integration and interplay between different modalities such as image, text, and video.

Within the domain of image-language, benchmarks like MME \citep{fu2023mme} offer a comprehensive evaluation of models' perception and cognition abilities across a diverse set of tasks, emphasizing the importance of intuitive and quantifiable analysis without the need for extensive prompt engineering.
MMBench \citep{liu2023mmbench} extends this by incorporating a vast dataset and a unique evaluation strategy, CircularEval, to robustly test models across a wide array of capabilities, including object localization and social reasoning, through single-choice questions derived from a broad spectrum of ability dimensions.
SEED-Bench \citep{li2023seedbench} and its successor SEED-Bench-2 \citep{li2023seedbench2} further contribute by providing a detailed assessment framework that covers generative comprehension capabilities across various dimensions, utilizing a mix of automatic filtering and manual verification to ensure the relevance and quality of questions.
MLLM-Bench \citep{ge2023mllmbench} aims to reflect user experiences more accurately by focusing on diverse scenarios ranging from perception to creative output, highlighting the gaps in performance between existing models and suggesting directions for future development.
MMMU \citep{yue2023mmmu} uniquely challenges models on college-level subject knowledge across a wide range of disciplines, requiring advanced perception and reasoning over complex multi-modal questions.
CMMMU \citep{zhang2024cmmmu} is designed to assess the proficiency of multimodal models in Chinese, featuring 12,000 questions across six disciplines and 30 subjects. It challenges models with complex reasoning tasks and a variety of image types.

In the video-language category, benchmarks like MVBench \citep{li2024mvbench} specifically target the temporal understanding capabilities of models by focusing on dynamic, video-based reasoning tasks that extend beyond static image understanding.
This involves evaluating models on their ability to interpret action sequences, object interactions, and scene transitions within video content.
VBench \citep{huang2023vbench} offers a nuanced approach to assessing video generation quality by breaking down the evaluation into specific dimensions and providing detailed feedback on models' performance across various content types, thereby enhancing our understanding of video generative models.



\begin{figure}[!t]
  \centering
  \includegraphics[width=\columnwidth]{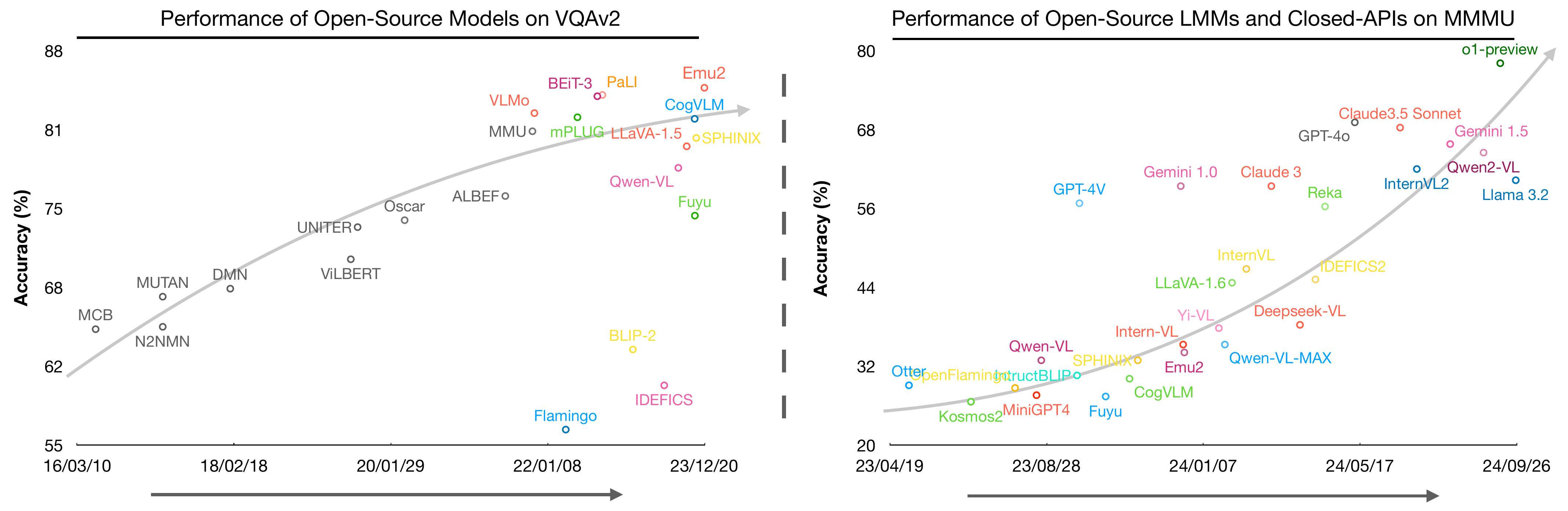}
  \caption{
    Performance evaluation of various models on the VQAv2~\citep{balanced_vqa_v2} and MMMU~\citep{yue2023mmmu} benchmarks.
    Colored representations signify the employment of the next token prediction architecture, whereas gray depictions denote alternative architectural frameworks.
  }
  \label{fig:comp_on_understanding}
\end{figure}

\begin{figure}[!t]
  \centering
  \includegraphics[width=\columnwidth]{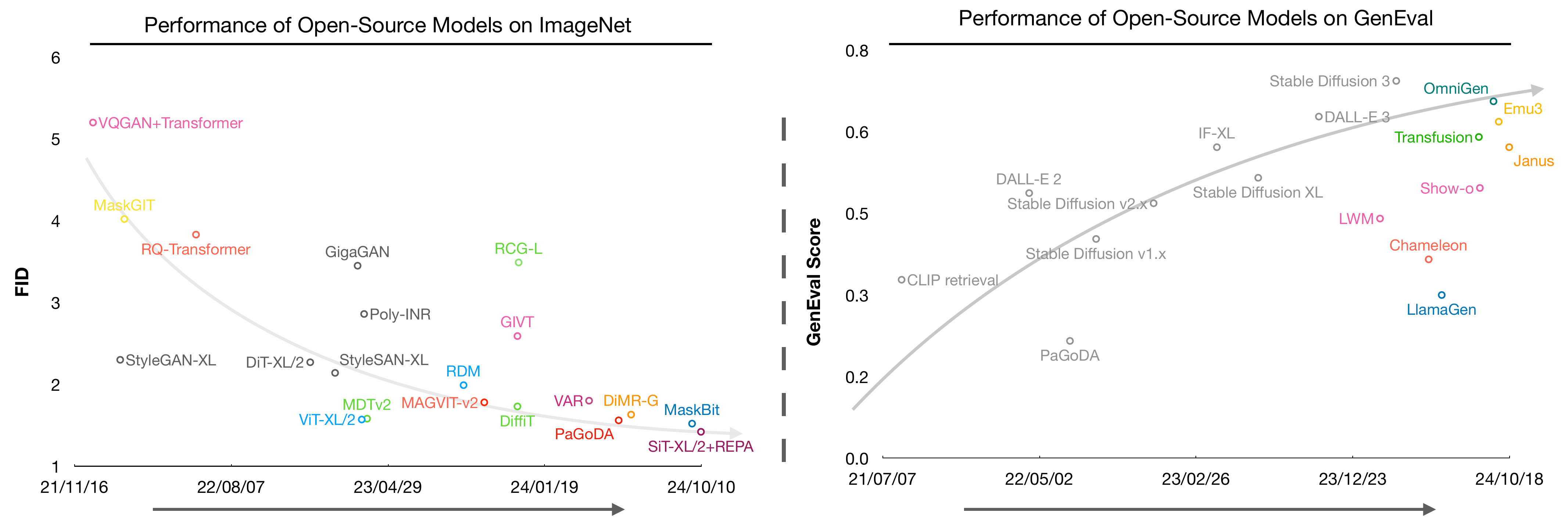}
  \caption{
    Performance evaluation of various models on the ImageNet~\citep{russakovsky2015imagenet} and GenEval~\citep{ghosh2023genevalobjectfocusedframeworkevaluating} benchmarks.
    Colored representations signify the employment of the next token prediction architecture, whereas gray depictions denote alternative architectural frameworks.
  }
  \label{fig:fig:comp_on_generation}
\end{figure}

\subsubsection{Emerging Evaluation Benchmarks}
Emerging benchmarks delve into more specialized and advanced aspects of multi-modal understanding, pushing the boundaries of model assessment.
SparklesEval \citep{huang2023sparkles} focuses on conversational competence in multi-modal contexts, emphasizing the ability of models to maintain coherent conversations involving multiple images and dialogue turns.
MathVista \citep{lu2024mathvista} challenges models on their mathematical reasoning abilities within visual contexts, incorporating a wide range of tasks that require a blend of visual understanding and compositional reasoning.
HallusionBench \citep{guan2024hallusionbench} is designed to test models on their ability to handle nuanced visual interpretations, particularly in the context of image-context reasoning, while Bingo \citep{cui2023holistic} addresses the critical issue of hallucinations in models, focusing on understanding and quantifying biases and interference effects.
MMC-Benchmark \citep{liu2023mmc} stands out for its focus on chart understanding, offering a unique set of tasks that evaluate models' abilities to extract and reason with information from visual charts, marking a significant challenge for even advanced models.
BenchLMM \citep{cai2023benchlmm} assesses performance across different visual styles, crucial for understanding and improving visual reasoning capabilities in diverse real-world scenarios. Lastly, TVGE \citep{wu2024better} introduces a novel metric, the Text-to-Video Score (T2VScore), for evaluating text-to-video generation models, providing a comprehensive tool for assessing alignment with textual descriptions and overall video quality.
MMCBench \citep{zhang2024benchmarking} is designed to evaluate LMMs robustness and self-consistency across text, image, and speech modalities, focusing on four generation tasks: text-to-image, image-to-text, text-to-speech, and speech-to-text.
The purpose of VQAv2-IDK \citep{cha2024visually} is to challenge and evaluate models on their ability to recognize and admit uncertainty or lack of information in visual question answering tasks, rather than generating incorrect or arbitrary responses.
Math-Vision \citep{wang2024measuring} benchmark is a comprehensive dataset of 3,040 mathematical problems with visual contexts, spanning 16 subjects and 5 difficulty levels, aimed at evaluating the reasoning capabilities of LLMs in mathematical scenarios.
TempCompass \cite{liu2024tempcompass} benchmark assesses Video LLMs' understanding of temporal dynamics in videos through diverse tasks and formats, highlighting significant gaps in models' ability to perceive time, using an LLM-based automatic evaluation method. 
MathVerse \cite{zhang2024mathverse} benchmark offers varying degrees of textual and image information content in multi-modality math problems, contributing to 2,612 test samples in total to investigate the ability of VLMs to gain information from pictures.

These holistic and emerging benchmarks provide a comprehensive framework for evaluating the current capabilities and identifying the limitations of multi-modal large language models, guiding the path towards more sophisticated, versatile, and capable multi-modal AI systems.

\section{Challenges}
\label{sec:challenges}
In this section, we propose four currently unsolved challenges, 
primarily stemming from the MMNTP training paradigm. 
We also recommend that the readers refer to surveys that address other open challenges, such as evaluation of multimodal LLMs~\citep{huang2024surveyevaluationmultimodallarge,fu2024mmesurveycomprehensivesurveyevaluation}, efficient LMM architectures~\citep{jin2024efficientmultimodallargelanguage}, generative approaches and auto-regressive models for vision~\citep{jiang2024surveyvisionautoregressivemodel}.

\subsection{Scaling Up MMNTP Models with Unlabeled Multimodal Data} 
\label{sec:challange_scaling_law}

The utilization of abundant unlabeled text data is one key to the success of LLMs~\citep{zhao2024surveylargelanguagemodels}. However, the potential of using large amounts of unlabeled multimodal data in training MMNTP models has not been fully investigated. Most large multimodal models currently rely on labeled pair-wise data, such as image-caption or audio-text pairs during training. Recent studies~\citep{chameleonteam2024chameleon,yang2024visionmodelpretraininginterleaved,luo2024deem} have attempted to leverage the enormous amounts of interleaved text-image data available on the internet. However, their performance on downstream tasks does not provide a significant advantage over models trained solely on pair-wise data, such as LLaVA~\citep{liu2023llava} and Qwen2-VL~\citep{Qwen2vl}. Consequently, determining how to better utilize unlabeled multimodal data such as text-image interleaved webpages, rich-text figures, text-free audios, videos, screenshots of graphical user interface (GUI), etc., remains a crucial question in the effort to scale up MMNTP models.

From another perspective, the benefits from scaling up unlabeled text data and model sizes can be curved with Scaling Law~\citep{kaplan2020scalinglaw,hoffmann2022trainingcomputeoptimallargelanguage}, which forms a fundamental basis and faith for the development of LLMs. It elucidates the intricate relationship between model performance, model size, and the amount of training data, while also guiding the optimal allocation of computational resources during LLM training. However, the return of scaling MMNTP models remains largely under-explored. Some studies \citep{aghajanyan2023scalinglawsgenerativemixedmodal,sun2024scalinglawhypothesismultimodal,VAR} have explored or hypothesized the scaling behaviors of MMNTP models in the self-supervised training manner. According to \citet{aghajanyan2023scalinglawsgenerativemixedmodal}, data from distinct modalities exhibit varying scaling behaviors. However, the reasons behind these differences and their impact on the performance of downstream tasks across various modalities remain unclear. Furthermore, it is still uncertain whether MMNTP models can develop similar emergent abilities \citep{wei2022emergent} in downstream tasks as LLMs do when the training is scaled up.


\subsection{Overcome Modalities' Interference and Boost Synergy in MMNTP Training}
\label{sec:challange_opti}

The second challenge comes with the multitask learning nature of MMNTP models, where predicting tokens belonging to different modalities could be viewed as different tasks. A critical challenge within this framework is maintaining the performance of each individual task from interference while investigating whether tasks from different modalities can provide mutual assistance~\citep{crawshaw2020multitasklearningdeepneural}.

A primary concern here is modality interference~\citep{zhang2024pretrainedlanguagemodelshelp}, where the performance of one task might negatively impact another due to conflicting information or noise from different modalities. Recent studies underscore how jointly training multimodal tasks in an NTP fashion can pose optimization challenges, especially when a single transformer decoder model is used to generate both text and image outputs \citep{chameleonteam2024chameleon}. To mitigate issues like gradient norm explosion, techniques such as QK-Norm have been employed \citep{henry2020querykeynormalizationtransformers}. However, the root causes of these optimization difficulties in MMNTP models remain largely unexplored. Further evidence of interference is seen when MMNTP models are built on pretrained large language models, as the language capability often deteriorates when adding more modalities  \citep{zhang2024pretrainedlanguagemodelshelp}. This demonstrates that interference can lead to suboptimal learning outcomes, as the model struggles to balance competing demands from the different modalities.



\subsection{Increase Efficiency in the Training and Inference of MMNTP Models}
\label{sec:challange_eff}

Efficiency remains a long-standing goal in the training and deployment of deep learning models~\citep{Menghani_2023}. Due to the similarities in backbone model structures and NTP training objectives between LLMs and MMNTP models, many advanced methods designed to enhance the efficiency of LLMs~\citep{wan2024efficientlargelanguagemodels} can also be effectively applied to MMNTP models. However, there are also several new challenges arising in the training and inference of MMNTP models due to the involvement of data from different modalities.


\paragraph{Training System Efficiency}

A big challenge in scaling up MMNTP models lie in the low efficiency in training large-scale MMNTP models on massive GPUs due to the inherent heterogeneity of both models and data across different modalities. Different from the text (1D) data in LLMs, MMNTP training involves high dimensional representations such as image (2D) and video (3D) data, where little has been done to optimize the training of these models from a system perspective. In particular, MMNTP exhibits scaling dependence with modality encoders. Recent studies have found that substantial idle GPU time (GPU Bubble) arises from the complex data dependencies when training MMNTP models that employ both visual encoders and a backbone LLM with pipeline parallelism~\citep{scale1}. To address the issue, several efficient and adaptive training frameworks have been developed to optimize the scheduling of encoder computations and overlap GPU communication with computation. For example, Optimus reduces training time by decomposing image encoder layer computations into smaller kernels and scheduling those kernel executions within LLM bubbles, minimizing the pipeline idle time~\citep{scale2}. DistTrain leverages disaggregated model orchestration and data reordering to improve the training efficiency and scalability of MMNTP, achieving significant improvements in model FLOPS utilization and throughput~\citep{scale3}. Despite these advancements, the unique challenges posed by multimodal architecture, how to develop more advanced system optimizations to train MMNTP on large-scale production clusters with thousands of GPUs remains an open research question. 

MMNTP training also exhibit highly variable sequence lengths. As MMNTP models move towards more complex tasks such as multi-image reasoning, multi-modal RAG, video understanding, supporting MMNTP with long and variable sequence length becomes critical. However, existing large model training systems and the underlying parallelism technologies (data, tensor, pipeline) are limited in their ability to support efficient long sequence training. Recent studies have proposed several sequence parallelism techniques, such as DeepSpeed-Ulysses~\citep{scale4}, Ring-Attention~\citep{scale5}, and Unified Sequence Parallelism~\citep{scale6}, to enable long sequence training for LLMs. However, applying these sequence parallelism strategies to MMNTP needs to take careful consideration in handling heterogeneous data from different modalities, each with distinct characteristics and sequence lengths, which motivates advanced system-algorithm co-design to address this challenge.

\paragraph{Multimodal Tokenization Efficiency in MMNTP Models}

As mentioned in the tokenization section, multimodal input like image, audio and video originally resides in a continuous space, which contains a lot of redundant information. The multimodal tokenization process has large room for efficiency improvement, where the core question is: can we use less tokens to represent the multimodal input while maintaining the performance? In the scope of single-modal image modeling and dual-encoder VLM architectures, various model compression methods—such as pruning, knowledge distillation, and quantization—are applied to accelerate image encoders. Model pruning \citep{zhu2021vision,Lin_2024_CVPR} sparsifies the encoder backbone and removes certain modules. Knowledge distillation \citep{yuan2021tokens,wu2023tinyclip} utilizes soft labels as supervision and train competent smaller dense models. Model quantization \citep{yu2022unified,liu2021post} replaces models and computation to low-precision counterparts. Though proven effective in single-model scenarios, the adaptability of these methods to training MMNTP models remains largely unexplored.

\paragraph{Modeling Efficiency in Understanding and Generation}

Although image tokens occupy a significant portion of the input sequence of MMNTP models, it is discovered that the the LLM backbone only pays a small portion of attention to the image tokens compared to the language tokens~\citep{chen2024image}. This phenomenon raises the question  whether we can reduce the number of image tokens during training and inference without sacrificing performance as they take up most of the sequence length but gain the least attention from the model. In single-modal image modeling, these tokens can be largely pruned using pre-trained priors \citep{marin2021token,xiong2024pyra} or through entirely training-free methods \citep{bolya2022token}, with minimal impact on performance. For MMNTP models such as Llava~\citep{liu2023llava} and QwenVL~\citep{QwenVL}, \citet{chen2024image} proposed a pruning approach that removes most image tokens without compromising performance. However, the reason behind the redundancy of image tokens and how to leverage this phenomenon remains underexplored for MMNTP models of different modalities.

In the realm of multimodal generation, the challenge of modeling efficiency is equally pronounced. A central issue is how to define an effective generative training objective that suits the Next Token Prediction (NTP) manner for various modalities, given that data from different modalities possess distinct structures. A vanilla NTP training objective may not adequately address these differences, prompting the development of specialized objectives tailored to the characteristics of each modality. For instance, methods like MaskGIT~\citep{MaskGIT}, MAGViT~\citep{magvit}, VAR~\citep{VAR}, and DnD-Transformer~\citep{dnd-transformer} have been proposed to better accommodate the unique aspects of different modalities. Moreover, enhancing generation quality efficiently can be achieved through post-generation refinement techniques such as super-resolution \citep{lu2023unifiedio2}. Super-resolution methods aim to upscale the outputs of language models by either fine-tuning the backbone model \citep{CogView,CogView2} or by incorporating additional modules \citep{text2image2,kondratyuk2023videopoet}. Despite significant advancements in this area, diffusion models remain the de facto model in visual generation applications. However, a detailed comparison of generation quality and efficiency between MMNTP models and diffusion models is still lacking in the literature, leaving room for further exploration and research.

\subsection{MMNTP as Universal Interfaces}
\label{sec:challange_exte}

LLMs have exhibited notable advancements in collaborating with external models in the framework of NTP~\citep{qin2023toolllm,shen2023hugginggpt,gupta2022visprog,hao2022languagemodelsgeneralpurposeinterfaces}. It highlights the growing potential for language models to extend their capabilities beyond mere text generation. In this survey, we have mentioned that the next token prediction paradigm has been unifying vision, audio and different multimodal task. However, an ultimate challenge is beyond current explored modalities, achieving a universal interface connecting tasks from various sources, such as robotics~\citep{brohan2023rt2}, molecular~\citep{Flam_Shepherd_2022} and proteins~\citep{Ruffolo2024DesigningPW}. The key problem is how to formulate a different task as next token prediction, and whether such formulation is efficient and scalable in solving the problem, which is largely underexplored outside the language, vision and audio data. 

\paragraph{Design NTP Training Objectives for Different Modalities}

For non-text modalities, simply linearizing the data into a 1D sequence and conducting NTP (Next Token Prediction) training may not be the most effective approach. This strategy poses two potential issues: it can overlook the inherent structure of multimodal data and lead to excessively long sequence lengths. For instance, the spatial relationships in images and temporal relationships in videos are crucial elements that the traditional NTP training fails to consider. The number of tokens required to represent an image or video increases linearly with the image's resolution and the video's duration. To address these challenges, several approaches have been developed to adapt NTP training objectives to various modalities, such as images and videos. Notable examples include MaskGiT~\citep{MaskGIT}, VAR~\citep{VAR}, DnD-Transformer~\citep{dnd-transformer}, and Next-Block-Prediction~\citep{anonymous2024next}. These efforts aim to better capture the unique structures present in different types of data and can reduce the inference time by generating multiple tokens at one time.

\paragraph{Comparison to Diffusion}

Diffusion models represent another popular framework for generative modeling and they have been extensively applied to multimodal data beyond the original image modality~\citep{yang2024diffusionmodelscomprehensivesurvey}. These applications extend to areas such as language~\citep{li2022diffusionlmimprovescontrollabletext}, robotics~\citep{chi2024diffusionpolicyvisuomotorpolicy}, and drug discovery~\citep{huang2024proteinligand}. When comparing NTP and diffusion models, both approaches share the fundamental idea of breaking down a complex generation task into multiple, more manageable steps. The major difference between these two approaches lies in how the task is decomposed. NTP breaks down the data according to its dimensional order, whereas diffusion models deconstruct the data globally in a coarse-to-fine manner. There is no consensus on which method is superior, and new approaches are emerging that combine these two modeling techniques. For instance, auto-regressive diffusion models~\citep{MAR,Transfusion} incorporate elements of both strategies. Exploring how different modeling methods perform on various modalities is a fascinating frontier in the field.

\section{Contributions and Acknowledgments}
\label{sec: ack}
\textbf{Liang Chen} leads the project. \textbf{Zekun Wang}, \textbf{Shuhuai Ren}, \textbf{Lei Li}, \textbf{Haozhe Zhao}, and \textbf{Yunshui Li} are the core contributors who drafted the initial version of this survey. \textbf{Hongcheng Guo} contributes to the general structure and tokenizer part. \textbf{Lei Zhang} contributes to the Datasets and Evaluation section. \textbf{Lingwei Meng}, \textbf{Shujie Hu}, and \textbf{Ge Zhang} contribute to the audio part. \textbf{Zefan Cai}, \textbf{Yichi Zhang} and \textbf{Ruoyu Wu} contribute to the Inference Enhancement section. \textbf{Yizhe Xiong} and \textbf{Minjia Zhang} contributes to the challenge section. \textbf{Qingxiu Dong} contributes to the In-Context Learning part and provides valuable feedback on survey writing. \textbf{Yulong Chen}, \textbf{Andreas Vlachos}, \textbf{Xu Tan}, \textbf{Junyang Lin}, \textbf{Jian Yang}, \textbf{Shuai Bai}, \textbf{Wen Xiao}, \textbf{Aaron Yee} and \textbf{Tianyu Liu} contribute to the revision and discussion throughout the writing of the survey. \textbf{Baobao Chang} is the supervisor of \textbf{Liang Chen}. We sincerely thank \textbf{Sanyuan Chen} (Meta), \textbf{Yuchen Yang} (PKU), \textbf{Jiarui Hai} (JHU), \textbf{Li Dong} (Microsoft) for their insightful and valuable feedback. The survey will be updated periodically, we also kindly welcome all kinds of comments and suggestions.

\bibliographystyle{ACM-Reference-Format}
\bibliography{sample-base}

\end{document}